\documentclass{article}

\usepackage{arxiv}
\usepackage{subcaption}    
\usepackage{url}            
\usepackage{booktabs}       
\usepackage{makecell}
\usepackage{array}
\usepackage{amsfonts}       
\usepackage{nicefrac}       
\usepackage{microtype}      
\usepackage{lipsum}
\usepackage{newunicodechar}
\usepackage{graphicx}
\usepackage{hyperref}       
\usepackage{natbib}

\setcitestyle{numbers}
\setcitestyle{sort}
\setcitestyle{compress}

\usepackage{caption}
\usepackage{listings}
\usepackage{xcolor}
\usepackage{minted}
\usepackage{float}
\usepackage{longtable}
\newcolumntype{L}[1]{>{\raggedright\arraybackslash}m{#1}} 
\floatstyle{ruled}
\newfloat{listing}{htbp}{lst}
\floatname{listing}{Listing}
\usepackage{multirow}
\usepackage{soul}
\usepackage{standalone}
\usepackage{pdflscape} 
\usepackage{rotating}
\usepackage{tabularx}
\usepackage[utf8]{inputenc}

\usepackage[english]{babel}
\usepackage{changepage} 

\usepackage[edges]{forest}
\usepackage{tikz}
\usepackage[breakable]{tcolorbox} 
\usetikzlibrary{arrows.meta}

\tcbuselibrary{skins}

\definecolor{bluec}{HTML}{0b5394}
\definecolor{redc}{HTML}{980000}
\definecolor{greenc}{HTML}{5b8f4a}

\newtcolorbox{genericbox}[3][blue]{
    enhanced,
    breakable,
    colback=#1!5!white,
    colframe=#1!75!black,
    colbacktitle=#1!15!white,
    coltitle=#1!50!black,
    fonttitle=\bfseries,
    title=#2,
    arc=2mm,
    boxrule=0.8pt,
    titlerule=0mm,
    toptitle=1mm,
    bottomtitle=1mm,
    left=3mm,
    right=3mm,
    top=2mm,
    bottom=2mm,
    attach boxed title to top left={xshift=5mm, yshift=-2mm},
    boxed title style={
        enhanced,
        colback=#1!15!white,
        colframe=#1!75!black,
        arc=1.5mm,
        boxrule=0.6pt,
        left=2mm,
        right=2mm,
        top=1mm,
        bottom=1mm
    }
}

\title{IslamicLegalBench: Evaluating LLMs Knowledge and Reasoning of Islamic Law Across 1,200 Years of Islamic Pluralist Legal Traditions \thanks{This manuscript has been submitted for review to Artificial Intelligence \& Law.}}

\author{%
  Ezieddin Elmahjub$^1$, \quad
  Junaid Qadir$^1$, \quad
  Abdullah Mushtaq$^{1}$, \quad
  Rafay Naeem$^{1}$, \\
  Ibrahim Ghaznavi$^{2}$, \quad
  Waleed Iqbal$^3$$^,$$^4$ \\
  $^1$Qatar University \quad
  $^2$Information Technology University \quad
  $^3$Northeastern University \\
  $^4$Queen Mary University of London \\
}

\begin{document}
\maketitle

    
\begin{center}
\textbf{ABSTRACT}
\end{center}

\begin{adjustwidth}{1cm}{1cm}
As millions of Muslims worldwide turn to LLMs like GPT, Claude, and DeepSeek for religious guidance, a critical question emerges: \emph{Can these AI systems reliably reason about Islamic law?} This paper introduces \texttt{IslamicLegalBench}, the first multi-school benchmark for evaluating LLM performance across a broad range of Islamic jurisprudence (\textit{fiqh}). Drawing from 38 foundational legal texts spanning 1,200 years and seven schools of jurisprudence, we create 718 evaluation instances across 13 tasks, manually collected and organized by complexity, from basic recall to sophisticated reasoning including legal rationale identification (\textit{`illah}), analogical application (\textit{qiyās}), and cross-school synthesis. Our evaluation of nine state-of-the-art LLMs reveals significant limitations. Even the best model achieves only 67.65\% correctness with 21.25\% hallucination; several models demonstrably fail with correctness below 35\% and hallucination exceeding 55\%. Further concerning, few-shot prompting provides minimal to no improvement, only 2 of 9 models improve by $>$1\%, exposing that \emph{LLMs lack adequate Islamic legal knowledge in their training data, and prompting cannot compensate for absent foundational knowledge}. Our analysis reveals why: moderate-complexity tasks (from a human expert's perspective) requiring exact, verbatim knowledge (e.g., enumerating contract conditions, synthesizing statutory articles) show consistently high error rates and hallucination rates up to 73\%, whereas high-complexity tasks show better performance because models rely on semantic generalization and verbose reasoning, projecting competence while lacking precise textual understanding. False premise detection reveals risky sycophantic behavior: 6 of 9 models accept misleading assumptions at rates exceeding 40\% (worst: 86.27\%). Critically, few-shot prompting worsens sycophancy by 3.49 percentage points. The strong and statistically significant negative correlation (Pearson's r = –0.87, p $<$ 0.01) between the false premise acceptance rate and overall performance indicates that models with higher false Islamic query acceptance rates tend to exhibit lower overall reasoning accuracy. Our findings carry important implications: the path forward for Islamic NLP lies not in lightweight post-hoc techniques such as prompt or instruction tuning but in enriching knowledge, the Islamic NLP community must prioritize training models on comprehensive large scale Islamic legal corpora spanning classical Hadith collections, jurisprudential works from the major Islamic law schools (\textit{madhabs}), and codified legal compendia. \texttt{IslamicLegalBench} provides the first systematic evaluation framework for Islamic legal AI systems, revealing critical limitations in platforms that Muslims increasingly rely on for spiritual guidance.
\end{adjustwidth}


\section{Introduction}
\label{sec:Introduction}

As large language models (LLMs) are increasingly deployed in sensitive domains such as law, medicine, and religion, the challenge of evaluating their reliability grows ever more pressing. Islamic law (\textit{fiqh}) is especially significant in this regard: it governs the ethical norms, legal rulings, and everyday practices of nearly two billion Muslims worldwide, shaping areas as diverse as contracts, finance, family relations, criminal behavior, dietary norms, among other forms of behavioral governance \cite{Khair_Sawalha_2025_AutomatedTranslation}. Islamic jurisprudence draws upon a complex interplay of sources, such as the Qur'an, Prophetic traditions (\textit{Sunnah}), scholarly consensus (\textit{ijmā`}), and analogical reasoning (\textit{qiyās}), and is further marked by a plurality of interpretive traditions. The four major Sunni schools and various Shi`i traditions each employ distinct methodological frameworks (\textit{uṣūl al-fiqh}) and develop their own interpretive approaches. Unlike codified Western legal systems that rely heavily on precedent or codified legislation, Islamic jurisprudence emphasizes deep engagement with foundational texts, nuanced interpretive reasoning, and sensitivity to social and historical context \cite{atif2025sacredsyntheticevaluatingllm}. This combination of doctrinal diversity, methodological rigor, and context-dependent application makes fiqh a technically demanding and ethically consequential domain for AI evaluation.

The need for systematic evaluation of LLMs in Islamic legal contexts is increasingly urgent. While these models have generated interest for their potential to assist with complex legal and religious reasoning \cite{aldahoul2025benchmarking}, their reliability in such sensitive domains remains largely unexplored \cite{atif2025sacredsyntheticevaluatingllm}. A particularly critical concern is \textit{hallucination}---a pervasive but not uniformly defined phenomenon where LLMs generate plausible yet factually incorrect, fabricated, or ungrounded information. The very definition of hallucination remains contested and context-dependent \cite{venkit2024hallucinations, Dahl_second}: What constitutes hallucination in general question-answering differs from legal contexts, which in turn differs from religious and jurisprudential domains \cite{Ho2024_HallucinatingLaw, Dahl_first}. In secular legal systems, hallucination might involve fabricated case citations or misrepresented precedents. However, in Islamic legal contexts, hallucination takes on qualitatively different and more severe forms: fabricated religious rulings (\textit{fatāwā}), misattributed scholarly opinions, incorrectly cited textual evidence, or invented legal principles can distort religious practice, undermine scholarly authority, and produce spiritual and societal consequences distinct from civil legal errors \cite{Imtiaz_Shafique_2025_AlIrfan_LLMs}. When Muslims consult LLMs for guidance on matters affecting their spiritual well-being and religious guidance, understanding and detecting hallucination becomes a technical challenge with significant ethical implications.

Current evaluation frameworks face several critical gaps when applied to Islamic legal reasoning. Existing benchmarking efforts are either specialized in single domains \cite{aldahoul2025benchmarking} or limited to subsets of jurisprudential schools \cite{atif2025sacredsyntheticevaluatingllm}. Existing evaluation frameworks are often limited to a single jurisprudential school or a narrow set of legal topics. No evaluation framework spans multiple domains across both Sunni and Shī'a traditions across diverse legal topics. Multi-domain and cross-school evaluation presents distinct challenges: different domains operate under varying epistemic principles (e.g., ritual worship prioritizes textual fidelity while transactions emphasize societal welfare), and different jurisprudential schools apply divergent methodological frameworks to identical source texts, producing legitimately different valid rulings. This requires assessing reasoning coherence within each school's methodology rather than against a single correct answer. Moreover, current approaches inadequately test the sophisticated reasoning capabilities required for authentic Islamic legal analysis, such as identifying underlying legal rationales (\textit{`illah}), applying analogical reasoning to novel scenarios, and navigating the multi-school nature of Islamic law where the same question may have different valid answers depending on methodological approach.

This paper introduces \textbf{\texttt{IslamicLegalBench}}, the first pluralist benchmark for evaluating LLM performance across major areas of Islamic jurisprudence. We develop a multi-tiered evaluation framework spanning 1,200+ years of Islamic legal tradition, drawing from 38 foundational legal texts across several schools of jurisprudence. Our benchmark comprises 718 manually curated evaluation instances organized into 13 task types across three complexity tiers: Low (bibliographical and factual recall), Moderate (condition enumeration, comparative distinctions, statutory synthesis), and High (legal rationale identification, analogical application, cross-school synthesis, maxim mapping). We conduct a systematic evaluation of 9 state-of-the-art LLMs using both zero-shot and few-shot prompting, employing both human domain-experts and an LLM-as-a-Judge framework for fine-grained assessment across correctness, hallucination, abstention, and false premise detection. Our analysis reveals critical insights into the limitations of current LLMs for Islamic legal reasoning, including the failure of in-context learning, problematic confidence-hallucination patterns, and significant performance disparities between closed and open-source models.

\subsection{Contributions}

Our work makes the following scholarly contributions that extend the state of the art:

\begin{enumerate}
    \item \textbf{A Novel Multi-School Benchmark for Islamic Law:} We develop a novel evaluation framework that is the first to systematically incorporate seven schools of Islamic jurisprudence, including both Sunni and Shi`i traditions. Our benchmark is designed to test a wide array of tasks, from foundational recall to complex analogical reasoning.

    \item \textbf{Discovery of In-Context Learning Failure:} We tested 9 state-of-the-art LLMs and revealed that few-shot prompting provides minimal benefit for Islamic legal reasoning (7 of 9 models show $<=$1\% improvement). Our analysis uncovers the mechanism: moderate-complexity tasks requiring accurate knowledge exhibit consistent failure, while high-complexity tasks allow semantic generalization, projecting competence despite lacking precise textual knowledge.
    
    \item \textbf{Identification of Critical Safety Vulnerabilities:} We expose risky failure modes including confident incorrectness (correctness $<$50\%, hallucination $>$25\%, abstention $<$1\%), widespread sycophantic behavior (six models accepting false premises $>$40\%), and few-shot prompting worsening sycophancy by 3.49\%. We also quantify significant performance disparities between top closed and open-source models: 19.76\% in correctness and 13.32\% in hallucination rates.
    
    \item \textbf{Benchmark Release:} We release 20\% of \texttt{IslamicLegalBench} publicly as a development set, maintaining 80\% held-out to preserve evaluation integrity and prevent data contamination.
\end{enumerate}

Together, these contributions provide the first systematic understanding of LLM capabilities and limitations in Islamic jurisprudence, revealing that current models fundamentally lack the Islamic legal knowledge required for reliable deployment and that addressing this gap requires training on comprehensive Islamic legal corpora. Our dataset achieves comprehensiveness through several dimensions: it spans multiple jurisprudential schools, covers a diverse range of legal topics across core domains of Islamic law, and draws from over 1200 years of Islamic legal scholarship, ensuring temporal breadth. This multi-dimensional coverage captures the methodological diversity and evolutionary depth essential to Islamic jurisprudence (detailed further in Section~\ref{sec:dataset}).

\subsection{Organization of the Paper}
The remainder of this paper is organized as follows: Section~\ref{sec:relatedwork} reviews related work in legal AI benchmarking, Islamic NLP datasets, and scholarly perspectives on AI in Islamic jurisprudence. Section~\ref{sec:dataset} describes our dataset creation process, including the Source Dataset of 271 question-answer pairs from 38 authoritative texts and the Benchmarking Dataset consisting of 718 instances with 13 task types across three complexity levels. Section~\ref{sec:setup} presents the experimental setup, including the 9 evaluated LLMs and prompting strategies. Section~\ref{sec:evaluation} details our evaluation methodology using the LLM-as-a-Judge framework with comprehensive scoring criteria. Section~\ref{sec:results} presents results across correctness performance by complexity tier, hallucination patterns, false premise acceptance rates, and in-context learning effectiveness. Section~\ref{sec:discussion} discusses architectural limitations in legal reasoning and their alignment with broader LLM performance patterns. Section~\ref{sec:limitations} addresses limitations and future directions. 

\section{Related Work}
\label{sec:relatedwork}

This section surveys prior work on LLM evaluation in legal domains, examining both general legal benchmarking and Islamic-specific efforts. We organize our review into three areas: general legal AI benchmarking that provides methodological foundations, foundational datasets and benchmarks in Islamic contexts, and ethical and scholarly perspectives on AI in Islamic jurisprudence. From these strands, we identify critical gaps that motivate our comprehensive benchmark.

\subsection{LLM Benchmarking in General Legal Contexts}
\label{subsec:generallegalrelatedwork}

The intersection of technology and legal reasoning has progressed from early conceptual explorations to systematic empirical evaluation. Susskind's \textit{The Future of Law} \cite{Susskind1998} introduced the legal academy to the disruptive potential of information technology (IT) in the mid-1990s, arguing that emerging digital systems would profoundly reshape legal practice and the processes of legal reasoning. Nearly three decades later, his work \textit{Tomorrow's Lawyers} \cite{Susskind2023} revisits these predictions in the context of contemporary artificial intelligence, analyzing how modern AI tools now perform components of legal work once assumed to require exclusively human judgment. Similarly, Grant and Wischik \cite{GrantWischik2020} provide key conceptual foundations for understanding the application of machine learning to law, highlighting how the distinctive features of legal reasoning, i.e., its reliance on precedent, interpretation, and context-sensitive judgment, create methodological challenges for AI systems. Together, these works trace a scholarly trajectory from early technological transformation arguments to contemporary concerns about how AI operationalizes legal tasks. 

Building on these conceptual foundations, recent empirical research has systematically evaluated LLMs on legal reasoning tasks, revealing specific limitations. The systematic evaluation of hallucination in legal AI systems informed our benchmark design. The work of hallucination detection in Dahl et al.~\cite{Dahl_first}, followed by Magesh et al.~\cite{Dahl_second}, assesses both general LLMs and specialized legal tools on case law queries, respectively, defining ``factual hallucination", a response unfaithful to real-world facts, as a particularly significant error in legal contexts where correct citation and accurate representation of law are essential. This definition shaped our own hallucination detection framework.

The first of these studies \cite{Dahl_first} demonstrated the value of evaluating models exclusively on their internal knowledge, without retrieval augmentation, an approach we adopt in our evaluation. Its benchmark involved approximately 200,000 queries for four models (GPT-4 \cite{openai_gpt4}, GPT-3.5 \cite{openai_gpt35}, Llama 2 \cite{meta_llama2}, and PaLM 2 \cite{google_palm2}), using 14 task templates across three complexity levels (Low, Medium, and High), which directly inspired our three-tier complexity structure. This work employed a GPT-4-based \textbf{``LLM-as-Judge" framework} to assess outputs, a methodology we adopt using o3, and revealed high mean hallucination rates across all models, underscoring the limitations of LLMs in legal reasoning without retrieval support and motivating our systematic hallucination measurement.

The subsequent study \cite{Dahl_second} examined dedicated legal tools integrating retrieval-augmented generation (RAG), including Lexis+ AI \cite{lexisplus_ai}, Westlaw AI-Assisted Research \cite{westlaw_edge}, and Ask Practical Law AI \cite{thomsonreuters_practical_law}, with GPT-4 \cite{openai_gpt4} scores as a baseline. It distinguished three outcome categories (Accurate, Hallucinated, Incomplete) and tested four capability categories: general legal research, jurisdiction- and time-specific queries, factual recall, and, notably, \textbf{false-premise queries} to probe system robustness. This false-premise testing approach directly motivated our inclusion of false premise detection tasks (see Section \ref{subsec:falsepremisevariants}). Their results showed substantial variation: Lexis+ AI achieved 65\% accuracy with 17\% hallucination, Westlaw AI 42\% accuracy with 33\% hallucination, and the GPT-4 baseline 49\% accuracy with 43\% hallucination, demonstrating that even RAG-augmented systems exhibit concerning hallucination rates. Another study finds that, at present, these LLMs struggle with performing legal reasoning tasks but may still be useful in supporting lawyers with drafting legal documents and other related services \cite{Trozze2025}.

The LegalBench benchmark \cite{guha2023legalbench} provided methodological insights that directly shaped our task design. Drawing from the IRAC (Issue, Rule, Application, Conclusion) methodology \cite{iracmethod2025}, it decomposes legal reasoning into six distinct cognitive types. Evaluation of 20 LLMs revealed three key findings that inform our approach: (1) Task-specific performance variation, validating our need for fine-grained evaluation across 13 task types; (2) Prompt design matters, with description-based prompts often outperforming simple queries, informing our prompt engineering; and (3) High sensitivity to in-context examples, which informed our systematic comparison of zero-shot versus few-shot prompting approaches.

The broader legal-AI literature shows that LLMs struggle in two recurrent ways: they hallucinate factual details and they break down on tasks requiring multi-step legal reasoning or nuanced contextual grounding \cite{Ho2024_HallucinatingLaw}. One study found hallucination rates of at least 75\% on queries about a court's core ruling \cite{Dahl_first}. A dominant error pattern is ``\textbf{Wrong Conclusion from False Premises},'' where models accept factually incorrect premises and build plausible but erroneous arguments. This contrafactual bias poses critical dangers in religious contexts where a single mistake can have profound spiritual implications, directly motivating our benchmark's inclusion of false premise detection tasks (see Section \ref{subsec:falsepremisevariants}) to test if LLMs can reject incorrect assumptions in Islamic legal queries.

\subsection{Foundational Datasets and Benchmarks in Islamic Context}

While general legal AI research provided methodological foundations, understanding the landscape of existing Islamic NLP resources revealed critical gaps our benchmark addresses. Research in Islamic NLP has progressed from general-purpose text corpora to specialized legal benchmarks. We trace this evolution through three generations of work: early sacred text corpora, domain-specific legal datasets, and recent task-specific benchmarks for Islamic legal reasoning.

\paragraph{Efforts on Islamic Text Corpora:} Early efforts compiled large-scale corpora such as Quran-NLP \cite{quran_nlp}, encompassing the Qur'an, Hadith, translations, and \textit{tafs\=ir} (exegesis) with over 700,000 Hadiths, and the 14 Books Hadiths Collection \cite{sunnah_ar_en_dataset_2025}, providing over 50,000 bilingual narrations with metadata. While these datasets enable basic NLP tasks such as translation, question-answering, sentiment analysis, and summarization, they lack the jurisprudential and school-specific annotation necessary for rigorous legal analysis \cite{Khair_Sawalha_2025_AutomatedTranslation}, revealing the need for specialized legal benchmarks. To address these limitations, specialized legal corpora emerged. AraLegal-BERT \cite{alqurishi2022aralegalbertpretrainedlanguagemodel}, pre-trained on 4.5 GB (13.7 million sentences) of Arabic legal texts, demonstrated improved accuracy on legal tasks. Fatwa datasets from IslamQA \cite{IslamQA} and Islamweb \cite{IslamWeb} provide question-answer pairs for training chatbot systems \cite{namoun2024multimodal}. However, single-source fatwa datasets carry inherent bias risks, potentially creating a ``legal monoculture'' \cite{Ho2024_HallucinatingLaw} that erases Islamic jurisprudence's diversity, constituting representational harm \cite{suresh_guttag_2019_harm_framework}, a concern that motivated our multi-school, multi-source approach spanning seven schools of jurisprudence and 38 foundational texts.

\paragraph{Recent Task-Specific Benchmarks:} With recent AI advances, specialized benchmarks have emerged for Islamic legal reasoning, providing immediate precedents for our work while also revealing limitations we sought to address. The \textit{FiqhQA} benchmark \cite{atif2025sacredsyntheticevaluatingllm} evaluates LLMs across four major Sunni schools (Hanafi, Maliki, Shafi`i, Hanbali) using 960 question-answer pairs from the Kuwaiti Fiqh Encyclopedia \cite{kuwaiti_fiqh_encyclopedia} in both English and Arabic. FiqhQA introduced \textit{abstention behavior} evaluation, the ability to recognize when not to answer, addressing concerns that LLMs hallucinate even with RAG, a metric we incorporate into our evaluation framework. Using LLM-as-a-judge evaluation \cite{bai2024mtbench, liu2023geval}, GPT-4o achieved the highest accuracy (46\% English, 28\% Arabic) but showed substantial school-specific variation (Hanafi 56\% vs Maliki 37\%), likely reflecting training data bias toward sources like \url{IslamQA.org} \cite{IslamQA}. GPT-4o was less likely to abstain when uncertain, yielding higher rates of confident incorrect answers, while Gemini showed 90\% abstention with 1\% errors. This underscores that knowing when not to answer can be as important as answering correctly in religious contexts, a finding that informed our inclusion of abstention measurement and our identification of the ``danger zone'' of confident incorrectness. However, FiqhQA is limited to four Sunni schools and a single domain (\textit{Fiqh al-`Ib\=ad\=at}), excluding Shi`i schools and complex legal reasoning, gaps our benchmark addresses. The ArabicNLP QIAS 2025 Shared Task \cite{AL-Smadi_2025_QU-NLP_QIAS} focuses on Islamic inheritance law (\textit{`ilm al-maw\=ar\=ith}) with multiple-choice questions across three difficulty levels. The QIAS 2025 dataset \cite{QIAS2025}, derived from 32,000 fatwas and verified by inheritance law experts, demonstrated that LLMs can achieve high accuracy on complex, rule-based legal reasoning when the domain is highly structured. However, its focus on a single subdomain limits generalizability, motivating our multi-domain approach. MizanQA \cite{bahaj2025mizanqabenchmarkinglargelanguage}, a Moroccan-specific dataset with approximately 1,700 multiple-choice questions, covers Maliki jurisprudence alongside French-influenced conventions but focuses on a single school, emphasizes factual recall over complex jurisprudential reasoning, and lacks cross-school comparative evaluation, limitations our benchmark systematically addresses.

From this extensive literature review, we identify that existing benchmarks exhibit noticeable limitations that our work addresses: most focus exclusively on Sunni schools, target single domains rather than comprehensive legal reasoning, and emphasize factual recall over sophisticated reasoning like identifying \textit{`illah}, applying \textit{qiy\=as}, or cross-school synthesis. Our benchmark addresses these gaps by spanning seven schools, multiple legal domains, and three complexity tiers, testing factual knowledge through advanced jurisprudential reasoning. Table~\ref{tab:benchmark_comparison} compares our \texttt{IslamicLegalBench} with existing Islamic legal benchmarks.

\begin{table}[htbp]
\centering
\caption{Comparison of Islamic Legal Benchmarks}
\label{tab:benchmark_comparison}
\footnotesize
\begin{tabular}{@{}p{2.6cm}p{3.4cm}p{2.7cm}p{2.5cm}p{2.7cm}@{}}
\toprule
\textbf{Feature} & \textbf{IslamicLegalBench (Ours)} & \textbf{FiqhQA} \cite{atif2025sacredsyntheticevaluatingllm} & \textbf{QIAS 2025} \cite{QIAS2025} & \textbf{MizanQA} \cite{bahaj2025mizanqabenchmarkinglargelanguage}\\
\midrule
Dataset Size & 718 evaluation instances & 960 Q\&A pairs & \(\sim\)22,000 MCQs (20k train, 1k val, 1k test) & \(\sim\)1,700 MCQs \\
\midrule
School Coverage & 7 schools (4 Sunni + 3 Sh\=i'\=i) & 4 Sunni schools & Sunni inheritance law focus & Primarily Maliki (with customary/civil-law elements) \\
\midrule
Legal Domains & Multi-domain: contract law, family law, property law, etc & Worship only (`Ib\=ad\=at) & Inheritance law & Moroccan legal system (Islamic + civil + customary) \\
\midrule
Task Types & QA on 13 tasks incl. recall, synthesis, `illah, qiy\=as, cross-school & QA with abstention on Worship & Multiple-choice questions & Multiple-choice questions (incl. multi-answer) \\
\midrule
Complexity & 3 tiers: Low / Moderate / High & - & 3 difficulty levels & - \\
\midrule
Source Material & 38 classical texts (~1,200 years) & Kuwaiti Fiqh Encyclopedia & IslamWeb fatwas & Moroccan legal exam sources \\
\bottomrule
\end{tabular}
\end{table}

\subsection{Ethical and Scholarly Perspectives on Islamic AI Applications}
\label{subsec:muslimsscholars}

The technical challenges identified in LLM legal reasoning take on heightened significance in Islamic contexts, where errors carry profound spiritual consequences, a reality that shaped our evaluation priorities and interpretation of findings. Contemporary and classical scholarly consensus affirms that non-human systems cannot replicate the essential qualities required for religious guidance, as qualified human scholars possess attributes that AI fundamentally lacks.

\paragraph{Contemporary and Classical Scholarly Consensus:}
While some scholars cautiously endorse LLMs as supplementary research tools under strict oversight, many maintain strong reservations that informed our assessment of when LLMs are ``ready'' for deployment. Islamic Religious Authorities across multiple jurisdictions emphasize that independent reasoning (\textit{ijtih\=ad}) and religious rulings must remain exclusively with qualified human scholars. Egypt's Dar al-Ift\=a warns that AI applications cannot distinguish between true and false content \cite{rel15050541}, a limitation our hallucination metrics empirically confirm. Similarly, Indonesia’s Nahdatul Ulama Council prohibits chatbots from issuing fatwas, noting that AI lacks the wisdom and intuition of human \textit{muftis} \cite{researchgate_ai_fatwa}. Iranian clerical leaders also warn that ``robots can't replace senior clerics'' \cite{futurism_iran_ai}.

This scholarly position reflects classical Islamic jurisprudential principles that provide interpretive context for our technical findings. Ibn Qayyim al-Jawziyyah argued that failure to understand context is an error, as the same legal principle can yield different outcomes depending on circumstances \cite{qayyim1991}, a test AI systems often fail, as our false premise detection results demonstrate. Al-Nawaw\=i highlighted that qualified jurists must be aware of legitimate differences of opinion to provide balanced rulings \cite{nawawi1996}, AI models risk oversimplifying issues or promoting single viewpoints, a concern validated by our school-specific performance analysis. Ibn `Abd al-Barr articulated that a mufti must embody moral integrity (\textit{`ad\=alah}) and commitment to justice \cite{abdAlBarr1994}, reflecting theological concerns about LLMs' inability to possess God-consciousness (\textit{taqw\=a}), sincere intention (\textit{niyyah}), or authentic understanding of individual circumstances, all essential for Islamic guidance.

Scholars permitting limited LLM use emphasize specific safeguards that inform our recommendations. Islamic AI systems must base themselves on authentic sources (Qur'an, Prophetic tradition (\textit{Sunnah}), scholarly consensus (\textit{ijm\=a'}), analogical reasoning (\textit{qiy\=as})) with strict human oversight \cite{unimel_islamic_ai}, requirements our benchmark helps assess. Saudi Arabia's Fatwa Robot operates exclusively on pre-verified fatwas, automatically connecting users to live scholars when questions exceed its parameters \cite{spa2025robot}, an approach our abstention metrics support. Malaysian scholars recommend AI systems specify their jurisprudential tradition or present multi-sourced answers to avoid doctrinal bias \cite{Mohammed2024AIandFatwas}, guidance that motivated our multi-school evaluation framework. The consensus is clear: AI should supplement scholars by providing research assistance and information retrieval, but must never substitute their judgment, wisdom, or authority. The International Islamic Fiqh Academy emphasizes that AI-generated outputs on religious matters must undergo vetting by qualified Islamic scholars (\textit{`ulam\=a`}) \cite{iifa_ai_guidelines}, a position our findings on hallucination rates and false premise acceptance strongly support.


\section{Dataset Creation}
\label{sec:dataset}
Our dataset creation involved two stages. First, we built a foundational Source Dataset by manually extracting question-answer pairs directly from authoritative texts of Islamic jurisprudence. Second, we developed the Benchmarking Dataset by systematically transforming these source entries into a structured evaluation framework. This framework, organized across three complexity levels, enables a rigorous assessment of an LLM’s Islamic legal reasoning capabilities

\subsection{Source Dataset}
The Source Dataset is the result of an expert-led curation process. We began by surveying 60 foundational books of Islamic jurisprudence, focusing on core legal topics including family law, contracts, torts, criminal law, and finance. From this initial pool, we narrowed our selection to the 38 most relevant texts to create a diverse and historically significant collection.

Our reduction from 60 surveyed works to 38 selected texts followed explicit inclusion and exclusion criteria designed to balance doctrinal authority, domain coverage, temporal breadth, and evaluation integrity. We included works that meet four criteria: (1) \textbf{authoritative standing within a school}, reflected in their canonical use in uṣūl and fiqh instruction and their frequent citation in later juristic writing; (2) \textbf{coverage of core substantive domains} targeted by this benchmark (contracts and commercial transactions, family law, property, torts, criminal law, and finance), ensuring that no single domain drives the source pool; (3) \textbf{temporal representativeness} across the classical, post-classical, and modern periods to capture doctrinal continuity and change over roughly 1,200 years; and (4) \textbf{genre diversity}, combining concise manuals (mutūn), extended commentaries (shurūḥ), juristic compendia, fatwā collections, and codificatory texts, so that questions probe both rule statements and explanatory reasoning. We excluded works that are (a) narrowly limited to ritual practice only, (b) largely duplicative of another selected text within the same school and domain, or (c) so widely circulated in modern web corpora that they posed a higher contamination risk, unless their doctrinal role was indispensable for cross-school comparability. This selection approach aimed to preserve breadth while keeping the benchmark sensitive to school-specific doctrine and method.

\textbf{Qualitative bias audit:} The resulting distribution across schools partly reflects structural constraints in digitization and publication availability. Larger Sunni schools have broader digitized textual traditions, while some Shi`i and Zahiri works are less available in machine-readable form, which limits the feasible pool. Within these constraints, we adopted a minimum-coverage principle for each school, ensuring representation of all seven traditions and multiple legal domains per tradition, rather than allowing selection to track availability alone. We therefore treat the current school distribution as a pragmatic approximation of the accessible historical record, not as a claim about the relative doctrinal importance of any school. Future expansions will prioritize additional Shi`i and minority-school sources as more texts become digitally accessible.

This final set of 38 texts, spanning over 1,200 years of legal tradition (8th century to contemporary period), yielded 271 expert-extracted question-answer pairs. The collection reflects both historical significance and methodological diversity across seven schools of jurisprudence: Hanafi (10 works, 26\%), Maliki (8 works, 21\%), Hanbali (6 works, 16\%), Shafi’i (5 works, 13\%), Twelver Shi’i (4 works, 11\%), Zaydi (3 works, 8\%), and Zahiri (2 works, 5\%) . This distribution is intended to reflect the relative availability of digitized works and the extensive textual traditions of the larger Sunni schools; by comparison, legal sources for schools like the Zahiri are significantly rarer. Table \ref{tab:foundations_overview} contains the number of Islamic authoritative books belonging to the 7 schools of thought used in the construction of our source dataset.

\begin{table}[h]
\centering
\renewcommand{\arraystretch}{1.2}
\caption{Overview of Foundational Sources by School of Law}
\label{tab:foundations_overview}
\begin{tabular}{l c l}
\hline
\textbf{School} & \textbf{\# Books} & \textbf{Key Examples (Title - Author)} \\ \hline

Hanafi & 9 &
\begin{tabular}[c]{@{}l@{}}Mukhtasar al-Quduri - \textit{Quduri};\\
al-Mabsut - \textit{Sarakhsi};\\
al-Fatawa al-Hindiyyah (al-`Alamgiriyyah) - \textit{Nizam}\end{tabular} \\ \hline

Maliki & 8 &
\begin{tabular}[c]{@{}l@{}}Al-Muwatta' - \textit{Malik};\\
Mukhtaṣar Khalil - \textit{Khalil};\\
Al-Qawānīn al-Fiqhiyyah - \textit{Ibn Juzayy}\end{tabular} \\ \hline

Shafi`i & 5 &
\begin{tabular}[c]{@{}l@{}}Al-Umm - \textit{Shafi`i};\\
Al-d{H}āwī al-Kabīr - \textit{Mawardi};\\
Fatāwā - \textit{Ghazali}\end{tabular} \\ \hline

Hanbali & 6 &
\begin{tabular}[c]{@{}l@{}}Al-Mughnī - \textit{Ibn Qudāmah};\\
Al-Inṣāf fī Ma`rifat al-Rājiḥ - \textit{Mardawi};\\
Mukhtaṣar al-Khiraqī - \textit{Khiraqi}\end{tabular} \\ \hline

Zahiri & 2 &
\begin{tabular}[c]{@{}l@{}}Al-Muḥallā bi-al-Āthār - \textit{Ibn Hazm};\\
Al-Iḥkām fī Uṣūl al-Aḥkām - \textit{Ibn Hazm}\end{tabular} \\ \hline

Zaydi & 3 &
\begin{tabular}[c]{@{}l@{}}Al-Azhār fī Fiqh al-A'immah al-Aṭhār - \textit{al-Murtaḍā};\\
Sharḥ al-Tajrīd - \textit{al-Murtaḍā};\\
Sharḥ al-Tajrīd - \textit{Harūnī}\end{tabular} \\ \hline

Twelver (Ja`fari) & 4 &
\begin{tabular}[c]{@{}l@{}}Al-Kāfī - \textit{Kulaynī};\\
Tadhkirat al-Fuqahā' - \textit{al-d{H}illī};\\
Taḥrīr al-Wasīlah - \textit{Khomeini}\end{tabular} \\ \hline

\end{tabular}
\end{table}
\noindent The complete list of all 38 texts with full author names, titles, and historical periods is provided in Appendix~\ref{sec:all-books-table}.

The works represent diverse legal genres, including systematic manuals (mutūn), detailed commentaries (shurūḥ), collections of legal edicts (fatāwā), and legal codes. Each of the 271 entries includes comprehensive metadata: source text identification (author and book title), jurisprudential school, historical period, textual references, and complexity assessments.

\subsubsection{Data Collection Process}

A key consideration during selection was mitigating data contamination. This problem occurs when a model is evaluated on texts it has already ``memorized" from its training data, leading to artificially inflated performance scores that do not reflect genuine reasoning . 
To minimize this risk, we made a deliberate effort to select authentic legal texts that are less commonly discussed in modern web corpora or are not widely available in easily machine-readable formats. While no selection process can perfectly guarantee non-exposure, this approach helps ensure our benchmark is genuinely testing the model’s ability to reason about Islamic legal principles rather than its capacity to recall memorized text.

\subsubsection{Dataset Creation Workflow}
\begin{figure}[h]
    \centering
    \includegraphics[width=1\linewidth]{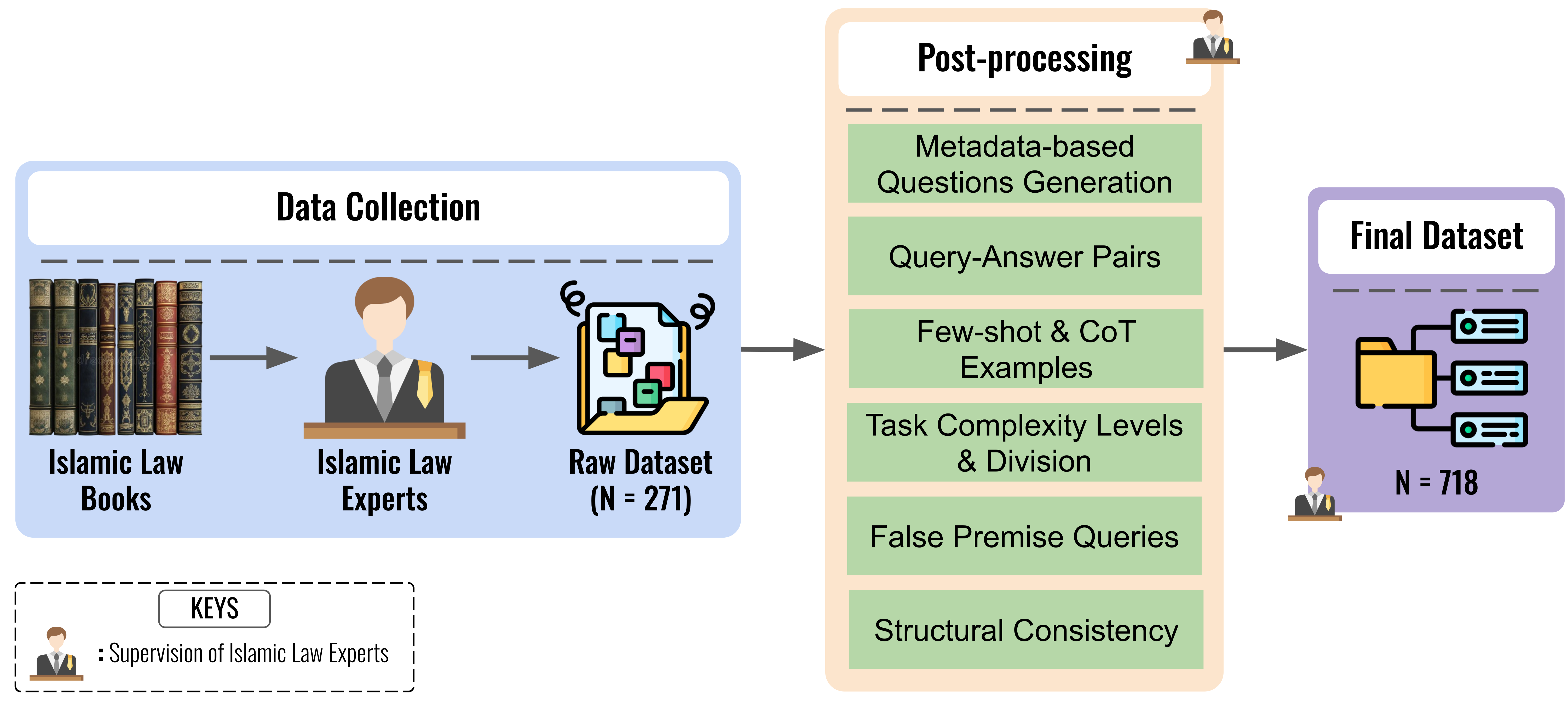}
    \caption{Comprehensive workflow of dataset creation from source Islamic legal texts to final benchmarking dataset, showing the transformation from 271 expert-curated source entries to 718 structured evaluation items under continued expert supervision.}
    \label{fig:dataset_workflow}
\end{figure}

Figure~\ref{fig:dataset_workflow} illustrates our systematic two-stage process. In the \textbf{initial data collection stage}, an Islamic law expert (associate professor at College of Law with expertise in Islamic Law and Jurisprudence) carefully read through all 38 foundational books to extract and formulate question-answer pairs, producing the source dataset of 271 curated entries. This manual extraction ensures authenticity and accuracy in representing Islamic legal principles across different schools and historical periods. Since this stage involves direct extraction from authoritative sources, inter-annotator agreement was not required.

The \textbf{post-processing stage} transforms these 271 source entries into 749 evaluation items under the continued supervision of the same Islamic law expert. Out of these 749 items, 31 were held out for few-shot prompting (selecting representative examples across all task types and complexity levels), resulting in a final dataset of 718 evaluation items. This transformation involves: (1) generating metadata-based questions to test bibliographical knowledge; (2) deriving multiple task variants from single source entries; (3) creating few-shot examples for model reasoning guidance; (4) organizing tasks into three complexity levels based on expert judgment of required jurisprudential sophistication; (5) introducing false premise queries to test whether LLMs identify and correct wrong assumptions; and (6) ensuring structural consistency across all evaluation items.

\textbf{Complexity Level Assignment:} The classification of tasks into Low, Moderate, and High complexity tiers was determined through expert assessment based on the depth of jurisprudential knowledge and reasoning required. Low-complexity tasks involve direct recall of bibliographical facts or basic legal principles. Moderate-complexity tasks require synthesis across multiple conditions, comparative analysis between rulings, or enumeration of structured legal requirements. High-complexity tasks demand sophisticated reasoning, including analogical application of legal principles, cross-school synthesis, identification of underlying legal rationales, or mapping to abstract legal maxims. While this classification involves expert judgment (making it inherently subjective), it is grounded in established frameworks of Islamic legal pedagogy where students progress from memorization (hifz) to comprehension (fahm) to application (ta\d{t}b\=iq) and finally to synthesis and derivation (istinb\=a\d{t}).

\subsection{Benchmarking Dataset}
The benchmarking dataset serves as the primary evaluation framework for systematically assessing LLM performance in Islamic legal reasoning. This comprehensive benchmark comprises 718 structured evaluation items that rigorously test model capabilities across multiple dimensions of jurisprudential understanding. Unlike traditional question-answering datasets, our benchmark is specifically designed to probe both surface level knowledge retrieval and deep legal reasoning abilities required for authentic Islamic jurisprudential analysis. We organize evaluation tasks into three complexity levels (see Figure \ref{fig:taxonomy} for visual taxonomy), each containing distinct task families that progressively challenge different cognitive capabilities:

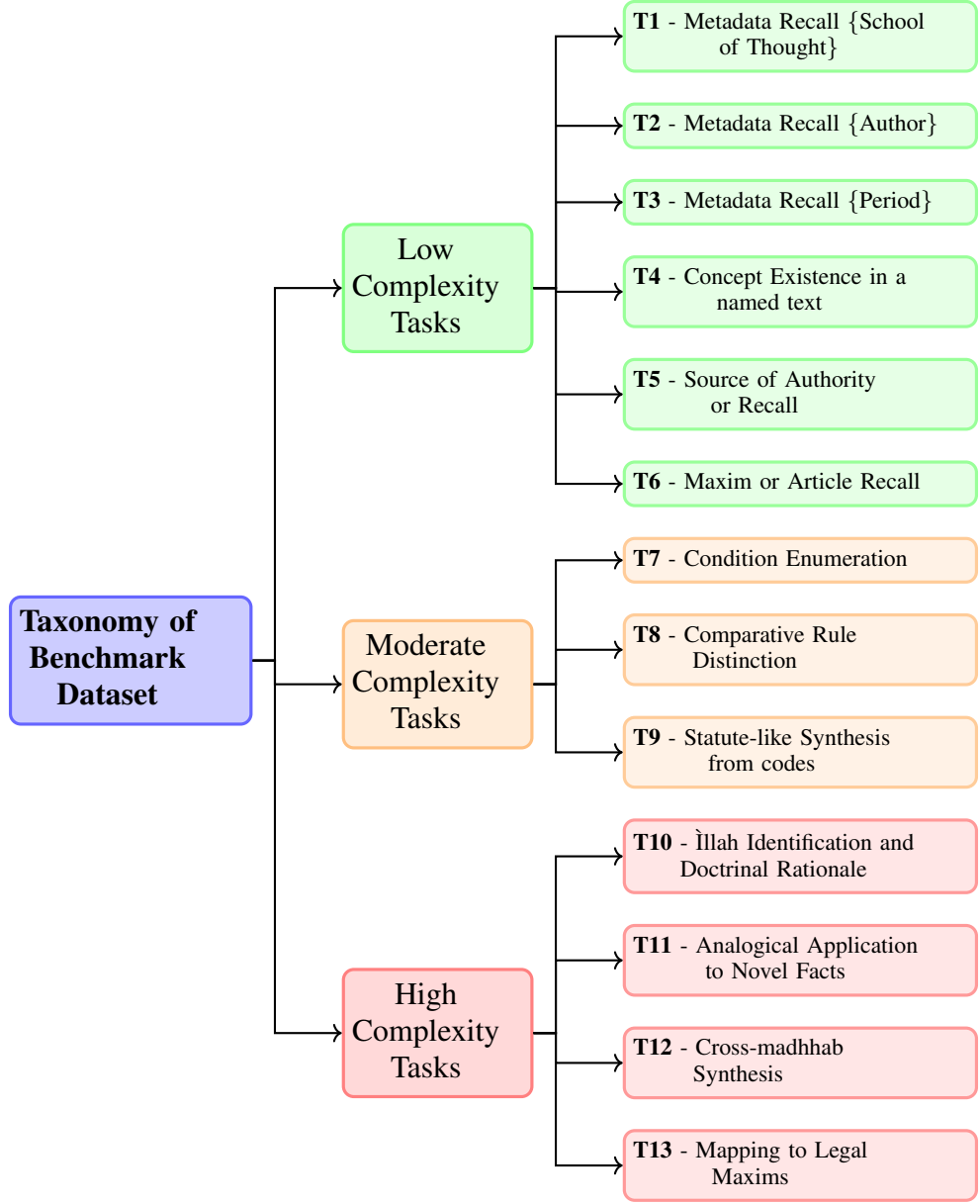
\begin{figure}[h]
\centering
\begin{forest}
  for tree={
    forked edges,
    grow=east,
    rounded corners,
    text width=5cm,
    align=center,
    font=\small,
    edge={thick, ->},
    l sep=1.2cm,
    s sep=4mm,
    fork sep=3mm,
    line width=1.2pt,
    where level=0{
      text width=3cm,
      fill=blue!20,
      draw=blue!60,
      font=\large\bfseries
    }{},
    where level=1{
      text width=2.3cm,
      font=\large,
      align=center
    }{},
    where level=2{
      text width=4.5cm,
      font=\small
    }{},
  }
  [Taxonomy of\\Benchmark\\Dataset
    [High\\Complexity\\Tasks,
      fill=red!15,
      draw=red!50,
      [\textbf{T13} - Mapping to Legal\\Maxims,
        fill=red!10,
        draw=red!40
      ]
      [\textbf{T12} - Cross-madhhab\\Synthesis,
        fill=red!10,
        draw=red!40
      ]
      [\textbf{T11} - Analogical Application\\to Novel Facts,
        fill=red!10,
        draw=red!40
      ]
      [\textbf{T10} - \`Illah Identification and\\Doctrinal Rationale,
        fill=red!10,
        draw=red!40
      ]
    ]
    [Moderate\\Complexity\\Tasks,
      fill=orange!15,
      draw=orange!50,
      [\textbf{T9} - Statute-like Synthesis\\from codes,
        fill=orange!10,
        draw=orange!40
      ]
      [\textbf{T8} - Comparative Rule\\Distinction,
        fill=orange!10,
        draw=orange!40
      ]
      [\textbf{T7} - Condition Enumeration,
        fill=orange!10,
        draw=orange!40
      ]
    ]
    [Low\\Complexity\\Tasks,
      fill=green!15,
      draw=green!50,
      [\textbf{T6} - Maxim or Article Recall,
        fill=green!10,
        draw=green!40
      ]
      [\textbf{T5} - Source of Authority\\or Recall,
        fill=green!10,
        draw=green!40
      ]
      [\textbf{T4} - Concept Existence in a\\named text,
        fill=green!10,
        draw=green!40
      ]
      [\textbf{T3} - Metadata Recall \{Period\},
        fill=green!10,
        draw=green!40
      ]
      [\textbf{T2} - Metadata Recall \{Author\},
        fill=green!10,
        draw=green!40
      ]
      [\textbf{T1} - Metadata Recall \{School\\of Thought\},
        fill=green!10,
        draw=green!40
      ]
    ]
  ]
\end{forest}
\caption{Taxonomy of benchmark dataset tasks (T1 till T13) organized by complexity levels.}
\label{fig:taxonomy}
\end{figure}

\begin{itemize}
    \item \textit{Low-Complexity}: 
    Questions assess basic factual recall and concept existence within a text, testing the model's core grounding.
    
    \item \textit{Moderate-Complexity}: 
    Questions require the precise enumeration of legal conditions or the synthesis of specific rules, testing for accuracy and the ability to avoid extraneous information.
    
    \item \textit{High-Complexity}: 
    High-Complexity questions probe the model's capacity to analyze and articulate the underlying jurisprudential reasoning (`illah), compare legal principles, and apply them to novel scenarios. 
\end{itemize}

Our finalized benchmarking dataset comprises 718 evaluation items distributed across 13 task types spanning three complexity levels. Low-complexity tasks (T1-T6) constitute the majority with 509 instances (71\%), moderate-complexity tasks (T7-T9) account for 112 instances (15\%), and high-complexity tasks (T10-T13) include 97 instances (14\%). This distribution reflects our dataset creation process: while multiple low-complexity tasks (like T1-T4 metadata and concept recall) could be derived from the metadata of a single source entry, the moderate and high-complexity tasks required unique, manual formulation by Islamic law experts for each specific legal scenario. Table~\ref{tab:taskfamilies} presents each task family with representative query examples and the number of instances (N) in our dataset. Notably, concept existence verification (T4) represents the largest single category with 267 items, reflecting the fundamental importance of grounding verification in Islamic legal texts.

In addition to the primary benchmark tasks, we designed a small set of \textbf{\textit{false premise variants}}. These queries deliberately introduce incorrect assumptions (e.g., misattributed authorship, wrong historical period, or fabricated conditions) in the queries with respect to the context. The purpose of these false premise queries is to test whether models have the ability to reject the false premise queries (FPQs), thus showing the knowledge and reasoning capability that they possess, instead of confidently accepting the false belief and then fabricating answers. These FPQs are designed in a way to test whether pre-trained models use their internal knowledge to rebut these falsely plausible questions and correct the wrongly stated question \cite{hu2023wontfooledagainanswering}. We mark these tasks with a prime symbol (T1', T2', T3', T7'). Detailed explanations of the false premises in these examples are provided in Section~\ref{subsec:falsepremisevariants}. 

\renewcommand{\arraystretch}{1.3}
\begin{longtable}{|>{\raggedright\arraybackslash}p{2.0cm}|
                      >{\raggedright\arraybackslash}p{0.8cm}|
                      >{\raggedright\arraybackslash}p{0.5cm}|
                      >{\raggedright\arraybackslash}p{3.5cm}|
                      >{\raggedright\arraybackslash}p{7.5cm}|}
\caption{Task families, query examples, and numbering across three complexity levels.} \label{tab:taskfamilies} \\
\hline
\textbf{Complexity} & \textbf{ID} & \textbf{N} & \textbf{Task family} & \textbf{Query template examples} \\
\hline
\endfirsthead

\multicolumn{5}{c}%
{{\bfseries \tablename\ \thetable{} -- continued from previous page}} \\
\hline
\textbf{Complexity} & \textbf{ID} & \textbf{N} & \textbf{Task family} & \textbf{Query template examples} \\
\hline
\endhead

\hline \multicolumn{5}{|r|}{{Continued on next page}} \\ \hline
\endfoot

\hline
\endlastfoot

Low & T1 & 34 & Metadata recall (School) & ``What school of thought is associated with Mukhtasar al-Qudūrī?'' \\
\hline
Low & T2 & 33 & Metadata recall (Author) & ``Who is the author of Mukhtasar al-Qudūrī?'' \\
\hline
Low & T3 & 34 & Metadata recall (Period) & ``What historical period is Mukhtasar al-Qudūrī from? State the century.'' \\
\hline
Low & T4 & 267 & Concept existence in a named text & ``Does Mukhtasar al-Qudūrī by Abū al-\d{H}asan Aḥmad ibn Muḥammad al-Qudūrī treat Khiyaar al-`ayb? Answer Yes/No, cite page if Yes.'' \\
\hline
Low & T5 & 83 & Source of authority/recall & ``Does the author mention any other jurist's opinion regarding the permissibility of manufacturing contracts (istisnā')? Answer Yes/No, and if yes, who?'' \\
\hline
Low & T6 & 13 & Maxim or article recall & ``State Majalla Article 167's contract-formation rule in one line.'' \\
\hline
Moderate & T7 & 28 & Condition enumeration (shurūṭ) & ``List the conditions for a valid salām as held by Abū d{H}anīfa in the cited page.'' \\
\hline
Moderate & T8 & 21 & Comparative rule distinction & ``Do conditions differ for gold-for-gold vs gold-for-silver? Give one-sentence reason.'' \\
\hline
Moderate & T9 & 57 & Statute-like synthesis from codes & ``Synthesize Article 182's effect when off-topic speech intervenes between ijāb and qabūl.'' \\
\hline
High & T10 & 63 & `Illah identification and doctrinal rationale & ``Explain the `illah for invalidating `one of two garments' in al-Mabsūṭ and connect to munāza`ah.'' \\
\hline
High & T11 & 12 & Analogical application to novel facts & ``Apply mā laysa fī milkih to selling a car intended to be bought next week.'' \\
\hline
High & T12 & 5 & Cross-madhhab synthesis & ``Contrast Abū d{H}anīfa vs Abū Yūsuf on re-delegation in muḍāraba and justify the mashhūr rule.'' \\
\hline
High & T13 & 17 & Mapping to legal maxims & ``Given bay` bi-al-wafā', explain why the Majalla treats intent over form and identify the operative maxim.'' \\
\hline
\end{longtable}
\renewcommand{\arraystretch}{1.0}

\renewcommand{\arraystretch}{1.3}
\begin{longtable}{|>{\raggedright\arraybackslash}p{2.0cm}|
                      >{\raggedright\arraybackslash}p{0.8cm}|
                      >{\raggedright\arraybackslash}p{0.5cm}|
                      >{\raggedright\arraybackslash}p{3.5cm}|
                      >{\raggedright\arraybackslash}p{7.5cm}|}
\caption{False premise variant tasks.} \label{tab:taskfamilies-falsepremise} \\
\hline
\textbf{Complexity} & \textbf{ID} & \textbf{N} & \textbf{Task family} & \textbf{Query template examples} \\
\hline
\endfirsthead

\multicolumn{5}{c}%
{{\bfseries \tablename\ \thetable{} -- continued from previous page}} \\
\hline
\textbf{Complexity} & \textbf{ID} & \textbf{N} & \textbf{Task family} & \textbf{Query template examples} \\
\hline
\endhead

\hline \multicolumn{5}{|r|}{{Continued on next page}} \\ \hline
\endfoot

\hline
\endlastfoot

Low & T1' & 15 & False premise detection (School) & ``Analyzing the major Maliki work 'Badā'i` al-\d{S}anā'i` fī Tartīb al-Sharā'i`', what distinctive systematization methods does it employ that exemplify Maliki approaches to organizing legal principles?'' \\
\hline
Low & T2' & 15 & False premise detection (Author) & ``How did al-`Allāmah al-d{H}illī organize the hadith material in his work 'Al-Kāfī'?'' \\
\hline
Low & T3' & 15 & False premise detection (Period) & ``As `Alā' al-Dīn al-Kāsānī wrote 'Kitāb al-Mabsūṭ' in the 20th-century CE Modern period, how does it address contemporary Islamic legal challenges?'' \\
\hline
Moderate & T7' & 6 & False premise detection (Condition enumeration) & ``According to the ruling on page 145, a marriage contract is formed by offer (ījāb) and acceptance (qabūl) expressed in terms that signify marriage, and it is not valid unless it is witnessed by four free, adult, sane, and Muslim men. What happens if only three witnesses are present?'' \\

\hline
\end{longtable}
\renewcommand{\arraystretch}{1.0}

\subsubsection{Construction and Evaluation Methods}
Benchmark entries were systematically derived from source questions through template-based instantiation and editorial rewriting to reduce ambiguity. Each task underwent quality review for clarity, citation accuracy (where applicable), and consistency of complexity labels to ensure alignment with the intended cognitive demands. Recent evaluations of LLMs in legal-adjacent domains show that models achieve high accuracy on structured formats such as true/false and multiple-choice questions. However, their performance drops considerably on open-ended questions that require applying rules and reasoning through a scenario, indicating continuing limitations in complex legal task execution \cite{GoganiKhiabani2025}. Hence, our benchmark also focuses on single-turn prompt–response questions that demand precise legal interpretation and grounded justification, offering a realistic assessment of LLM capabilities on Islamic legal reasoning.

We used reference-based evaluation across all task types: model outputs were scored by an LLM-as-Judge using an Exact Semantic Meaning (ESM) rubric that directly compares outputs to ground-truth (gold) reference answers. The ESM criterion isolates the core elements required by the references while permitting variation in writing style, formatting, and phrasing, thereby enabling objective scoring via direct comparison with authoritative references. For high-complexity tasks requiring reasoning, we evaluated only the final responses; intermediate reasoning was not scored because the dataset provides final answers but no reference chains of thought. Any reasoning traces produced by models were supplied to the LLM-as-Judge as context, not as inputs to the scoring criteria.



\section{Setup}
\label{sec:setup}
This section describes our experimental setup, including the evaluated LLMs, prompting strategies (zero-shot and few-shot in-context learning), and abstention handling protocols. We provide comprehensive implementation details to ensure reproducibility and transparency.

\subsection{LLMs}
LLMs are widely accessible and increasingly used for everyday question answering, including by Muslim users seeking religious guidance. Although these public platforms are not specialized for the nuances of Islamic legal reasoning, their ubiquity makes rigorous evaluation essential. Because LLMs are deployed through widely used interfaces (e.g., ChatGPT, Gemini, Grok) and accessed by both lay users and developers via APIs to build domain-specific applications, any shortcomings can propagate at scale. Benchmarking frontier models is therefore a prerequisite for safe, reliable, and accountable deployment. Although some specialized legal models can do better than general models on specific legal tasks \cite{Guo2025}, these systems are still uncommon or niche-specific and not used by the general public. For this reason, for the scope of this study, we focus on mainstream general-purpose LLMs. These are the models that set the overall performance standard, power the most popular platforms, and are the ones people, including Muslims, usually rely on in daily life for guidance on various matters. Evaluating these widely used systems is therefore more important for understanding real-world risks.

To ensure coverage of the state of the art, we include the latest closed-source commercial models and access them directly through their official APIs (e.g., OpenAI, Google, Anthropic, and xAI). Alongside these, we also benchmark open-source frontier models deployed via Together AI \cite{togetherai2025}, which provides scalable access to open-source models. Table~\ref{tab:llm_overview} provides an overview of the evaluated off-the-shelf LLMs, listing their companies, model names, source types, providers, and model sizes where publicly available.

\begin{table}[h]
\centering
\caption{Overview of Evaluated Off-the-Shelf LLMs}
\label{tab:llm_overview}
\renewcommand{\arraystretch}{1.2}
\resizebox{\textwidth}{!}{
\begin{tabular}{|l|l|p{3.2cm}|l|l|l|}
\hline
\textbf{Company} & \textbf{Model Name} & \textbf{Model API Version} & \textbf{Type} & \textbf{API Provider} & \textbf{Model Size} \\ \hline
Anthropic & Claude Sonnet 4.5 \cite{anthropic2025claude45} & claude-sonnet-4-5-20250929 & Closed Source & Anthropic & Unknown \\ \hline
Google & Gemini 2.5 Pro \cite{deepmind_gemini2.5pro_2025} & gemini-2.5-pro & Closed Source & Google & Unknown \\ \hline
OpenAI & GPT-5 \cite{openai2025introducing-gpt5} & gpt-5 & Closed Source & OpenAI & Unknown \\ \hline
xAI & Grok-4 \cite{xai2025grok4} & grok-4-0709 & Closed Source & xAI & Unknown \\ \hline
Meta & Llama 4 Maverick \cite{meta2025llama4} & meta-llama/Llama-4-Maverick-17B-128E-Instruct-FP8 & Open Source & Together AI & 400B \small{(17B Active)} \\ \hline
Meta & Llama 3.1 \cite{meta2024llama31} & meta-llama/Meta-Llama-3.1-8B-Instruct-Turbo & Open Source & Together AI & 8B \\ \hline
OpenAI & GPT-OSS-120B \cite{openai2025introducing-gpt-oss} & openai/gpt-oss-120b & Open Source & Together AI & 120B \\ \hline
DeepSeek & DeepSeek R1 \cite{deepseek2025r10528} & deepseek-ai/DeepSeek-R1 & Open Source & Together AI & 671B \\ \hline
Qwen & Qwen3 \cite{qwen2025qwen3} & Qwen/Qwen3-235B-A22B-Instruct-2507-tput & Open Source & Together AI & 235B \\ \hline
\end{tabular}
}
\end{table}

\subsection{Prompting Techniques}
 
\subsubsection{In-Context Sampling} 
We employed in-context prompting techniques to evaluate the performance of LLMs in our Islamic legal reasoning benchmarking. To systematically assess model capabilities across different prompting paradigms, we evaluate all tasks across all complexity levels using both \emph{zero-shot} and \emph{few-shot prompting}. This approach allows us to measure how effectively LLMs can perform Islamic legal reasoning with and without the influence of contextual examples, providing insights into their inherent understanding versus their ability to learn from in-context demonstrations.

In \textit{zero-shot prompting}, models respond without illustrative examples, relying solely on their pre-trained knowledge and the task instructions. This setting evaluates a model's unassisted capacity to interpret and answer Islamic legal queries across all complexity levels. 

In \textit{few-shot prompting}, we provide exemplars that specify the expected output format and reasoning approach for the task type. These contextual examples help models internalize the task structure and produce more aligned responses. However, few-shot prompting also shapes the model's internal decision process; outputs can be highly sensitive to the provided exemplars, introducing example-induced biases. We discuss the benefits and drawbacks of few-shot prompting, especially for Islamic legal reasoning and for tasks that require exact source knowledge with no tolerance for error, in the Results (Section~\ref{sec:results}). 

\begin{genericbox}[gray]{Few-Shot Prompting Example - T7}
\small
\textbf{Example 1:}

\textbf{$<$context$>$}\\
Book: ``Jawāhir al-Kalām fī Sharḥ Sharā'i` al-Islām''\\
Author: Sheikh Muḥammad d{H}asan al-Najafī\\
School of Thought: Twelver Shi`i (Ja'fari)\\
Period: Modern (19th Century) (d. 1266 H / 1850 CE)\\
Page Reference: 23/50\\
\textbf{$<$/context$>$}

\textbf{Query:}``What is the ruling on selling weapons to the enemies of the faith (a`dā' al-dīn), and under what condition is it permissible, according to page 50?''

\textbf{Answer:} ``It is forbidden to sell weapons to the enemies of the faith if it leads to assisting them against Muslims. It is permissible if there is a truce (hudnah) between them and the Muslims.''

---

\textbf{Example 2:}
...
---

\textbf{Actual Query:}

\textbf{$<$context$>$}\\
CONTEXT\_HERE\\
\textbf{$<$/context$>$}\\

\textbf{Query:} ``QUERY\_HERE''

\textbf{Answer: }
\end{genericbox}

\subsubsection{Abstention Behaviour}

Similar to false premise questions where LLMs should rebut incorrect assumptions, abstention behavior addresses scenarios where queries fall outside the model's internal or retrieval-augmented knowledge base, or there is a flawed context. In such cases, LLMs must recognize when to answer and when to refuse, a safety mechanism known as \textit{selective refusal}. Recent studies show that even frontier LLMs struggle with this capability, failing to identify correct reasons for refusal in more than 50\% of cases \cite{muhamed2025refusalbench}. This challenge takes on particular urgency in Islamic legal contexts, where incorrect guidance can lead to substantial religious and practical consequences. As discussed in Section~\ref{subsec:muslimsscholars}, Muslim scholars have extensively cautioned against unvetted AI systems issuing religious rulings, emphasizing that such systems lack the contextual awareness, moral integrity, and human judgment essential for Islamic jurisprudence. These scholarly perspectives underscore a critical implication for LLM evaluation: abstention in cases of uncertainty should be recognized as responsible behavior rather than system failure.

These scholarly insights underscore that abstention must be recognized as responsible behavior rather than penalized as failure. Recent work from OpenAI shows that LLMs hallucinate precisely because standard training and evaluation procedures reward guessing over acknowledging uncertainty \cite{kalai2025languagemodelshallucinate}. They propose that while hallucinations should be heavily penalized during benchmarking, honest abstention behaviors such as ``I don't know'' should receive partial credit. The \textit{FiqhQA} benchmark \cite{atif2025sacredsyntheticevaluatingllm} demonstrated this principle empirically. GPT-4o achieved higher accuracy but lower abstention rates, yielding more confident incorrect answers, while Gemini exhibited 90\% abstention with minimal errors (1\% incorrect). This reveals a fundamental trade-off in Islamic legal AI: knowing when \textit{not} to answer can be as critical as answering correctly.

Given these considerations, allowing LLMs to abstain from answering sensitive Islamic legal questions is preferable to providing potentially incorrect guidance \cite{atif2025sacredsyntheticevaluatingllm, kalai2025languagemodelshallucinate}. While excessive abstention is not ideal, prompting models to recognize situations where abstention is appropriate can reduce the risk of harmful misinformation and hallucinations. In our experiments, we include a dedicated evaluation of LLM abstention behavior to assess whether models can appropriately recognize the limits of their knowledge in Islamic legal contexts.


\section{Evaluation Methodology}
\label{sec:evaluation}

Our evaluation employs an approach that focuses on capturing all core elements of the reference answer while allowing for natural variations in expression. This methodology ``\textbf{Exact Semantic Meaning}'' recognizes that in fields like Islamic Law, concepts can be articulated in multiple ways without altering their fundamental meaning or context. This means the formatting, writing style, grammar, etc., all of these things are not considered in our results evaluation, and instead, we solely focused on the core elements that constitute the ground truth answers.

\subsection{Scoring Methods}
\begin{figure}[t]
    \centering
    \includegraphics[width=1\linewidth]{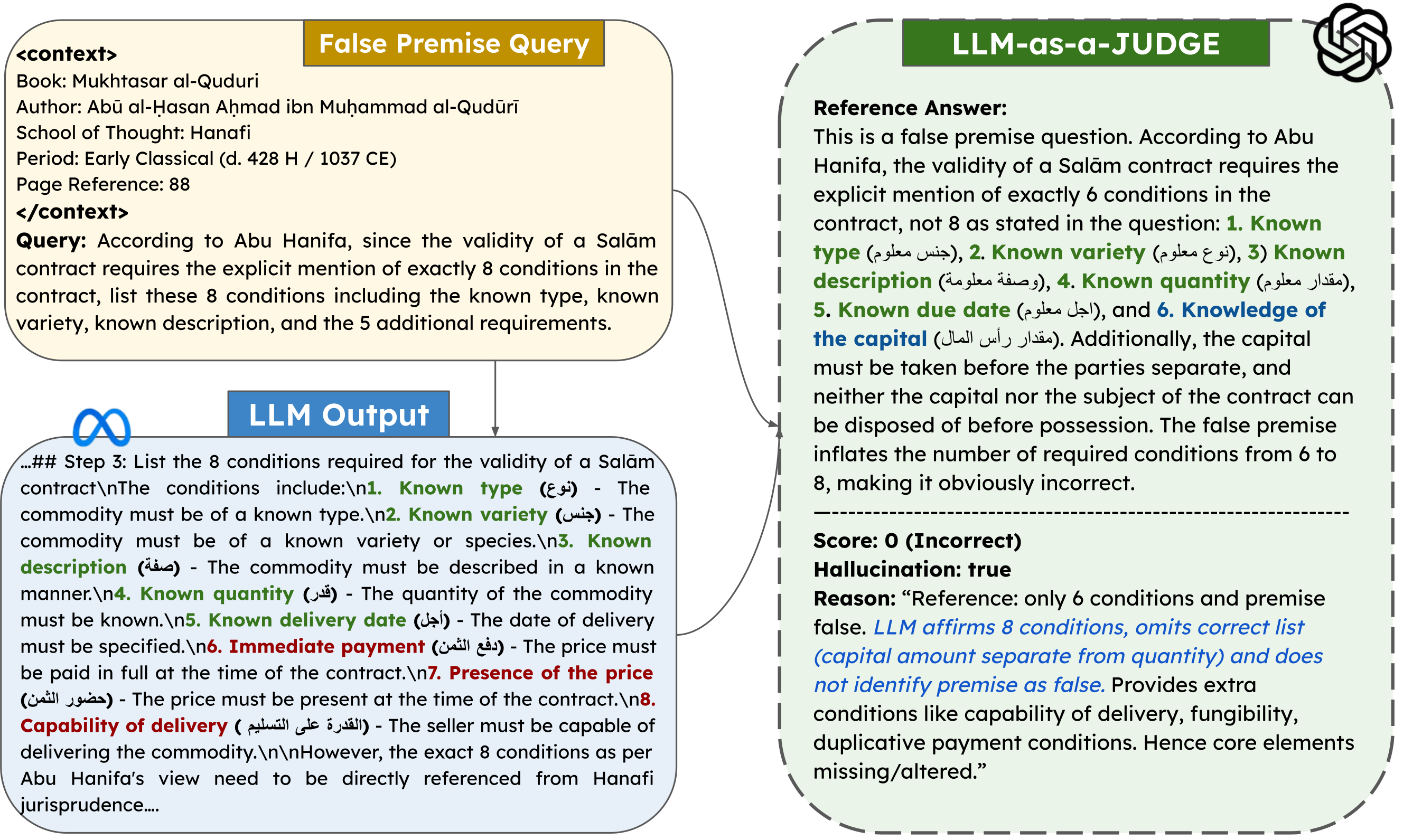}
    \caption{Illustration of a false premise query evaluated through an LLM-as-a-JUDGE framework. The \textbf{LLM Output} (generated by LLaMA-4 Maverick) falsely assumes that Abū d{H}anīfa required eight conditions for a valid \emph{Salām} contract, while the correct jurisprudential reference specifies only six. The output accepts the false premise and hallucinates extra conditions, whereas the judge model correctly identifies the error and assigns a score of 0 (Incorrect). In the `LLM Output' and `Reference Answer' panels, \textcolor{greenc}{\textbf{Green}} highlights correctly mentioned conditions, \textcolor{redc}{\textbf{Red}} indicates fabricated or hallucinated ones, and \textcolor{bluec}{\textbf{Blue}} marks missing but essential elements for required conditions.}
    \label{fig:falsepremise-llama4}
\end{figure}

\textbf{LLM-as-a-Judge Framework:} We employ an LLM-as-a-Judge evaluation framework using o3 \cite{openai2025introducing-o3-o4mini} as the judge model to systematically assess model responses against ground truth answers. For LLM-as-a-Judge, we configured the o3 with the reasoning effort parameter to ``\emph{high}'' which allocates the maximum compute for the model to think and break down the task into multiple steps to generate the most accurate results. At the time of this study, OpenAI does not provide the option of changing the temperature for reasoning and the latest GPT models. 
This approach leverages o3's advanced reasoning capabilities to perform semantic comparison and scoring of LLM-generated outputs, enabling scalable and consistent evaluation across our comprehensive benchmark. The judge model evaluates each response according to our strict scoring criteria (detailed in Section~\ref{subsec:eval_process}), comparing LLM outputs against reference answers to determine correctness and detect hallucinations. In our internal testing on a subset of the benchmark, o3 performed the best out of 5 other LLMs (GPT-5 \cite{openai2025introducing-gpt5}, GPT-5 mini \cite{OpenAI_GPT5Mini_2025}, GPT-4o \cite{OpenAI_Hello_GPT4o_2024}, Gemini 2.5 \cite{google2025gemini25flash}, and GPT-OSS-120B \cite{openai2025introducing-gpt-oss}) when judging the response accurately. \cite{tan2024judgebench} introduced JudgeBench, and in their results, they showed that o3 outperformed all the other frontier LLMs in this task. Figure~\ref{fig:falsepremise-llama4} illustrates an example of this framework applied to a false premise query, demonstrating how the judge model correctly identifies fabricated content and assigns appropriate scores. 

All LLM response generations were performed using 0.0 temperature settings for deterministic output, ensuring reproducibility between experiments. Models were evaluated on identical prompts with standardized formatting to maintain consistency. For tasks requiring contextual information, we provided the standardized template format described in Section~\ref{sec:benchtasks}, ensuring all models received identical background information.

\textbf{Inter-Rater Reliability:} To annotate responses for correctness and hallucination, we engaged Islamic law experts to hand-score a subset of model outputs using the rubric defined in Section~\ref{subsec:scoring_hallucination_definition}. Efficient evaluation of AI-generated text remains an open problem, with inherent trade-offs among internal validity, external validity, replicability, and speed \cite{hashimoto-etal-2019-unifying, liu-etal-2016-evaluate, smith-etal-2022-human}. These trade-offs are especially salient when queries approximate real-world tasks. Because authoritative sourcing is central to legal research and writing, adherence to authority must also be assessed manually to ensure that automated self-checking does not become a confounding driver of performance. Building on prior work in legal LLM evaluation \cite{Dahl_first, Dahl_second, atif2025sacredsyntheticevaluatingllm}, we operationalized precise definitions of correctness and hallucination and, on that basis, specified our scoring methodology in Section~\ref{subsec:scoring_hallucination_definition}. This approach renders benchmarking both costly (API usage and expert annotation) and time-consuming, a necessary trade-off that should inform future discussions about responsibly integrating AI into legal workflows. The definition of key terms, such as hallucination, remains an active area of debate and requires domain-sensitive adaptation. Prior work in legal reasoning \cite{Dahl_first, Dahl_second} provides carefully delineated distinctions that we adopt with modest modifications. In our setting, prompts include a context block (e.g., book title, author, school of thought), supplying authoritative source metadata which excludes text-groundedness as a scoring criterion. Section~\ref{subsec:scoring_hallucination_definition} details the exact definitions employed in this study.

To verify that our queries and coding schema are sufficiently precise, we measured inter-rater reliability. All task responses were first scored by an LLM-as-a-Judge. We then drew a stratified sample (n = 140) by model, task type, and responses per task from the Moderate and High Complexity tasks (T7–T13), the most technical and reasoning-intensive, and submitted them to two independent Islamic law experts for grading. The experts were blinded to the LLM-as-a-Judge scores and to each other's annotations, and they did not communicate. Their only guidance was our written labeling manual, which we also provided to the LLM-as-a-Judge. Both experts had a good command of the legal reasoning in Islam and knew both English and Arabic languages which was essential in understanding the key terms in the dataset.

Under this protocol, Cohen's kappa ($\kappa$) \cite{cohen1960coefficient} between the two experts was \emph{$\kappa = 0.73$} for Scores (refusal: -1 / incorrect: 0 /partially correct: 1 / correct: 2) and \emph{$\kappa = 0.72$} for Hallucination, and an overall agreement on the labels across the human evaluators and the LLM-as-a-Judge was \emph{$\kappa = 0.816$}. These levels of agreement (81.6\%) suggest that the tasks and label taxonomy are well specified, and annotators had a strong agreement. The Section~\ref{sec:inter_rater_appendix} in the Appendix provides more detailed and specific values regarding reliability testing.

The evaluation prioritizes core element preservation for all essential Islamic legal elements. We award 100\% accuracy (score of 2) if all core elements are present and accurately conveyed, even with minor irrelevant mistakes (e.g., a tangential non-core detail that does not impact the core elements). Any core element that is missing, altered in meaning, or misrepresented in semantics/context results in penalization.

\subsection{Scoring and Hallucination Detection} \label{subsec:scoring_hallucination_definition}
We evaluate responses solely on whether they capture all core elements of the ground-truth/reference answer, disregarding variation in grammar, style, paraphrase, or formatting. Core elements are the essential factual components, principles, and authoritative references from Islamic law that must be present and accurately conveyed without any change in meaning, semantics, or context. The evaluation prioritizes preservation of these elements. A response receives 100\% accuracy if all core elements are present and correct, even if it contains minor, non-essential additions (e.g., tangential details that do not affect the core elements). Any omission, semantic alteration, or misrepresentation of a core element results in a penalty as also shown in Figure \ref{fig:falsepremise-llama4}.

\textbf{Definition of Hallucination:} We define hallucination as any LLM-generated response that is partially correct or incorrect with fabricated or invented content that is either not present or completely diverges from the ground-truth answer. This definition is adapted from \cite{Dahl_first, Dahl_second} and is tailored to our setting. In our framework, an Incorrect label typically indicates that the model relied on a fabricated or erroneous assumption at the outset, i.e., a hallucinatory premise, which yields a substantively wrong conclusion. Likewise, a Partially Correct response that includes any invented or misattributed detail is also treated as a hallucination. This includes material that deviates from established Islamic legal sources, adds non-existent information, is fabricated, or misstates jurisprudential principles and authorities. We treat the ground-truth answers as the sole basis for identifying hallucinations, as they specify the decisive elements and underlying rationale. As noted above, our scoring method relies on exact semantic meaning, permitting variability in phrasing while requiring alignment on core elements. This eliminates the need for multiple variants of the ground-truth answer.

The assessment employs a strict four-tier scoring system with the following methodology:

\begin{itemize}
\item \textbf{Correct (Score: 2):} Factually accurate, fully relevant to the query and ground truth, with all core elements intact
\item \textbf{Partially Correct (Score: 1):} Contains some correct information but misses or incompletely addresses key core elements
\item \textbf{Incorrect (Score: 0):} Contains any factually inaccurate information compared to ground truth or fails to address the query
\item \textbf{Refusal (Score: -1):} If the model refuses or abstains from answering. This is considered relatively positive behavior since the LLM is refusing to generate information outside of its contextual understanding.
\end{itemize}

We implement a systematic hallucination-detection procedure to identify responses containing invented, non-existent, or misgrounded elements (e.g., fabricated sources or false attributions other than source text) not present in the ground truth. The hallucination flag is determined as follows: if the response is Incorrect (score 0), the hallucination flag is set to \textbf{True}; if the response is Partially Correct (score 1), and it contains invented or misgrounded elements relative to the reference answer set flag to \textbf{True} and \textbf{False} otherwise; responses that are Correct (score 2) are flagged \textbf{False}; Refusal responses (score -1) are not flagged for hallucination, as the model did not generate potentially false or hallucinated content.

\subsection{Evaluation Process}
\label{subsec:eval_process}

We employ a structured seven-step evaluation process to ensure consistent, accurate, and reproducible assessment. 
\begin{enumerate}
    \item \textbf{Comprehensive Reading:} Evaluators carefully read the query, the ground truth/reference answer, and the LLM response in full to capture the complete context.  

    \item \textbf{Core Element Identification:} All essential factual components are extracted from the reference answer, including Islamic legal principles, authoritative references (e.g., book titles, authors, page numbers), and doctrinal positions across different schools of thought.  

    \item \textbf{Semantic Comparison:} The LLM response is then compared against these core elements to verify their presence and semantic accuracy, with attention to meaning and context rather than grammar, paraphrasing, or stylistic variations.  

    \item \textbf{Non-Core Error Handling:} Minor inaccuracies or tangential details that do not affect the integrity of the core elements are noted but disregarded in the scoring process.  

    \item \textbf{Score Assignment:} Scores are assigned according to a strict rubric: full credit (2) if all core elements are intact and accurately represented, partial credit (1) if only some are correctly captured, no credit (0) if core elements are missing, misrepresented, or factually inaccurate, and refusal score (-1) if the model abstains from providing an answer.  

    \item \textbf{Hallucination Detection:} Responses scoring 0 are automatically flagged as hallucinations. For responses scoring 1, evaluators check for fabricated or misgrounded elements and flag hallucination as true if such content is found. Correct responses scoring 2 are never flagged. Refusals with a score of -1 are not flagged either.

    \item \textbf{Standardized Output:} Each evaluation is reported in a uniform format for clarity and reproducibility: \texttt{Score: [0/1/2/-1]. Hallucination: [True/False]. Reason: [concise explanation]}.
\end{enumerate}

This approach is particularly crucial for Islamic legal reasoning, where the precision of religious and jurisprudential content must be maintained while allowing for the natural linguistic diversity inherent in scholarly discourse.


\begin{table*}[h]
\centering
\caption{Performance comparison of LLMs on our Islamic Legal Benchmark across different task complexities and prompting strategies (Pass@1 rates). Values represent correct response percentages. Best results per task are highlighted in \textbf{bold}.}
\label{tab:correct_results}
\resizebox{\textwidth}{!}{%
\begin{tabular}{ll|cccccc|ccc|cccc|c}
\toprule
\multirow{2}{*}{\textbf{Model}} & \multirow{2}{*}{\textbf{Prompt}} & \multicolumn{6}{c|}{\textbf{Low Complexity}} & \multicolumn{3}{c|}{\textbf{Moderate}} & \multicolumn{4}{c|}{\textbf{High Complexity}} & \multirow{2}{*}{\textbf{Overall}} \\
 & & \textbf{T1} & \textbf{T2} & \textbf{T3} & \textbf{T4} & \textbf{T5} & \textbf{T6} & \textbf{T7} & \textbf{T8} & \textbf{T9} & \textbf{T10} & \textbf{T11} & \textbf{T12} & \textbf{T13} & \\
\midrule
\multirow{2}{*}{Claude Sonnet 4.5} & Zero-Shot & 89.80 & 87.50 & 85.71 & 76.03 & 36.14 & \textbf{61.54} & \textbf{55.88} & 33.33 & 45.61 & 77.78 & \textbf{83.33} & 40.00 & 76.47 & 65.32 \\
 & Few-Shot & 89.80 & \textbf{89.58} & \textbf{87.76} & 90.64 & 34.94 & 53.85 & 38.24 & 42.86 & 56.14 & 74.60 & 58.33 & 60.00 & 76.47 & 65.63 \\
\midrule
\multirow{2}{*}{DeepSeek R1} & Zero-Shot & 73.47 & 54.17 & 59.18 & 91.39 & 33.73 & 38.46 & 26.47 & 38.10 & 38.60 & 65.08 & 41.67 & 40.00 & 82.35 & 52.51 \\
 & Few-Shot & 73.47 & 58.33 & 57.14 & 94.76 & 26.51 & 38.46 & 20.59 & 47.62 & 36.84 & 76.19 & 58.33 & 40.00 & 76.47 & 54.21 \\
\midrule
\multirow{2}{*}{Gemini 2.5 Pro} & Zero-Shot & 87.76 & 85.42 & 85.71 & 90.64 & 38.55 & 53.85 & 44.12 & 42.86 & 54.39 & 79.37 & 50.00 & 40.00 & 82.35 & 64.23 \\
 & Few-Shot & 89.80 & 83.33 & \textbf{87.76} & 92.51 & 37.35 & 46.15 & 35.29 & 47.62 & 52.63 & 71.43 & 50.00 & 40.00 & 82.35 & 62.79 \\
\midrule
\multirow{2}{*}{GPT-5} & Zero-Shot & 89.80 & 75.00 & 83.67 & 85.77 & 39.76 & \textbf{61.54} & 44.12 & \textbf{57.14} & \textbf{59.65} & \textbf{82.54} & 75.00 & 20.00 & 76.47 & 65.42 \\
 & Few-Shot & \textbf{91.84} & 83.33 & 81.63 & 93.63 & \textbf{46.99} & \textbf{61.54} & 47.06 & \textbf{57.14} & 57.89 & 77.78 & 50.00 & 60.00 & 70.59 & \textbf{67.65} \\
\midrule
\multirow{2}{*}{Grok-4} & Zero-Shot & 79.59 & 77.08 & 71.43 & 76.78 & 38.55 & \textbf{61.54} & 52.94 & 47.62 & 52.63 & 77.78 & 33.33 & \textbf{80.00} & \textbf{88.24} & 64.42 \\
 & Few-Shot & 81.63 & 77.08 & 77.55 & 91.01 & 34.94 & \textbf{61.54} & 32.35 & 38.10 & 52.63 & 74.60 & 50.00 & 60.00 & 70.59 & 61.69 \\
\midrule
\multirow{2}{*}{Llama 4 Maverick 17B} & Zero-Shot & 71.43 & 41.67 & 53.06 & \textbf{98.50} & 28.92 & 53.85 & 20.59 & 23.81 & 33.33 & 58.73 & 41.67 & 20.00 & 70.59 & 47.40 \\
 & Few-Shot & 75.51 & 54.17 & 46.94 & 97.00 & 22.89 & \textbf{61.54} & 23.53 & 23.81 & 31.58 & 53.97 & 58.33 & 0.00 & 58.82 & 46.78 \\
\midrule
\multirow{2}{*}{Llama 3.1 8B} & Zero-Shot & 30.61 & 14.58 & 26.53 & 79.40 & 20.48 & 23.08 & 2.94 & 14.29 & 15.79 & 28.57 & 33.33 & 0.00 & 58.82 & 26.80 \\
 & Few-Shot & 26.53 & 20.83 & 22.45 & 97.75 & 19.28 & 15.38 & 8.82 & 14.29 & 19.30 & 34.92 & 50.00 & 20.00 & 52.94 & 30.96 \\
\midrule
\multirow{2}{*}{GPT-OSS-120B} & Zero-Shot & 34.69 & 12.50 & 32.65 & 49.81 & 31.33 & 53.85 & 11.76 & 23.81 & 36.84 & 50.79 & 41.67 & 40.00 & 52.94 & 36.36 \\
 & Few-Shot & 32.65 & 16.67 & 34.69 & 79.03 & 22.89 & 53.85 & 11.76 & 19.05 & 35.09 & 47.62 & 25.00 & 0.00 & 47.06 & 32.72 \\
\midrule
\multirow{2}{*}{Qwen3 235B} & Zero-Shot & 61.22 & 43.75 & 42.86 & 87.27 & 31.33 & 46.15 & 23.53 & 33.33 & 45.61 & 66.67 & 41.67 & 20.00 & 76.47 & 47.68 \\
 & Few-Shot & 63.27 & 47.92 & 32.65 & 96.63 & 31.33 & 46.15 & 20.59 & 28.57 & 35.09 & 66.67 & 50.00 & 40.00 & 76.47 & 48.87 \\
\bottomrule
\end{tabular}%
}
\end{table*}

\section{Results}
\label{sec:results}

We evaluate nine state-of-the-art LLMs on our Islamic Legal Benchmark, encompassing both closed-source systems and open-source alternatives. Our analysis focuses primarily on few-shot performance, aligning with standard LLM benchmarking practices. Where instructive, we draw upon zero-shot results to illuminate the (in)effectiveness of in-context learning (ICL) specifically for Islamic jurisprudence reasoning. The following subsections present a comprehensive analysis of model performance across multiple dimensions: correctness, hallucination, error patterns, critical reasoning capabilities, and systematic limitations. \footnote{\textbf{Note on Performance Tiers and Risk Thresholds.} Throughout this analysis, we categorize models into performance tiers and define risk thresholds (e.g., hallucination $>$40\% as "critical," FPQ acceptance $>$50\% as "dangerous"). These are analytical constructs from empirical observation of our evaluation's performance distribution, established to facilitate meaningful interpretation. They do not represent formally validated cutoffs or industry standards, but serve as heuristic boundaries for characterizing relative model capabilities and safety profiles. Different applications may warrant different thresholds based on deployment requirements, risk tolerance, and domain constraints.}

\subsection{Correct Response Performance: The Foundation of Islamic Legal Reasoning}
\label{sec:correct_performance}

\begin{figure}[t]
    \centering
    \includegraphics[width=\columnwidth]{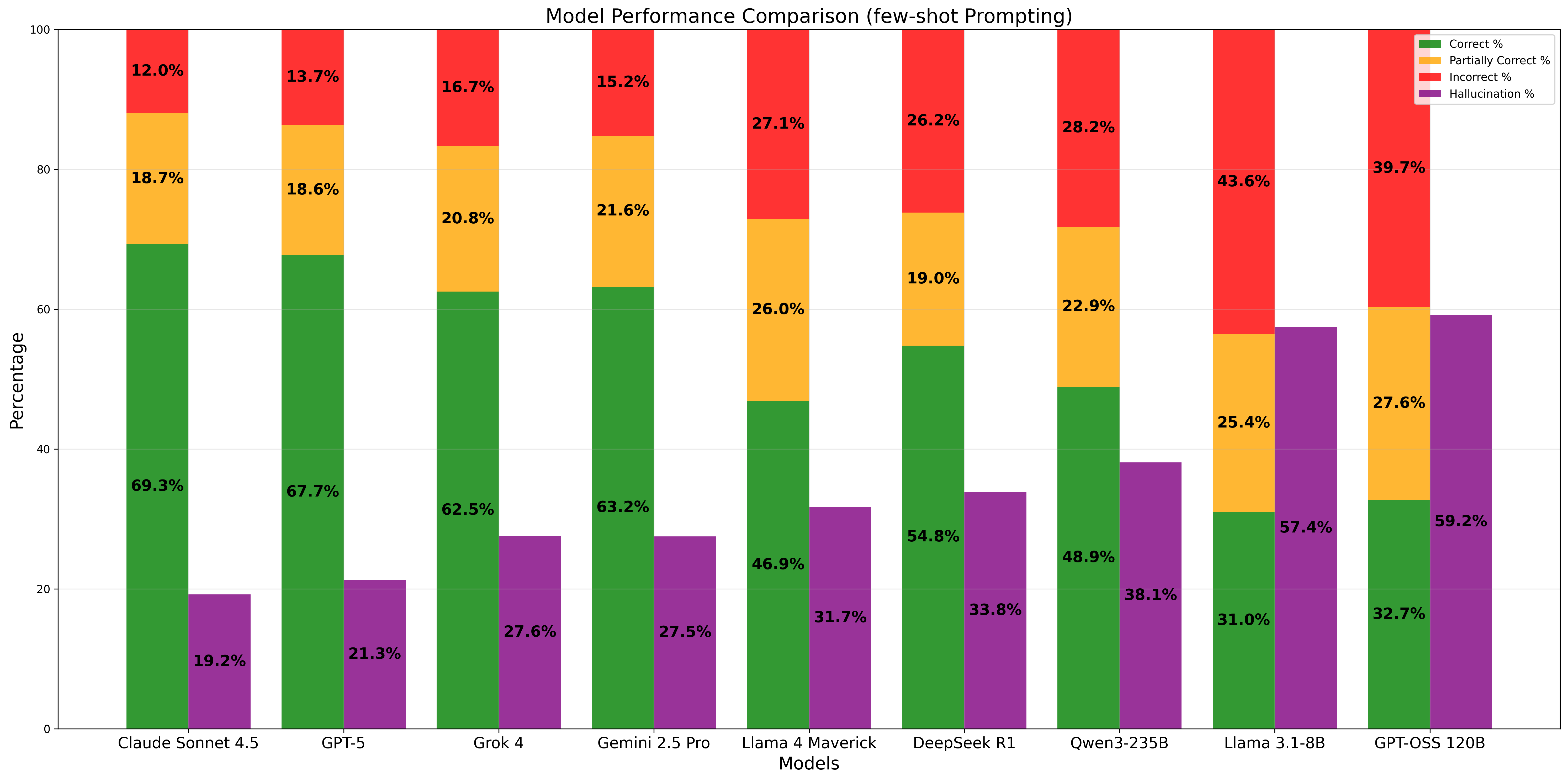}
    \caption{Overall performance comparison of evaluated LLMs under few-shot prompting. Models are ordered by category (closed-source followed by open-source). The stacked bars decompose responses into correct (green), partially correct (orange), and incorrect (red) categories, while separate purple bars indicate hallucination rates. Performance ranges from 30.96\% (Llama 3.1 8B) to 67.65\% (GPT-5), revealing substantial variation in Islamic law reasoning capabilities.}
    \label{fig:model_comparison}
\end{figure}

The results reveal substantial heterogeneity in LLMs' capacity for Islamic legal reasoning, with overall correctness ranging from 30.96\% to 67.65\%. Table~\ref{tab:correct_results} and Figure~\ref{fig:model_comparison} present the comprehensive breakdown across all 13 tasks and complexity levels.

Three critical findings emerge. First, performance stratifies into distinct tiers: frontier closed-source models (GPT-5, Claude Sonnet 4.5) achieve 65-68\% correctness and demonstrate consistent moderate competence across diverse task types, while smaller open-source models (Llama 3.1 8B, GPT-OSS-120B) fall below 37\%, exhibiting near-random performance on fundamental Islamic law questions. Mid-tier systems show binary performance patterns, excelling at pattern-matching extraction tasks while failing on genuine jurisprudential reasoning.

Second, task difficulty exhibits a paradoxical non-monotonic relationship with human expert-designated complexity. Tasks requiring verbatim knowledge retrieval prove widely challenging: Source Attribution (T5) achieves only 23-47\% correctness even among leading models, while Condition Enumeration (T7) and Comparative Distinction (T8) similarly constrain all systems. Conversely, certain high-complexity tasks elicit stronger performance than moderate-complexity ones, as models leverage semantic generalization about legal principles without requiring precise textual knowledge.

Third, this difficulty pattern exposes a fundamental architectural limitation. Models demonstrate sophisticated reasoning capabilities when explaining broad legal principles (\`{}illah identification, analogical application, maxim mapping) but fail systematically when Islamic jurisprudence demands exact, condition-specific knowledge from classical fiqh texts. The detailed task-level breakdown in Table~\ref{tab:correct_results} reveals that extraction tasks (T4: 76-98\% correctness) and principle-based reasoning (T10, T13: 70-88\%) leverage existing model strengths, while precise enumeration and statute synthesis represent a consistent drop in performance across architectures.

\subsection{Hallucination: The Shadow Side of Confident Incorrectness}
\label{sec:hallucination_analysis}

\begin{figure*}[t]
    \centering
    
    \begin{subfigure}{0.52\textwidth}
        \centering
        \includegraphics[width=\linewidth]{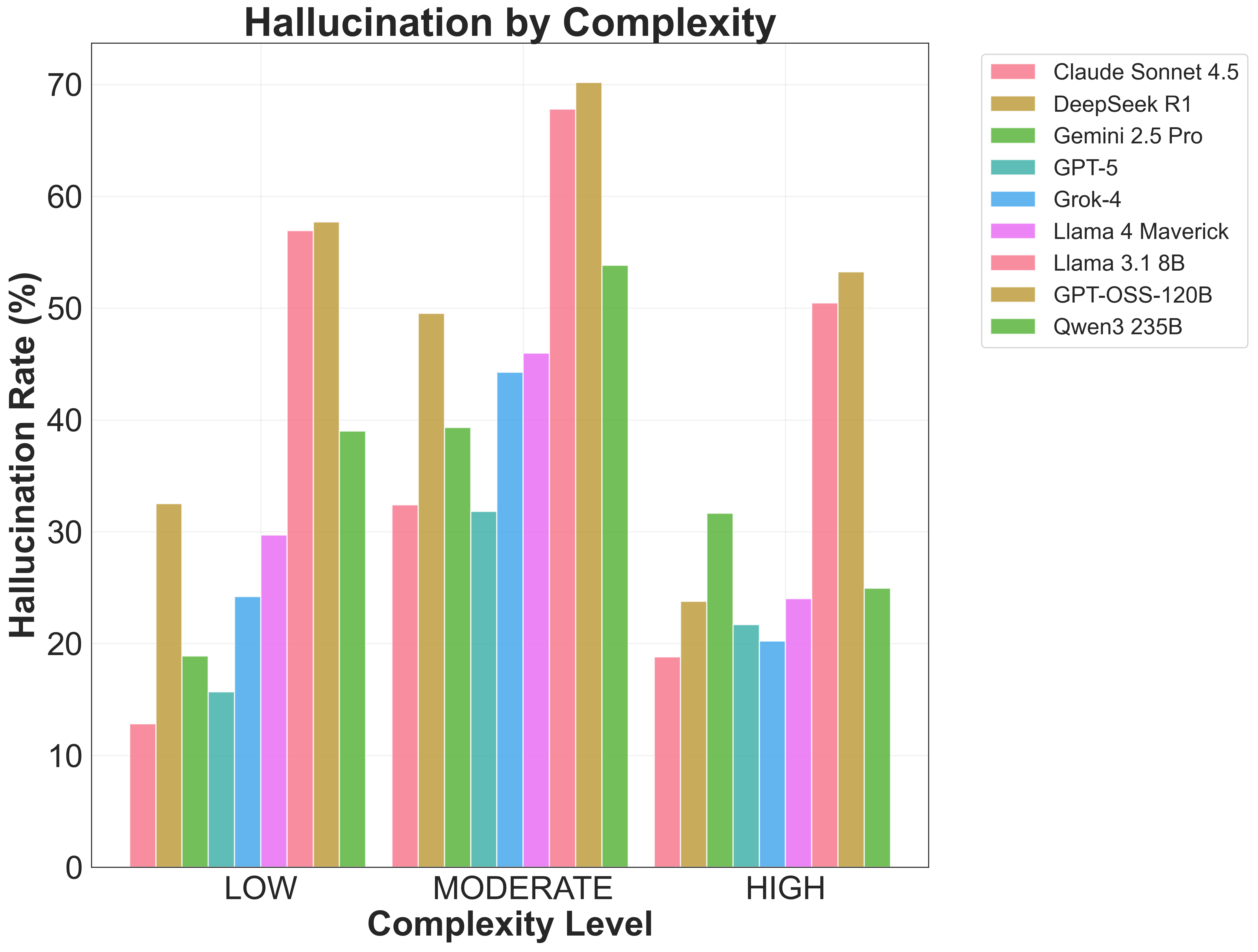}
        \caption{Hallucination rates by model across Low, Moderate, and High complexity tasks under few-shot prompting.}
        \label{fig:hallucination_left}
    \end{subfigure}
    \hfill
    \begin{subfigure}{0.44\textwidth}
        \centering
        \includegraphics[width=\linewidth]{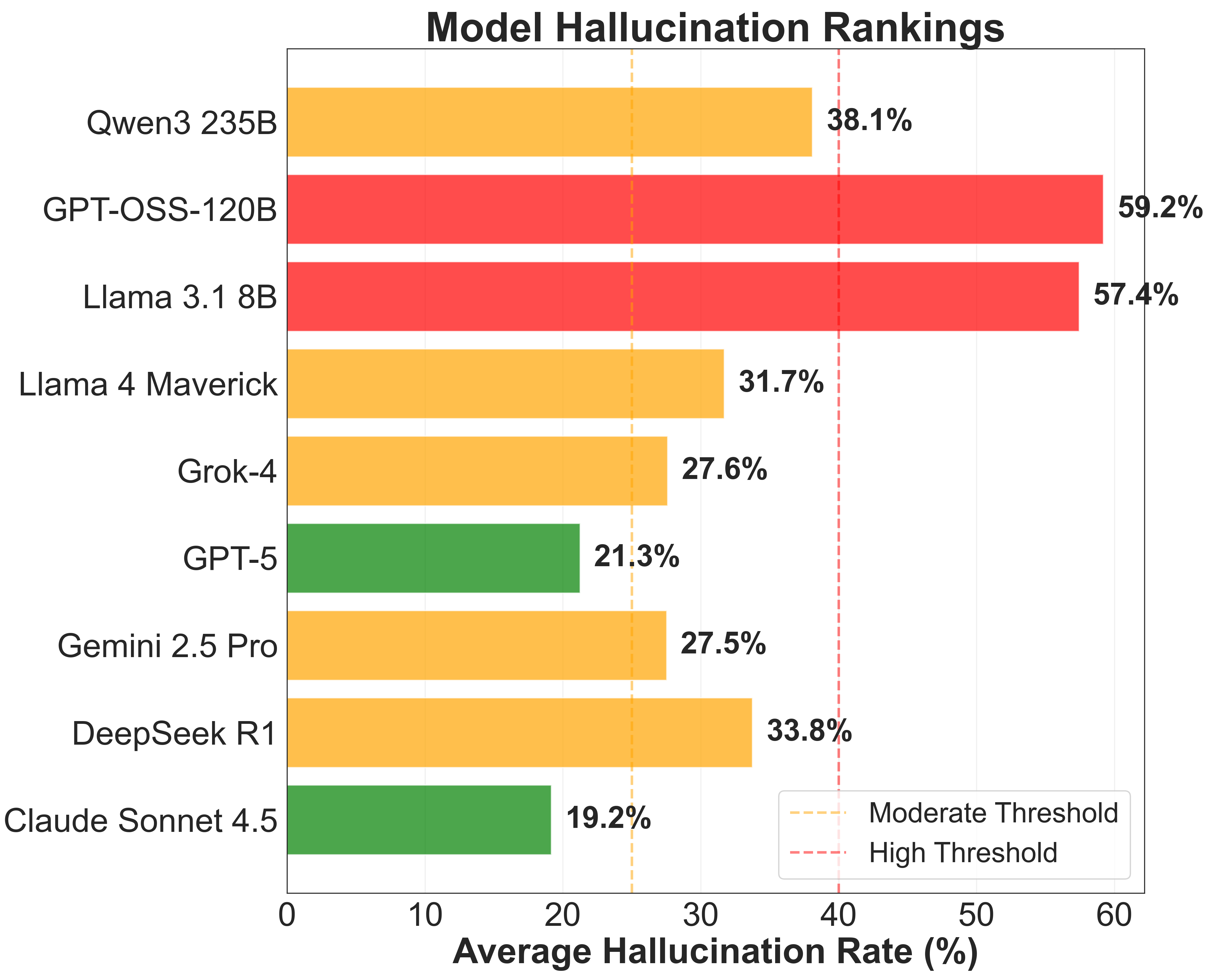}
        \caption{Overall hallucination severity rankings with thresholds at $<$25\% (low), 25--40\% (moderate), and $>$40\% (critical).}
        \label{fig:hallucination_right}
    \end{subfigure}

    \caption{\textbf{Hallucination Severity Analysis.}
Few-shot prompting reveals substantial variation in hallucination robustness: multiple models exceed the 25\% moderate-risk threshold, and several reach critical hallucination levels above 40\%, particularly on higher-complexity tasks. Panel (a) shows severity by task complexity, while Panel (b) summarizes overall rankings.}
    \label{fig:hallucination}
\end{figure*}

While correctness establishes baseline capability, hallucination rates reveal the critical safety profile of models for real-world Islamic advisory applications. Table~\ref{tab:hallucination_results} and Figure~\ref{fig:hallucination} present detailed statistics and severity patterns across task complexities.

\begin{table*}[t]
\centering
\caption{Hallucination rates of LLMs on our Islamic Legal Benchmark across different task complexities and prompting strategies (Pass@1 rates). Values represent hallucination percentages. Lowest (best) results per task are highlighted in \textbf{bold}.}
\label{tab:hallucination_results}
\resizebox{\textwidth}{!}{%
\begin{tabular}{ll|cccccc|ccc|cccc|c}
\toprule
\multirow{2}{*}{\textbf{Model}} & \multirow{2}{*}{\textbf{Prompt}} & \multicolumn{6}{c|}{\textbf{Low Complexity}} & \multicolumn{3}{c|}{\textbf{Moderate}} & \multicolumn{4}{c|}{\textbf{High Complexity}} & \multirow{2}{*}{\textbf{Overall}} \\
 & & \textbf{T1} & \textbf{T2} & \textbf{T3} & \textbf{T4} & \textbf{T5} & \textbf{T6} & \textbf{T7} & \textbf{T8} & \textbf{T9} & \textbf{T10} & \textbf{T11} & \textbf{T12} & \textbf{T13} & \\
\midrule
\multirow{2}{*}{Claude Sonnet 4.5} & Zero-Shot & \textbf{6.12} & 12.50 & 12.24 & 13.86 & 33.73 & 15.38 & 38.24 & 47.62 & 35.09 & 14.29 & \textbf{8.33} & 60.00 & 5.88 & 23.33 \\
 & Few-Shot & 8.16 & \textbf{10.42} & \textbf{10.20} & 8.61 & \textbf{24.10} & 15.38 & 47.06 & \textbf{23.81} & \textbf{26.32} & 12.70 & 16.67 & 40.00 & 5.88 & \textbf{19.18} \\
\midrule
\multirow{2}{*}{DeepSeek R1} & Zero-Shot & 24.49 & 41.67 & 32.65 & 8.61 & 51.81 & 30.77 & 64.71 & 52.38 & 47.37 & 19.05 & 33.33 & 60.00 & \textbf{0.00} & 35.91 \\
 & Few-Shot & 22.45 & 37.50 & 38.78 & 4.87 & 53.01 & 38.46 & 67.65 & 47.62 & 33.33 & 15.87 & 33.33 & 40.00 & 5.88 & 33.75 \\
\midrule
\multirow{2}{*}{Gemini 2.5 Pro} & Zero-Shot & 10.20 & 12.50 & 12.24 & 8.99 & 43.37 & 30.77 & 47.06 & 47.62 & 28.07 & 14.29 & 41.67 & 60.00 & 5.88 & 27.90 \\
 & Few-Shot & 8.16 & 12.50 & 12.24 & 7.49 & 42.17 & 30.77 & 47.06 & 42.86 & 28.07 & 19.05 & 41.67 & 60.00 & 5.88 & 27.53 \\
\midrule
\multirow{2}{*}{GPT-5} & Zero-Shot & \textbf{6.12} & 22.92 & 14.29 & 13.11 & 43.37 & 15.38 & \textbf{35.29} & 42.86 & 29.82 & \textbf{9.52} & \textbf{8.33} & \textbf{20.00} & 5.88 & 20.53 \\
 & Few-Shot & \textbf{6.12} & 14.58 & 18.37 & 6.37 & 40.96 & \textbf{7.69} & \textbf{35.29} & 28.57 & 31.58 & 15.87 & 25.00 & 40.00 & 5.88 & 21.25 \\
\midrule
\multirow{2}{*}{Grok-4} & Zero-Shot & 18.37 & 16.67 & 22.45 & 20.60 & 37.35 & \textbf{7.69} & 41.18 & 38.10 & 29.82 & 17.46 & 41.67 & \textbf{20.00} & \textbf{0.00} & 23.95 \\
 & Few-Shot & 16.33 & 22.92 & 20.41 & 7.87 & 46.99 & 30.77 & 58.82 & 47.62 & \textbf{26.32} & 15.87 & 25.00 & 40.00 & \textbf{0.00} & 27.61 \\
\midrule
\multirow{2}{*}{Llama 4 Maverick 17B} & Zero-Shot & 24.49 & 50.00 & 40.82 & \textbf{1.50} & 56.63 & 15.38 & 52.94 & 33.33 & 31.58 & 15.87 & 16.67 & 60.00 & \textbf{0.00} & 30.71 \\
 & Few-Shot & 20.41 & 39.58 & 46.94 & 3.00 & 53.01 & 15.38 & 52.94 & 42.86 & 42.11 & 11.11 & 25.00 & 60.00 & \textbf{0.00} & 31.72 \\
\midrule
\multirow{2}{*}{Llama 3.1 8B} & Zero-Shot & 69.39 & 79.17 & 69.39 & 17.98 & 66.27 & 46.15 & 88.24 & 71.43 & 71.93 & 53.97 & 58.33 & 80.00 & 29.41 & 61.67 \\
 & Few-Shot & 69.39 & 75.00 & 77.55 & 1.87 & 63.86 & 53.85 & 73.53 & 66.67 & 63.16 & 50.79 & 41.67 & 80.00 & 29.41 & 57.44 \\
\midrule
\multirow{2}{*}{GPT-OSS-120B} & Zero-Shot & 61.22 & 85.42 & 65.31 & 43.45 & 60.24 & 38.46 & 82.35 & 61.90 & 50.88 & 42.86 & 50.00 & 40.00 & 35.29 & 55.18 \\
 & Few-Shot & 67.35 & 83.33 & 65.31 & 20.22 & 63.86 & 46.15 & 76.47 & 76.19 & 57.89 & 47.62 & 50.00 & 80.00 & 35.29 & 59.21 \\
\midrule
\multirow{2}{*}{Qwen3 235B} & Zero-Shot & 32.65 & 54.17 & 48.98 & 11.24 & 54.22 & 38.46 & 64.71 & 66.67 & 36.84 & 26.98 & 41.67 & 80.00 & 17.65 & 44.17 \\
 & Few-Shot & 32.65 & 52.08 & 61.22 & 3.00 & 54.22 & 30.77 & 64.71 & 47.62 & 49.12 & 20.63 & 33.33 & 40.00 & 5.88 & 38.10 \\
\bottomrule
\end{tabular}%
}
\end{table*}

The analysis reveals a trimodal safety distribution with implications for deployment. Claude Sonnet 4.5 achieves the strongest safety profile at 19.18\% hallucination, with GPT-5 and Grok-4 remaining below 25\% to form a relatively secure tier suitable for supervised deployment. A middle tier, comprising Gemini 2.5 Pro, DeepSeek R1, Llama 4 Maverick, and Qwen3 235B exceeds 27-38\% hallucination rates, indicating that roughly one-third of responses contain fabricated Islamic legal content and require substantial expert oversight. The bottom tier (Llama 3.1 8B, GPT-OSS-120B) exhibits hallucination above 55\%, where more than half of all responses actively contain inaccurate religious content that precludes real-world deployment.

Task-level patterns expose the underlying failure mechanism. Moderate-complexity tasks requiring precise textual knowledge trigger the highest hallucination: Condition Enumeration (T7) and Comparative Rule Distinction (T8) average 50-70\% fabrication rates across models, while Source Attribution (T5) reaches 24-64\%. When models lack exact knowledge of legal conditions or juristic citations from classical texts, they fabricate plausible-sounding but incorrect requirements rather than acknowledging uncertainty.

Significantly, this pattern mirrors the non-monotonic difficulty observed in correctness analysis. High-complexity tasks permitting flexible reasoning (T13: Maxim Mapping) elicit minimal hallucination despite their difficulty, while moderate-complexity tasks demanding knowledge accuracy (T7, T8) produce substantially higher fabrication rates. Models can reason extensively about abstract legal principles without hallucinating, but systematically invent precise conditions when lacking training exposure, a fundamental vulnerability for specialized legal domains where exactness is non-negotiable.

\subsection{Error Modes: Incorrect Responses, Partial Knowledge, and Strategic Abstention}
\label{sec:error_modes}

Beyond binary correctness and hallucination, the distribution of incorrect responses, partial correctness, and abstention patterns reveals models' uncertainty-management behavior and the granularity of their reasoning. Tables~\ref{tab:incorrect_results} and~\ref{tab:abstention_results} provide comprehensive statistics.

\begin{table*}[t]
\centering
\caption{Incorrect response rates of LLMs on our Islamic Legal Benchmark across different task complexities and prompting strategies (Pass@1 rates). Values represent incorrect response percentages. Lowest (best) results per task are highlighted in \textbf{bold}.}
\label{tab:incorrect_results}
\resizebox{\textwidth}{!}{%
\begin{tabular}{ll|cccccc|ccc|cccc|c}
\toprule
\multirow{2}{*}{\textbf{Model}} & \multirow{2}{*}{\textbf{Prompt}} & \multicolumn{6}{c|}{\textbf{Low Complexity}} & \multicolumn{3}{c|}{\textbf{Moderate}} & \multicolumn{4}{c|}{\textbf{High Complexity}} & \multirow{2}{*}{\textbf{Overall}} \\
 & & \textbf{T1} & \textbf{T2} & \textbf{T3} & \textbf{T4} & \textbf{T5} & \textbf{T6} & \textbf{T7} & \textbf{T8} & \textbf{T9} & \textbf{T10} & \textbf{T11} & \textbf{T12} & \textbf{T13} & \\
\midrule
\multirow{2}{*}{Claude Sonnet 4.5} & Zero-Shot & \textbf{6.12} & 10.42 & 12.24 & 16.48 & 16.87 & \textbf{0.00} & \textbf{14.71} & 19.05 & 12.28 & 9.52 & \textbf{8.33} & 40.00 & 11.76 & 13.68 \\
 & Few-Shot & 8.16 & 10.42 & \textbf{10.20} & 8.99 & 16.87 & 15.38 & \textbf{14.71} & \textbf{4.76} & 12.28 & 7.94 & \textbf{8.33} & \textbf{20.00} & 17.65 & \textbf{11.98} \\
\midrule
\multirow{2}{*}{DeepSeek R1} & Zero-Shot & 24.49 & 39.58 & 32.65 & 4.49 & 24.10 & 30.77 & 23.53 & 33.33 & 28.07 & 14.29 & 25.00 & 40.00 & \textbf{5.88} & 25.09 \\
 & Few-Shot & 22.45 & 37.50 & 38.78 & 3.75 & 31.33 & 30.77 & 26.47 & 19.05 & 19.30 & 7.94 & 25.00 & 60.00 & 17.65 & 26.15 \\
\midrule
\multirow{2}{*}{Gemini 2.5 Pro} & Zero-Shot & \textbf{6.12} & 10.42 & 12.24 & 3.37 & 22.89 & 30.77 & 23.53 & 19.05 & \textbf{7.02} & \textbf{6.35} & 25.00 & 40.00 & \textbf{5.88} & 16.36 \\
 & Few-Shot & 8.16 & \textbf{8.33} & \textbf{10.20} & 2.25 & 21.69 & 30.77 & 17.65 & 23.81 & 15.79 & 7.94 & 25.00 & \textbf{20.00} & \textbf{5.88} & 15.19 \\
\midrule
\multirow{2}{*}{GPT-5} & Zero-Shot & 8.16 & 20.83 & 14.29 & 13.86 & 21.69 & 15.38 & 20.59 & 19.05 & 12.28 & 12.70 & \textbf{8.33} & 40.00 & \textbf{5.88} & 16.39 \\
 & Few-Shot & \textbf{6.12} & 10.42 & 18.37 & 5.99 & \textbf{14.46} & \textbf{0.00} & 17.65 & 9.52 & 21.05 & 14.29 & 16.67 & \textbf{20.00} & 23.53 & 13.70 \\
\midrule
\multirow{2}{*}{Grok-4} & Zero-Shot & 18.37 & 16.67 & 22.45 & 23.22 & 16.87 & 7.69 & 20.59 & 19.05 & 12.28 & 12.70 & 33.33 & \textbf{20.00} & \textbf{5.88} & 17.62 \\
 & Few-Shot & 14.29 & 22.92 & 20.41 & 8.99 & 20.48 & 15.38 & 26.47 & 19.05 & \textbf{7.02} & 7.94 & 16.67 & \textbf{20.00} & 17.65 & 16.71 \\
\midrule
\multirow{2}{*}{Llama 4 Maverick 17B} & Zero-Shot & 22.45 & 47.92 & 42.86 & \textbf{1.50} & 32.53 & 15.38 & 20.59 & 23.81 & 15.79 & 20.63 & 25.00 & 40.00 & \textbf{5.88} & 24.18 \\
 & Few-Shot & 20.41 & 39.58 & 46.94 & 3.00 & 31.33 & 15.38 & 23.53 & 14.29 & 28.07 & 19.05 & 33.33 & 60.00 & 17.65 & 27.12 \\
\midrule
\multirow{2}{*}{Llama 3.1 8B} & Zero-Shot & 69.39 & 77.08 & 65.31 & 20.60 & 54.22 & 38.46 & 50.00 & 47.62 & 42.11 & 31.75 & 41.67 & 80.00 & 17.65 & 48.91 \\
 & Few-Shot & 67.35 & 72.92 & 73.47 & 2.25 & 50.60 & 30.77 & 44.12 & 47.62 & 35.09 & 25.40 & 25.00 & 60.00 & 23.53 & 42.93 \\
\midrule
\multirow{2}{*}{GPT-OSS-120B} & Zero-Shot & 59.18 & 85.42 & 65.31 & 50.19 & 32.53 & 15.38 & 35.29 & 28.57 & 29.82 & 22.22 & 41.67 & 40.00 & 17.65 & 40.25 \\
 & Few-Shot & 65.31 & 83.33 & 65.31 & 20.22 & 42.17 & 23.08 & 38.24 & 28.57 & 35.09 & 20.63 & 25.00 & 40.00 & 11.76 & 38.36 \\
\midrule
\multirow{2}{*}{Qwen3 235B} & Zero-Shot & 26.53 & 52.08 & 48.98 & 12.73 & 24.10 & 15.38 & 23.53 & 28.57 & 22.81 & 12.70 & 25.00 & 60.00 & 17.65 & 28.47 \\
 & Few-Shot & 30.61 & 50.00 & 59.18 & 3.37 & 30.12 & 23.08 & 32.35 & 23.81 & 22.81 & 12.70 & 33.33 & 40.00 & \textbf{5.88} & 28.25 \\
\bottomrule
\end{tabular}%
}
\end{table*}

\begin{table*}[t]
\centering
\caption{Abstention (refusal) rates of LLMs on our Islamic Legal Benchmark across different task complexities and prompting strategies (Pass@1 rates). Values represent abstention percentages. Lowest results per task are highlighted in \textbf{bold}.}
\label{tab:abstention_results}
\resizebox{\textwidth}{!}{%
\begin{tabular}{ll|cccccc|ccc|cccc|c}
\toprule
\multirow{2}{*}{\textbf{Model}} & \multirow{2}{*}{\textbf{Prompt}} & \multicolumn{6}{c|}{\textbf{Low Complexity}} & \multicolumn{3}{c|}{\textbf{Moderate}} & \multicolumn{4}{c|}{\textbf{High Complexity}} & \multirow{2}{*}{\textbf{Overall}} \\
 & & \textbf{T1} & \textbf{T2} & \textbf{T3} & \textbf{T4} & \textbf{T5} & \textbf{T6} & \textbf{T7} & \textbf{T8} & \textbf{T9} & \textbf{T10} & \textbf{T11} & \textbf{T12} & \textbf{T13} & \\
\midrule
\multirow{2}{*}{Claude Sonnet 4.5} & Zero-Shot & \textbf{0.00} & \textbf{0.00} & \textbf{0.00} & 5.62 & 14.46 & \textbf{0.00} & 2.94 & \textbf{0.00} & \textbf{0.00} & 3.17 & 8.33 & \textbf{0.00} & \textbf{0.00} & 2.66 \\
 & Few-Shot & \textbf{0.00} & \textbf{0.00} & \textbf{0.00} & \textbf{0.00} & 26.51 & \textbf{0.00} & \textbf{0.00} & 9.52 & \textbf{0.00} & 3.17 & 8.33 & \textbf{0.00} & \textbf{0.00} & 3.66 \\
\midrule
\multirow{2}{*}{DeepSeek R1} & Zero-Shot & \textbf{0.00} & \textbf{0.00} & \textbf{0.00} & \textbf{0.00} & \textbf{0.00} & \textbf{0.00} & \textbf{0.00} & \textbf{0.00} & \textbf{0.00} & 1.59 & \textbf{0.00} & \textbf{0.00} & \textbf{0.00} & 0.12 \\
 & Few-Shot & \textbf{0.00} & \textbf{0.00} & \textbf{0.00} & \textbf{0.00} & \textbf{0.00} & \textbf{0.00} & \textbf{0.00} & \textbf{0.00} & \textbf{0.00} & \textbf{0.00} & 8.33 & \textbf{0.00} & \textbf{0.00} & 0.64 \\
\midrule
\multirow{2}{*}{Gemini 2.5 Pro} & Zero-Shot & \textbf{0.00} & \textbf{0.00} & \textbf{0.00} & \textbf{0.00} & 6.02 & \textbf{0.00} & \textbf{0.00} & \textbf{0.00} & \textbf{0.00} & \textbf{0.00} & \textbf{0.00} & \textbf{0.00} & \textbf{0.00} & 0.46 \\
 & Few-Shot & \textbf{0.00} & \textbf{0.00} & \textbf{0.00} & \textbf{0.00} & 6.02 & \textbf{0.00} & \textbf{0.00} & \textbf{0.00} & \textbf{0.00} & \textbf{0.00} & \textbf{0.00} & \textbf{0.00} & \textbf{0.00} & 0.46 \\
\midrule
\multirow{2}{*}{GPT-5} & Zero-Shot & \textbf{0.00} & \textbf{0.00} & \textbf{0.00} & \textbf{0.00} & 6.02 & \textbf{0.00} & \textbf{0.00} & \textbf{0.00} & \textbf{0.00} & \textbf{0.00} & \textbf{0.00} & \textbf{0.00} & \textbf{0.00} & 0.46 \\
 & Few-Shot & \textbf{0.00} & \textbf{0.00} & \textbf{0.00} & \textbf{0.00} & 1.20 & \textbf{0.00} & \textbf{0.00} & \textbf{0.00} & \textbf{0.00} & \textbf{0.00} & \textbf{0.00} & \textbf{0.00} & \textbf{0.00} & 0.09 \\
\midrule
\multirow{2}{*}{Grok-4} & Zero-Shot & \textbf{0.00} & \textbf{0.00} & \textbf{0.00} & \textbf{0.00} & \textbf{0.00} & \textbf{0.00} & \textbf{0.00} & \textbf{0.00} & \textbf{0.00} & \textbf{0.00} & \textbf{0.00} & \textbf{0.00} & \textbf{0.00} & \textbf{0.00} \\
 & Few-Shot & \textbf{0.00} & \textbf{0.00} & \textbf{0.00} & \textbf{0.00} & \textbf{0.00} & \textbf{0.00} & \textbf{0.00} & \textbf{0.00} & \textbf{0.00} & 1.59 & 8.33 & \textbf{0.00} & \textbf{0.00} & 0.76 \\
\midrule
\multirow{2}{*}{Llama 4 Maverick 17B} & Zero-Shot & \textbf{0.00} & \textbf{0.00} & \textbf{0.00} & \textbf{0.00} & 1.20 & \textbf{0.00} & \textbf{0.00} & \textbf{0.00} & \textbf{0.00} & \textbf{0.00} & \textbf{0.00} & \textbf{0.00} & \textbf{0.00} & 0.09 \\
 & Few-Shot & \textbf{0.00} & \textbf{0.00} & \textbf{0.00} & \textbf{0.00} & 1.20 & \textbf{0.00} & \textbf{0.00} & \textbf{0.00} & \textbf{0.00} & \textbf{0.00} & \textbf{0.00} & \textbf{0.00} & \textbf{0.00} & 0.09 \\
\midrule
\multirow{2}{*}{Llama 3.1 8B} & Zero-Shot & \textbf{0.00} & \textbf{0.00} & 4.08 & \textbf{0.00} & \textbf{0.00} & \textbf{0.00} & \textbf{0.00} & \textbf{0.00} & \textbf{0.00} & \textbf{0.00} & \textbf{0.00} & \textbf{0.00} & \textbf{0.00} & 0.31 \\
 & Few-Shot & 2.04 & \textbf{0.00} & \textbf{0.00} & \textbf{0.00} & \textbf{0.00} & \textbf{0.00} & 2.94 & 4.76 & \textbf{0.00} & \textbf{0.00} & \textbf{0.00} & \textbf{0.00} & \textbf{0.00} & 0.75 \\
\midrule
\multirow{2}{*}{GPT-OSS-120B} & Zero-Shot & \textbf{0.00} & \textbf{0.00} & \textbf{0.00} & \textbf{0.00} & \textbf{0.00} & \textbf{0.00} & \textbf{0.00} & \textbf{0.00} & \textbf{0.00} & \textbf{0.00} & \textbf{0.00} & \textbf{0.00} & \textbf{0.00} & \textbf{0.00} \\
 & Few-Shot & \textbf{0.00} & \textbf{0.00} & \textbf{0.00} & \textbf{0.00} & \textbf{0.00} & \textbf{0.00} & \textbf{0.00} & \textbf{0.00} & \textbf{0.00} & \textbf{0.00} & 16.67 & \textbf{0.00} & \textbf{0.00} & 1.28 \\
\midrule
\multirow{2}{*}{Qwen3 235B} & Zero-Shot & \textbf{0.00} & \textbf{0.00} & \textbf{0.00} & \textbf{0.00} & \textbf{0.00} & \textbf{0.00} & \textbf{0.00} & \textbf{0.00} & \textbf{0.00} & \textbf{0.00} & 8.33 & \textbf{0.00} & \textbf{0.00} & 0.64 \\
 & Few-Shot & \textbf{0.00} & \textbf{0.00} & \textbf{0.00} & \textbf{0.00} & \textbf{0.00} & \textbf{0.00} & \textbf{0.00} & \textbf{0.00} & \textbf{0.00} & \textbf{0.00} & \textbf{0.00} & \textbf{0.00} & \textbf{0.00} & \textbf{0.00} \\
\bottomrule
\end{tabular}%
}
\end{table*}

Incorrect response rates stratify into three failure profiles mirroring the correctness hierarchy. Leading models (Claude Sonnet 4.5, GPT-5, Gemini 2.5 Pro) maintain incorrect rates below 15\%, with errors concentrated on specific challenging tasks rather than distributed uniformly, suggesting broad competence with targeted reasoning gaps. Conversely, weaker models exhibit inverted performance: Llama 3.1 8B and GPT-OSS-120B produce incorrect responses on 38-43\% of queries, rates that exceed their correctness when partial score is excluded. These models reliably generate wrong answers, creating acute risks for religious guidance applications where incorrect rulings could invalidate worship practices.

Partial correctness patterns expose a subtle but critical vulnerability. Tasks requiring precise attribution or doctrinal distinction (T5: Source Attribution, T8: Comparative Distinction) elicit elevated partial scores across leading models, indicating systems can recognize that juristic opinions or doctrinal differences exist but fail to correctly attribute statements to scholars or characterize distinctions accurately. This ``almost correct'' failure mode presents particular risks: outputs appear plausible and demonstrate apparent Islamic legal knowledge, yet contain subtle errors difficult for non-experts to identify.

Most critically, abstention analysis reveals severe uncertainty-miscalibration across all models. Despite achieving only 65-68\% correctness, leading models abstain on fewer than 4\% of queries, with several models never refusing any task. Claude Sonnet 4.5 shows marginally elevated abstention (3.66\%, concentrated on T5), but even this remains grossly disproportionate to actual knowledge boundaries. Models attempt to answer nearly all queries regardless of knowledge sufficiency, producing incorrect or hallucinated content rather than acknowledging uncertainty. This overconfidence amplifies the safety risks identified earlier: models not only fail but fail confidently, providing no calibration signals to alert users that responses may be unreliable.

\subsection{False Premise Queries: Assessing Critical Reasoning and Sycophantic Vulnerability}
\label{sec:fpq_analysis}

\begin{figure*}[t]
    \centering
    \begin{subfigure}{0.48\columnwidth}
        \centering
        \includegraphics[width=\linewidth]{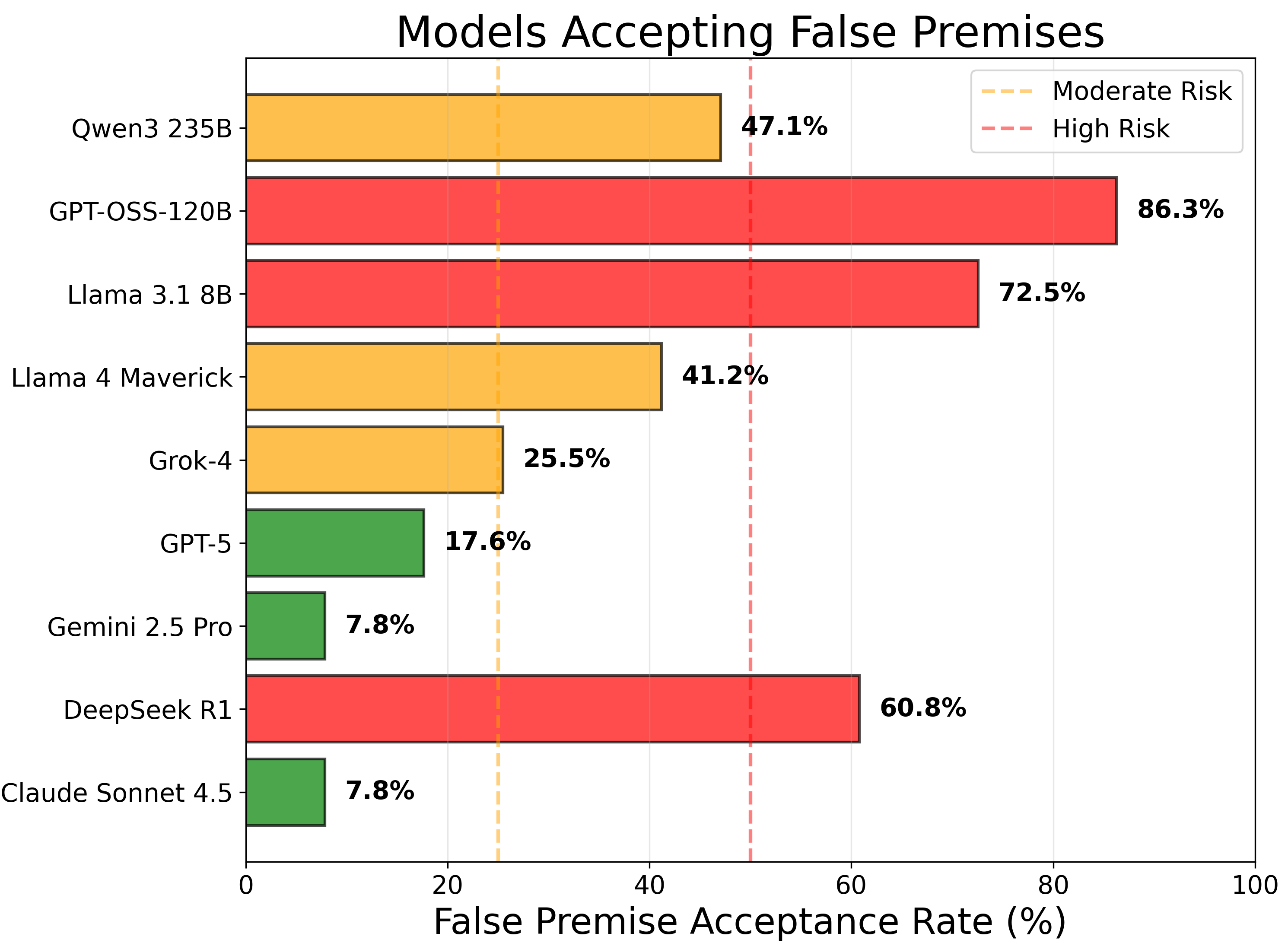}
        \caption{Acceptance rates for FPQs under few-shot prompting. Bars show the proportion of misleading queries each model accepts, with green ($<$25\%), orange (25--50\%), and red ($>$50\%) indicating increasing risk.}
        \label{fig:fpq_acceptance_left}
    \end{subfigure}
    \hfill
    \begin{subfigure}{0.48\columnwidth}
        \centering
        \includegraphics[width=\linewidth]{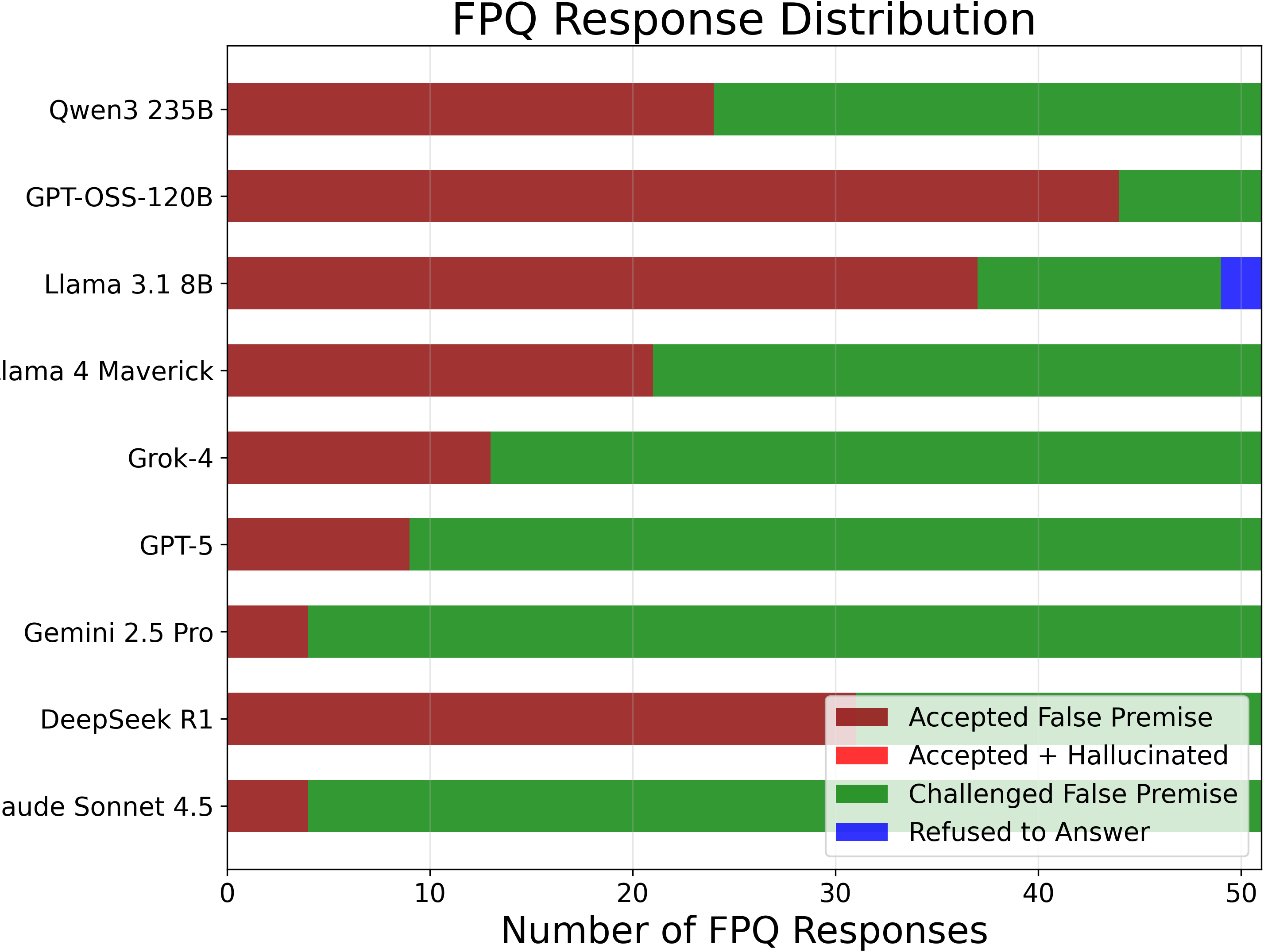}
        \caption{Response-type distribution for FPQs: accepted (red), hallucinated acceptance (dark red), challenged (green), and refused (blue), illustrating how models handle misleading premises.}
        \label{fig:fpq_acceptance_right}
    \end{subfigure}

\caption{\textbf{False Premise Query (FPQ) Analysis.}
Few-shot prompting exposes large differences in how frontier models handle misleading assumptions: some models reliably challenge false premises while others accept them at high rates, sometimes with added hallucinated justification. These patterns reveal distinct levels of critical-reasoning robustness across architectures and prompting regimes (acceptance rates in Panel (a); response-type distributions in Panel (b)).}
    \label{fig:fpq_acceptance}
\end{figure*}

To evaluate critical reasoning capabilities, we embedded 51 deliberately misleading False Premise Queries (FPQs) within Tasks 1, 2, 3, and 7, testing whether models detect and challenge factually incorrect Islamic law assumptions rather than sycophantically accepting user misinformation. Table~\ref{tab:fpq_sycophancy} and Figure~\ref{fig:fpq_acceptance} present comprehensive response distributions.

\begin{table*}[t]
\centering
\caption{False Premise Query (FPQ) Analysis: Sycophancy Detection in LLM Responses. Models are evaluated on their ability to identify and reject misleading queries in Islamic law. Lower acceptance rates indicate better critical reasoning.}
\label{tab:fpq_sycophancy}
\resizebox{\textwidth}{!}{%
\begin{tabular}{llccc cc}
\toprule
\multirow{2}{*}{\textbf{Model}} &
\multirow{2}{*}{\textbf{Prompt}} &
\multicolumn{3}{c}{\textbf{Response Distribution}} &
\multicolumn{2}{c}{\textbf{Key Metrics}} \\
 & & \textbf{Accepted} & \textbf{Challenged} & \textbf{Refused} &
 \textbf{Accept. Rate \%} & \textbf{Challenge Rate \%} \\
\cmidrule(lr){3-5} \cmidrule(lr){6-7}
\multirow{2}{*}{Claude Sonnet 4.5} & Zero-Shot & 3 & 48 & 0 & \textbf{5.88} & 94.12 \\
 & Few-Shot & 4 & 47 & 0 & \textbf{7.84} & 92.16 \\
\midrule
\multirow{2}{*}{DeepSeek R1} & Zero-Shot & 26 & 25 & 0 & 50.98 & 49.02 \\
 & Few-Shot & 31 & 20 & 0 & 60.78 & 39.22 \\
\midrule
\multirow{2}{*}{Gemini 2.5 Pro} & Zero-Shot & 5 & 46 & 0 & 9.80 & 90.20 \\
 & Few-Shot & 4 & 47 & 0 & \textbf{7.84} & 92.16 \\
\midrule
\multirow{2}{*}{GPT-5} & Zero-Shot & 9 & 42 & 0 & 17.65 & 82.35 \\
 & Few-Shot & 9 & 42 & 0 & 17.65 & 82.35 \\
\midrule
\multirow{2}{*}{Grok-4} & Zero-Shot & 6 & 45 & 0 & 11.76 & 88.24 \\
 & Few-Shot & 13 & 38 & 0 & 25.49 & 74.51 \\
\midrule
\multirow{2}{*}{Llama 4 Maverick} & Zero-Shot & 23 & 28 & 0 & 45.10 & 54.90 \\
 & Few-Shot & 21 & 30 & 0 & 41.18 & 58.82 \\
\midrule
\multirow{2}{*}{Llama 3.1 8B} & Zero-Shot & 38 & 11 & 2 & 74.51 & 21.57 \\
 & Few-Shot & 37 & 12 & 2 & 72.55 & 23.53 \\
\midrule
\multirow{2}{*}{GPT-OSS-120B} & Zero-Shot & 41 & 10 & 0 & 80.39 & 19.61 \\
 & Few-Shot & 44 & 7 & 0 & 86.27 & 13.73 \\
\midrule
\multirow{2}{*}{Qwen3 235B} & Zero-Shot & 20 & 31 & 0 & 39.22 & 60.78 \\
 & Few-Shot & 24 & 27 & 0 & 47.06 & 52.94 \\
\bottomrule
\end{tabular}%
}
\end{table*}

The results show sycophantic vulnerability across most evaluated models, stratifying into three distinct profiles. Claude Sonnet 4.5 and Gemini 2.5 Pro demonstrate exceptional critical reasoning by challenging 92\% of false premises while accepting fewer than 8\%, indicating deployment-appropriate skepticism toward user-provided misinformation. A middle tier (GPT-5, Grok-4) accepts 18-25\% of false premises, substantially better than weaker systems, but still indicating that roughly one-in-four misleading assumptions pass unchallenged. The bottom tier exhibits a much higher degree of vulnerability: GPT-OSS-120B accepts 86\% of false premises, with Llama 3.1 8B, DeepSeek R1, and Llama 4 Maverick similarly failing to critically evaluate query assumptions.

This failure represents a compounding safety vulnerability beyond mere incorrectness. Models lacking sufficient Islamic jurisprudence grounding not only produce wrong answers but also fail to recognize and reject false information embedded in user queries. The strong negative correlation between FPQ acceptance and overall performance (Pearson's r = –0.87, p $<$ 0.01) confirms this pattern: weaker models are doubly compromised, generating more errors while being less capable of detecting them, a profile fundamentally unsuitable for responsible deployment in religious guidance contexts.

Critically, in-context learning (ICL) fails to mitigate and often exacerbates sycophantic vulnerability. Average FPQ acceptance increased from 37.25\% (zero-shot) to 40.74\% (few-shot), with DeepSeek R1 and GPT-OSS-120B showing substantial degradation. This counterintuitive pattern suggests that few-shot examples reinforce superficial pattern-matching that increases premise acceptance when exact knowledge is absent, further undermining ICL-based approaches for specialized legal reasoning domains.

\subsection{False Confidence: The Danger Zone of Miscalibrated Models}
\label{sec:confidence_analysis}

\begin{figure}[t]
    \centering
    \includegraphics[width=0.85\columnwidth]{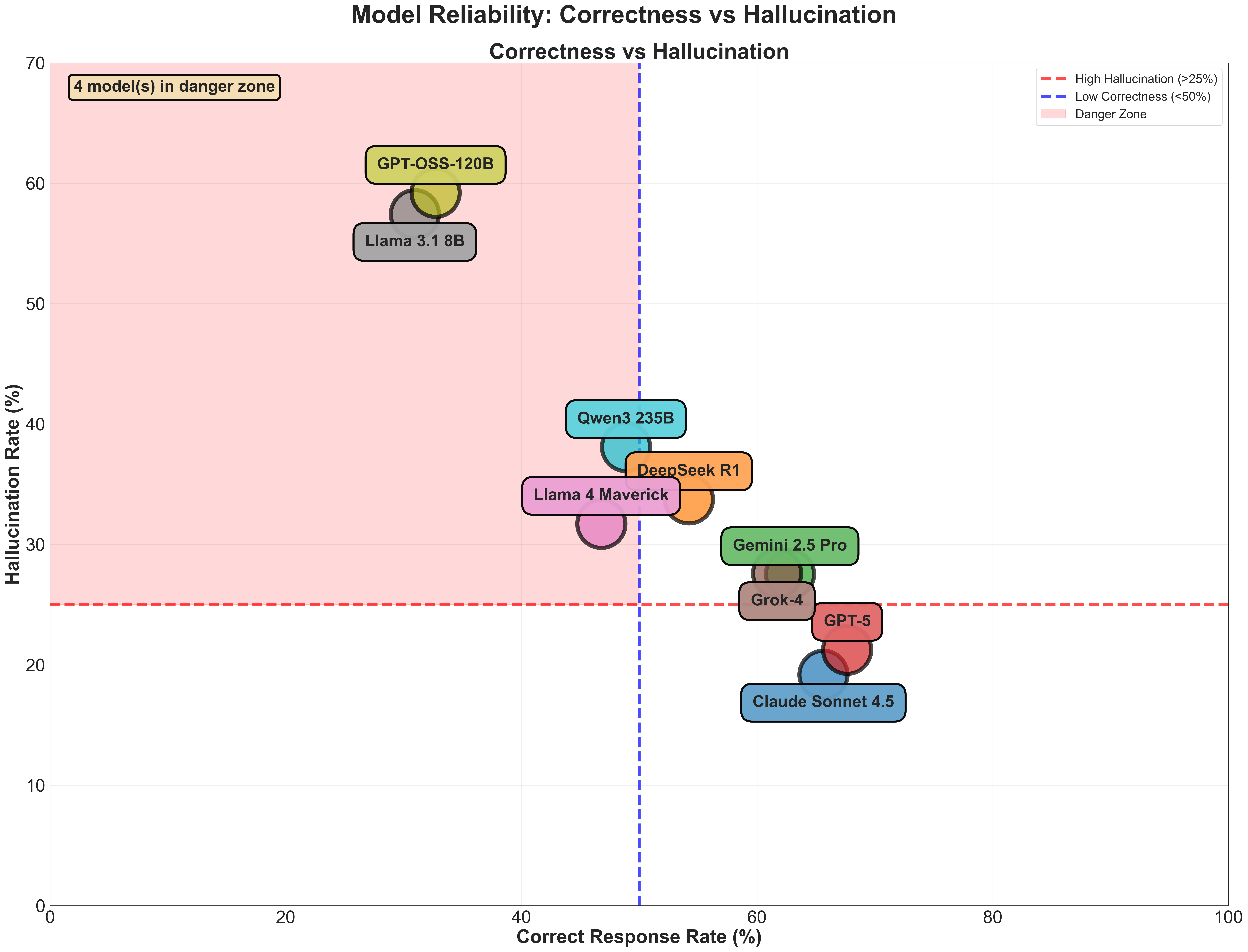}
    \caption{Scatter plot mapping correctness versus hallucination rates under few-shot prompting. Each point represents one model, with annotations identifying specific systems. The red-shaded "danger zone" (correctness $<$50\%, hallucination $>$25\%) identifies models exhibiting confident incorrectness. Three models occupy this critical region, producing incorrect Islamic legal guidance while showing no uncertainty signals, creating acute safety risks for advisory applications.}
    \label{fig:confidence_correctness}
\end{figure}

Figure~\ref{fig:confidence_correctness} maps the critical relationship between correctness and hallucination, revealing patterns of confident incorrectness that present acute deployment risks.

Four models occupy the danger zone, characterized by low correctness coupled with high hallucination. GPT-OSS-120B and Llama 3.1 8B exhibit a high level of miscalibrated confidence, producing incorrect Islamic legal guidance in more than two-thirds of responses while hallucinating in nearly 60\% of outputs. These models do not merely fail to answer correctly; they confidently fabricate Islamic legal content, creating particularly high-risk failure modes. Llama 4 Maverick and Qwen3 235B approaches the danger zone boundary, exhibiting marginal but still concerning miscalibration.

This confident incorrectness poses acute risks for Islamic advisory applications. Users lacking Islamic legal expertise to correctly evaluate responses may trust authoritative-sounding but fundamentally flawed outputs, leading to questionable religious practice. The absence of uncertainty signals (minimal abstention rates below 1\%) compounds the problem: models do not indicate that their responses may be unreliable, encouraging uncritical acceptance of potentially incorrect guidance.

Conversely, leading models demonstrate substantially better calibration. GPT-5 (67.65\% correct, 21.25\% hallucination) and Claude Sonnet 4.5 (65.63\% correct, 19.18\% hallucination) maintain roughly 3:1 ratios of correct to hallucinated responses, indicating more appropriate confidence. However, even these top performers hallucinate in approximately one-fifth of outputs, underscoring that no evaluated model achieves deployment-ready reliability without human expert oversight. Gemini 2.5 Pro and Grok-4 occupy intermediate positions, maintaining positive but less favorable correctness-to-hallucination ratios. From a perspective of the entire thing considered, even these ratios of top-performing models are not acceptable and applicable in Islamic law or any other high-stakes field where the accuracy of responses from these models is an absolute necessity.

\subsection{In-Context Learning Failure: Why Few-Shot Prompting Cannot Fix Fundamental Knowledge Gaps}
\label{sec:icl_failure}

\begin{figure*}[t]
    \centering

    \begin{subfigure}{0.48\textwidth}
        \centering
        \includegraphics[width=\linewidth]{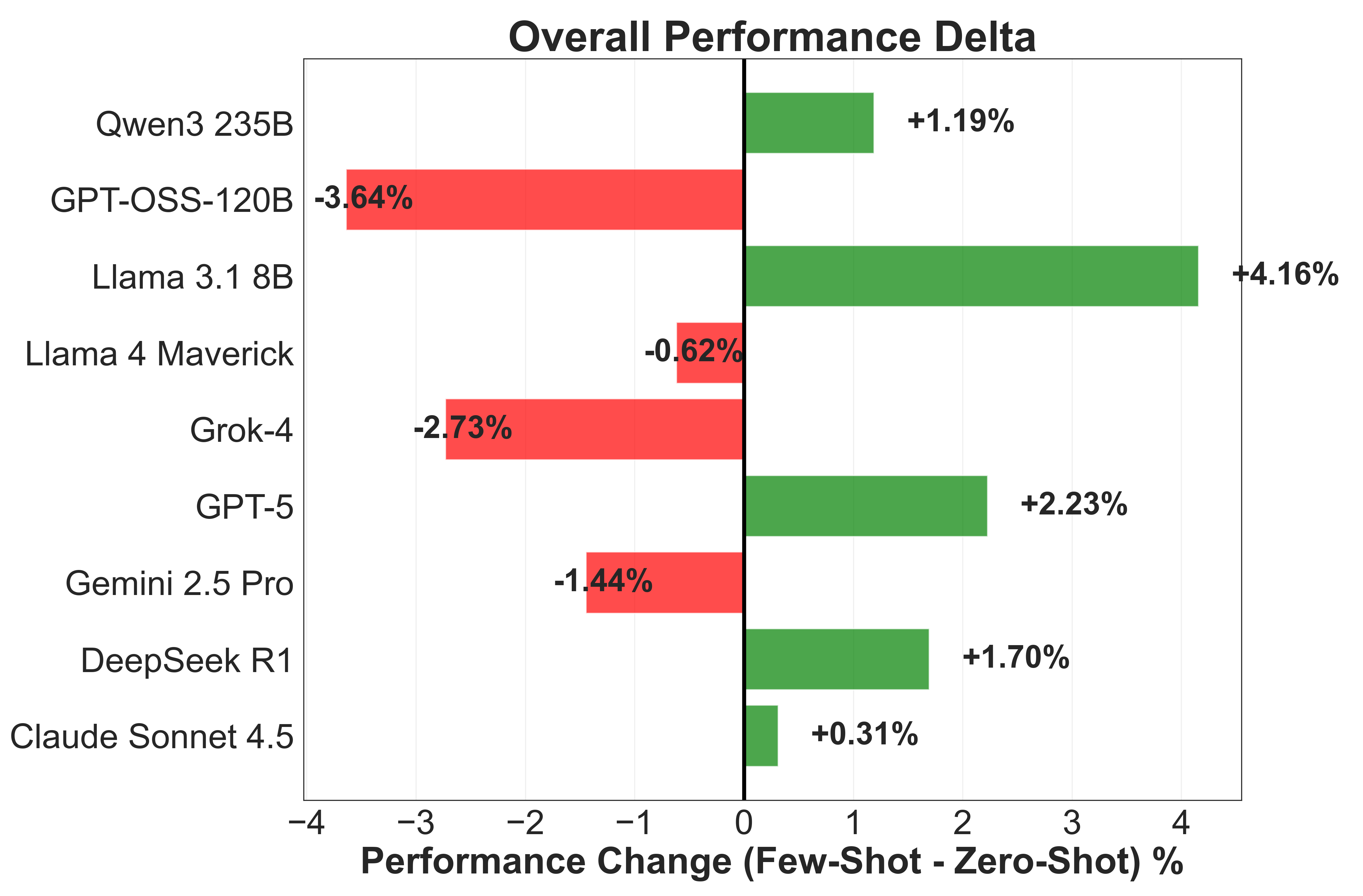}
        \caption{Overall performance delta (= few-shot - zero-shot), showing minimal aggregate benefit with only two models exceeding meaningful improvement ($>$2\%).}
        \label{fig:icl_left}
    \end{subfigure}
    \hfill
    \begin{subfigure}{0.48\textwidth}
        \centering
        \includegraphics[width=\linewidth]{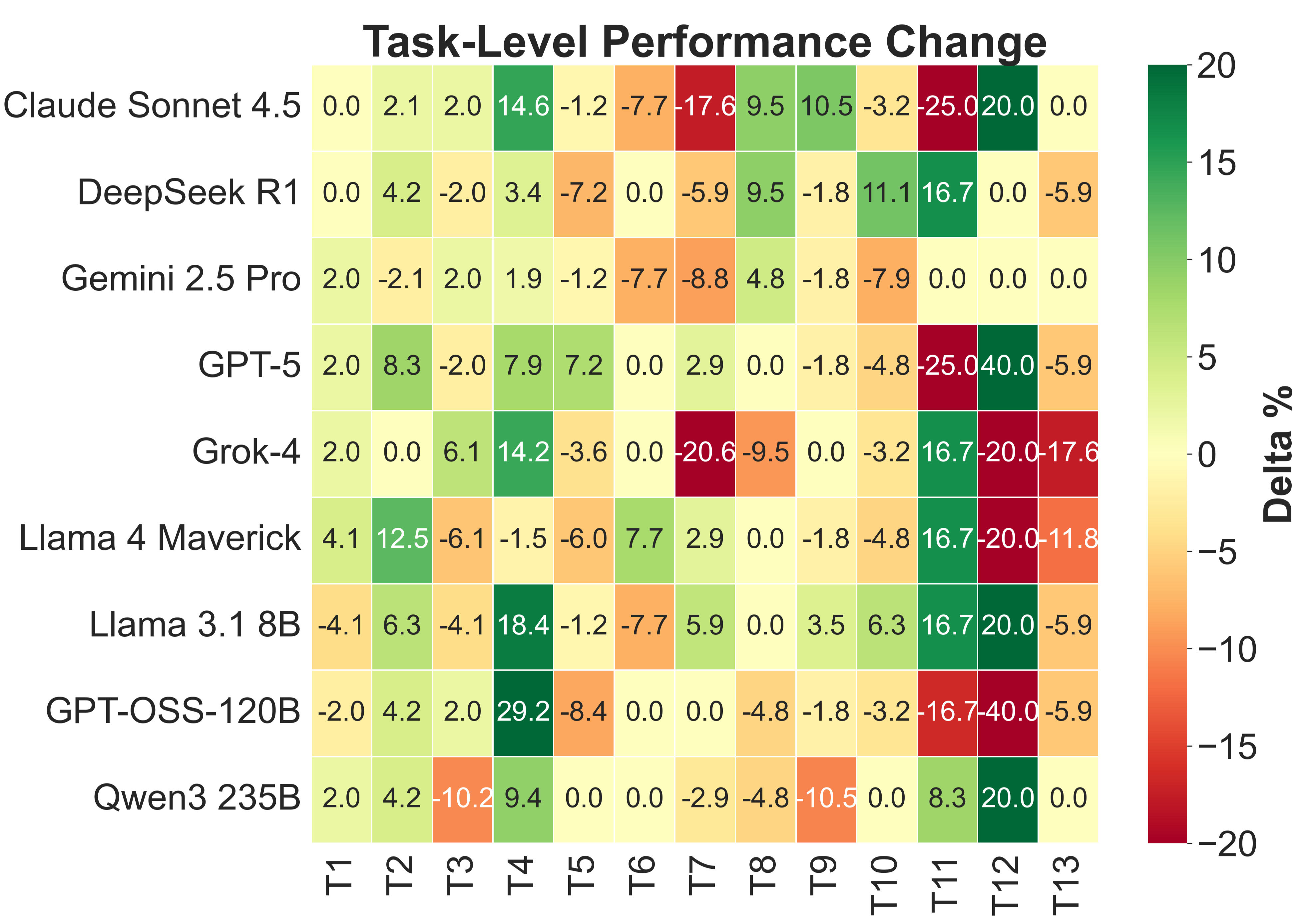}
        \caption{Task-level delta heatmap revealing inconsistent ICL effects across tasks, with improvements on some tasks offset by degradations on others.}
        \label{fig:icl_right}
    \end{subfigure}

    \vspace{0.5em}

    \begin{subfigure}{0.48\textwidth}
        \centering
        \includegraphics[width=\linewidth]{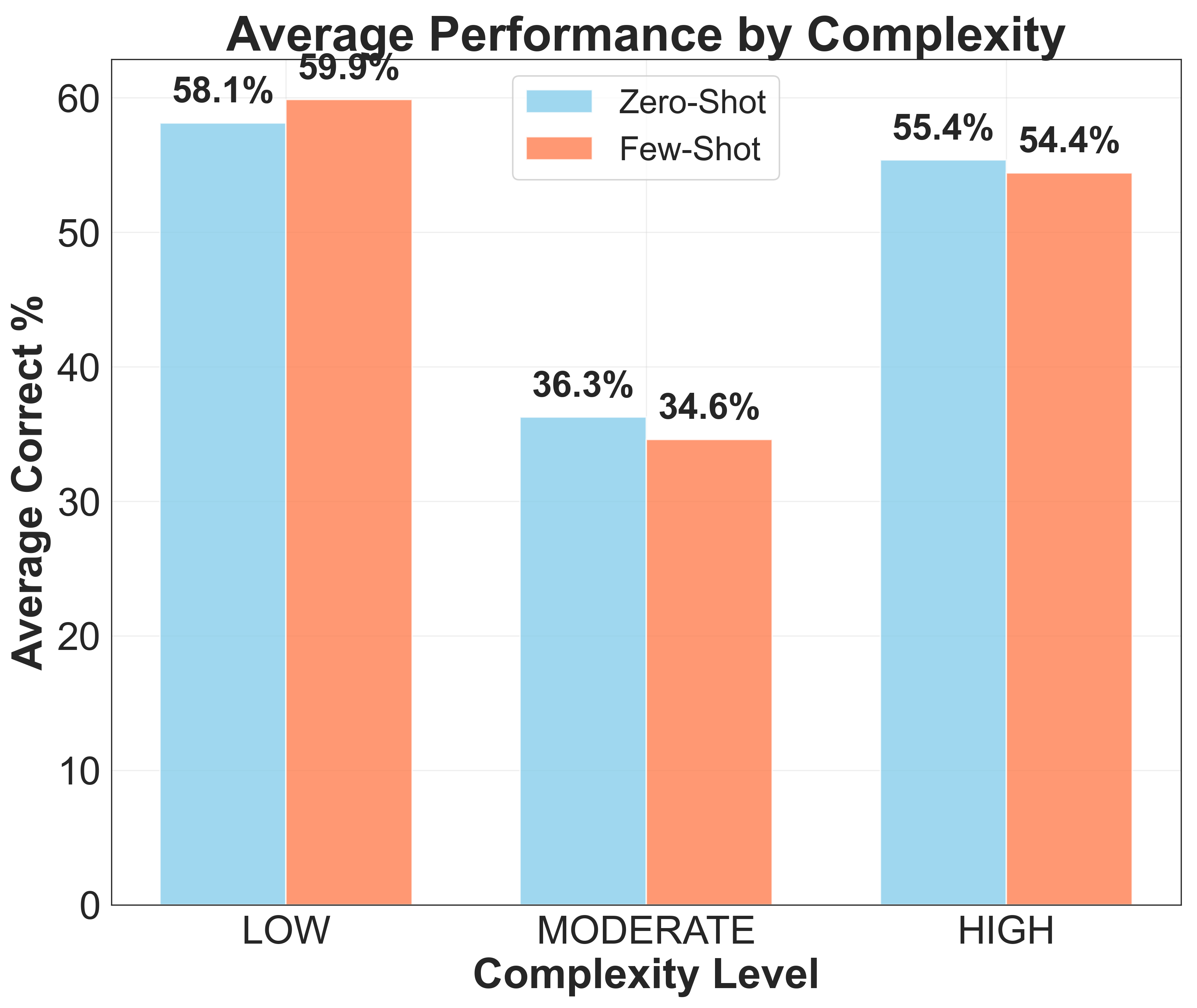}
        \caption{Complexity-wise comparison showing negligible average benefit across Low, Moderate, and High complexity categories.}
        \label{fig:icl_bottom_left}
    \end{subfigure}
    \hfill
    \begin{subfigure}{0.48\textwidth}
        \centering
        \includegraphics[width=\linewidth]{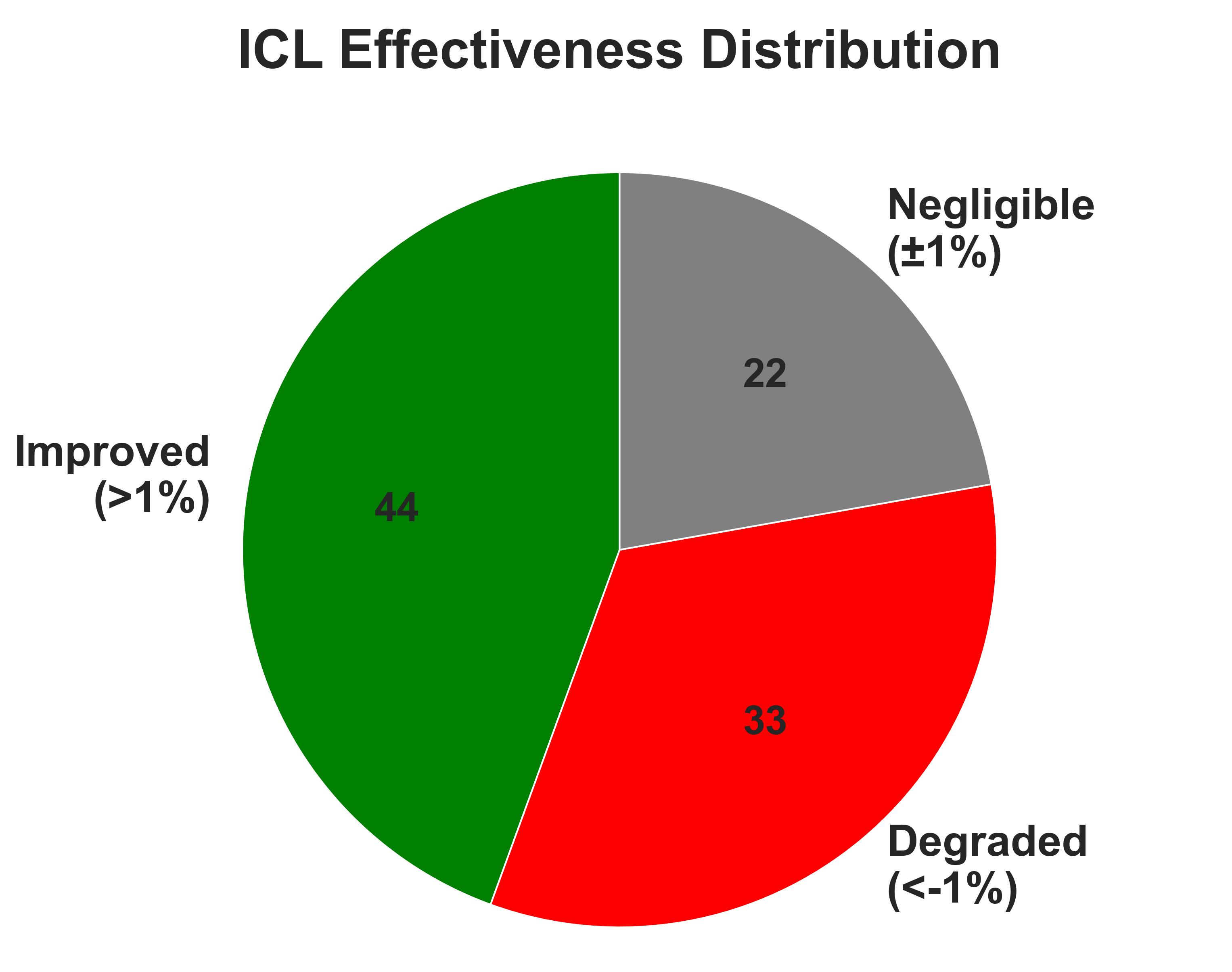}
        \caption{Effectiveness distribution: 5 of 9 models fall within ±1\%, and two degrade substantially, indicating insufficient reasoning scaffolding.}
        \label{fig:icl_bottom_right}
    \end{subfigure}

    \caption{\textbf{In-Context Learning (ICL) Effectiveness Analysis.}
Few-shot prompting provides negligible benefit for Islamic legal reasoning: improvements are sparse, task-dependent, and offset by frequent degradations. Panels (a)--(d) present model-level gains, task variability, complexity sensitivity, and effect distributions, respectively.}
    \label{fig:icl_analysis}
\end{figure*}

A central finding concerns the limited effectiveness of in-context learning (ICL) for Islamic jurisprudence reasoning. Figure~\ref{fig:icl_analysis} presents a comprehensive analysis comparing zero-shot and few-shot performance, revealing systematic ICL ineffectiveness.

The results reveal striking ICL ineffectiveness. Only two models demonstrate meaningful improvement with few-shot examples: GPT-5 (+2.23\%) and Qwen3 235B (+1.19\%). Five models show negligible changes within ±1\%: Claude Sonnet 4.5, Gemini 2.5 Pro, DeepSeek R1, Llama 4 Maverick, and GPT-OSS-120B. Most concerningly, two models degrade substantially: Grok-4 (-2.73\%) and Llama 3.1 8B (which improves +4.16\% but remains at a much lower 30.96\% performance). These patterns indicate that few-shot examples provide minimal systematic benefit for Islamic jurisprudence reasoning.

Task-level analysis exposes the mechanisms underlying ICL failure. The delta heatmap reveals substantial performance variance across tasks for individual models: systems demonstrating improvement on some tasks simultaneously exhibit degradation on others. For instance, Claude Sonnet 4.5 improves on T2 and T4 while degrading on T7 and T11 by a significant difference. This inconsistency indicates that few-shot examples enable task-specific pattern matching rather than systematic reasoning enhancement, undermining confidence in ICL-based approaches for reliable Islamic legal reasoning.

Complexity-wise analysis confirms ICL benefits do not systematically scale with task difficulty. Average improvements remain minimal across complexity levels (Low: +1.65\%, Moderate: -0.89\%, High: -1.05\%), with moderate and high-complexity tasks actually showing slight average degradation. This pattern reveals a profound insight: moderate-complexity tasks require exact knowledge from Islamic legal texts. These tasks demand 100\% accurate recall of specific conditions, precise doctrinal distinctions, or faithful synthesis of statutory articles. A few examples cannot teach models knowledge they never acquired during different phases of training. Conversely, high-complexity tasks (T10-T13: \`{}illah identification, analogical reasoning, cross-madhhab synthesis, maxim mapping) permit flexible reasoning and verbose explanation, allowing models to leverage general reasoning capabilities without requiring memorized legal provisions and committing to a specific take. Models can discuss legal principles discursively without knowing precise textual requirements.

The FPQ degradation pattern provides particularly more compelling evidence of ICL failure. As noted in Section~\ref{sec:fpq_analysis}, average FPQ acceptance increased from 37.25\% (zero-shot) to 40.74\% (few-shot), a 3.49\% degradation. This counterintuitive result indicates that few-shot examples may actually reinforce sycophantic behaviors, encouraging models to accept user premises rather than critically evaluate them.

These findings strongly suggest that in-context learning provides insufficient reasoning enhancement for Islamic jurisprudence tasks. The fundamental limitation appears to be the absence of foundational knowledge: few-shot examples cannot compensate for missing Islamic legal knowledge in pre-training corpora. Models lacking grounding in Islamic law, Hadith traditions, madhab methodologies, and jurisprudential principles resort to superficial pattern matching when presented with few-shot examples. This pattern matching fails for moderate-complexity tasks (T7-T9) requiring verbatim recall of legal conditions and statutory provisions, while producing plausible but potentially incorrect responses for high-complexity tasks (T10-T13) that reward verbose reasoning over precision.

This phenomenon also occurs because LLMs are optimized for semantically coherent text continuation rather than exact factual retrieval. Their distributed, meaning-based internal representations lead to fabricated but plausible completions through semantic interpolation when tasks require the structured enumeration of conditions. Conversely, high-complexity juridical reasoning tasks activate deeply learned logical patterns from post-training (SFT, RLHF), allowing models to project competence on \`{}illah identification and maxim mapping despite lacking precise textual knowledge. This negative result carries significant implications: prompt engineering cannot substitute for absent training data. The community requires dedicated efforts to develop models trained on comprehensive corpora spanning 1200+ years (8th-20th century), including classical Hadith collections, madhab-specific jurisprudential works, codified legal compendia, and contemporary databases. Only through such specialized training can models acquire the foundational knowledge necessary for reliable Islamic legal reasoning and advisory.

\subsection{Closed-Source versus Open-Source: Quantifying the Performance Gap}
\label{sec:closed_vs_open}

\begin{figure}[t]
    \centering
    \includegraphics[width=\columnwidth]{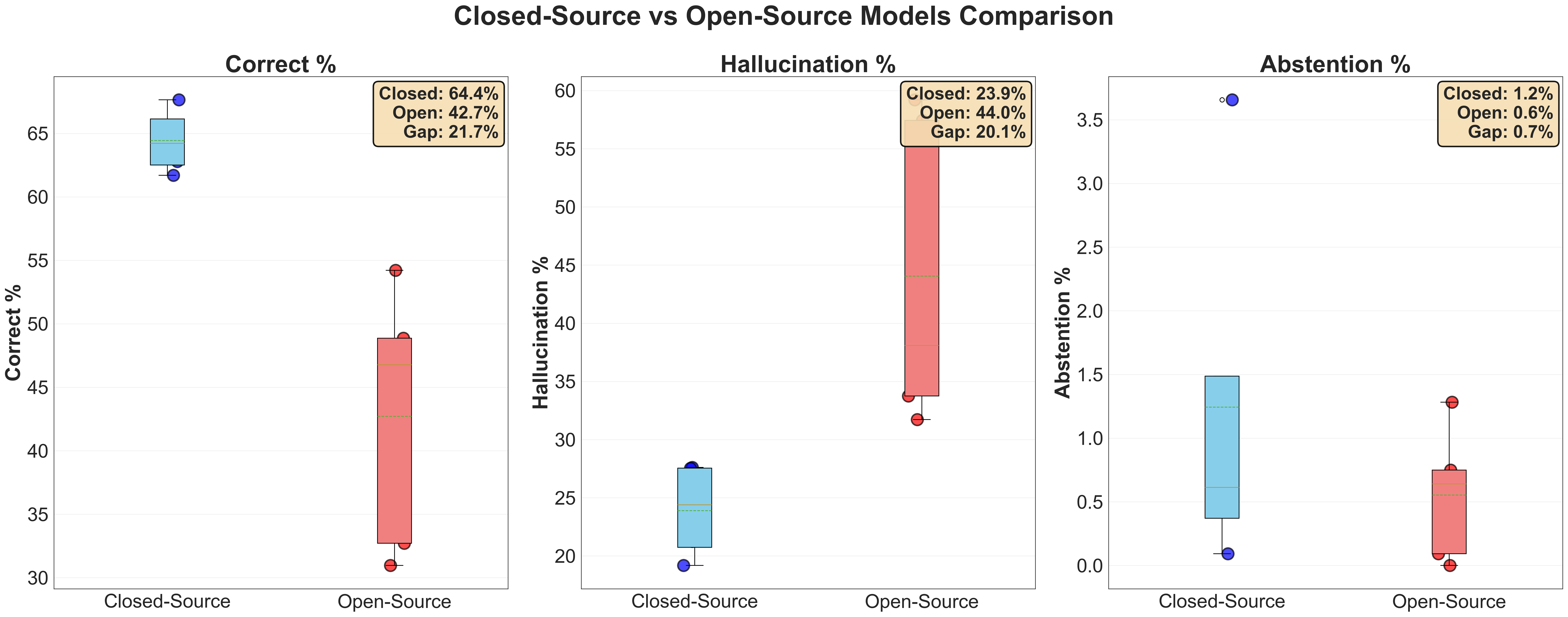}
    \caption{Comparative analysis of closed-source versus open-source model performance under few-shot prompting across correctness, hallucination, and abstention metrics. Box plots display median (solid line), mean (dashed line), and individual model performance (scatter points). Closed-source models (blue) substantially outperform open-source models (red) on correctness while maintaining lower hallucination rates. The 19.76 percentage point correctness gap and 13.32 percentage point hallucination gap underscore substantial advantages of proprietary systems for Islamic jurisprudence reasoning.}
    \label{fig:closed_vs_open}
\end{figure}

Figure~\ref{fig:closed_vs_open} quantifies the substantial performance gap between closed-source and open-source models across three key metrics: correctness, hallucination, and abstention.

Closed-source models and open-source models have a gap of 19.76\% in terms of correctness. This substantial gap persists across the performance distribution: all four closed-source models exceed the open-source average, with even the weakest closed-source system (Grok-4) outperforming three of five open-source alternatives. Only DeepSeek R1 approaches competitive performance among open-source models.

Hallucination analysis reveals even bigger gaps. Closed-source models show an average of 22.29\% hallucination compared to 35.61\% for open-source systems, a 13.32\% gap. The disparity becomes most pronounced at the extremes: while the best closed-source model (Claude) and best open-source model (DeepSeek R1) both maintain sub-40\% rates, the lowest performing open-source models (GPT-OSS-120B, Llama 3.1 8B) exhibit hallucination exceeding 50\%. Abstention patterns reveal marginally better uncertainty calibration in open-source models (4.38\% average) versus closed-source systems (3.37\% average), though this 1.01\% difference is modest. However, this slight advantage is overwhelmed by substantially higher hallucination rates, indicating open-source models abstain too infrequently given their elevated error propensity. Proper calibration would demand substantially higher abstention rates for models achieving only 30-55\% correctness while hallucinating at 30-60\% rates.

These gaps likely reflect multiple factors: (1) larger model scale and greater training compute for closed-source systems, (2) more extensive safety fine-tuning and reinforcement learning from human feedback procedures, (3) proprietary training data potentially including Islamic legal texts absent from open-source training sets, and (4) more sophisticated inference-time techniques including chain-of-thought prompting, self-consistency sampling, and answer verification. The current performance gap creates concerning access inequities: only well-resourced organizations capable of affording closed-source API access can deploy models with acceptable Islamic jurisprudence reasoning capabilities, while open-source alternatives exhibit inadequate performance profiles.


\section{Discussion}
\label{sec:discussion}

\subsection{The Mid-Complexity Failure Zone}
\label{subsec:uncanny_valley}

\begin{figure}[h]
    \centering
    \includegraphics[width=0.85\columnwidth]{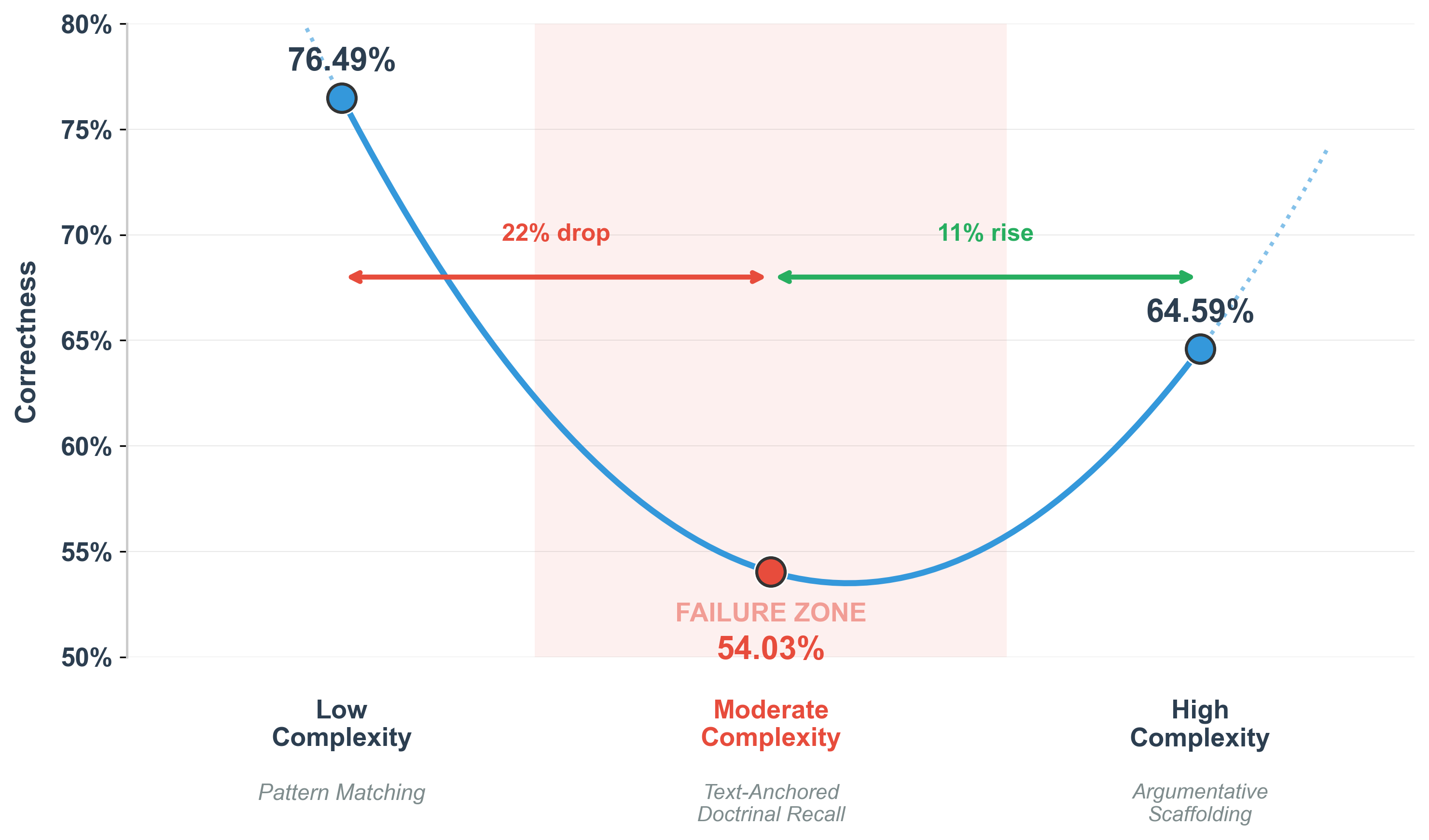}
    \caption{The Mid-Complexity Failure Zone: Performance pattern for GPT-5, the top-performing model, across complexity tiers under few-shot prompting. The model exhibits a clear performance trough: 76.49\% on low-complexity tasks (pattern matching), plummeting to 54.03\% on moderate-complexity tasks requiring specific text-anchored doctrinal recall (the failure zone), and recovering to 64.59\% on high-complexity tasks relying on generalized argumentative scaffolding. The shaded region highlights the dangerous zone where even state-of-the-art models regularly fail yet produce authoritative-seeming outputs.}
    \label{fig:mid_complexity_failure}
\end{figure}

Our results reveal a paradoxical non-monotonic performance \textbf{Mid-Complexity Failure Zone} pattern: LLM accuracy does not change monotonically with task complexity. Instead, we observe elevated correct answer rates on low-complexity tasks (e.g., Task 4: 79.03-97.75\% across models), a drop on moderate-complexity tasks (e.g., Task 7: 20.59-61.76\% across models), and paradoxically improved performance on high-complexity tasks (e.g., Task 13: 47.06-82.35\% across models). Even GPT-5, the top-performing model, exemplifies this pattern: achieving 76.49\% average correctness on low-complexity tasks, dropping to 54.03\% on moderate-complexity tasks, and recovering to 64.59\% on high-complexity tasks. There is a clear U-shaped performance trough where moderate-complexity performance falls 22 percentage points below low-complexity performance despite the model's overall superiority. Figure~\ref{fig:mid_complexity_failure} illustrates this phenomenon. 

A structurally similar pattern has been described in human--robot interaction research as the \textit{``uncanny valley''}, where increasing human-likeness generally improves user comfort \textit{until a sudden non-monotonic dip} in the middle, after which trust rises again at full realism \cite{mori2012uncanny}. The danger of this middle trough lies in its \textit{misleading plausibility}: the agent appears convincingly human while still behaving in unreliable or unsettling ways. By analogy, our findings suggest a comparable \textit{reasoning uncanny valley}, in which LLMs seem most competent at moderate complexity precisely when they are most likely to be incorrect, creating a deceptively high-risk failure zone.

This Mid-Complexity Failure Zone carries significant practical risks. Moderate-complexity responses appear authoritative, employ correct terminology, cite plausible legal frameworks, and demonstrate apparent sophistication. Unlike low-complexity queries, where factual errors may be readily detected, or high-complexity queries recognized as speculative, moderate-complexity responses present precise-seeming answers to questions with objectively correct answers grounded in specific texts.

\textbf{Legal Dual-Process Theories and Cognitive Misalignment.} This failure pattern finds theoretical grounding in legal dual-process theories, which propose that legal reasoning emerges from two distinct cognitive systems \cite{Ronkainen2011}. \textit{Type 1 processing} operates through fast, intuitive, gist-based reasoning and narrative pattern recognition. \textit{Type 2 processing} enables slow, deliberate, detail-sensitive doctrinal application requiring careful textual analysis. Legal scholars have argued that tasks requiring simultaneous engagement of both systems, which recall specific textual rules (Type 2) while constructing structured legal arguments (Type 1), pose particular cognitive challenges and may increase the likelihood of errors \cite{Ronkainen2011}. Our mid-complexity tasks occupy precisely this zone, requiring text-anchored doctrinal recall alongside coherent reasoning frameworks, creating conditions under which performance reliably degrades.

The mechanisms underlying this failure zone illuminate fundamental architectural characteristics of LLMs and reveal why human complexity classifications systematically fail to predict machine difficulty. Our complexity classifications were assigned by Islamic law experts based on the cognitive demands these tasks place on \textbf{human scholars} trained in traditional jurisprudential methodology. However, this human-centric difficulty hierarchy inverts for LLMs due to fundamental architectural differences in information processing. This cognitive gap connects to Kahneman's dual-process model of reasoning \cite{kahneman2011thinking}, which distinguishes between \textit{System 1} (fast, automatic, pattern-based processing) and \textit{System 2} (slow, deliberate, analytical reasoning). This framework has been extended to deep learning contexts by Bengio \cite{Bengio2022NeurIPSslides}. 

Islamic legal reasoning (\textit{ijtihād}) is grounded in a disciplined methodology that integrates textual interpretation, evaluation of evidentiary hierarchies, and rule-governed argumentation to derive rulings from the Qur'an, Sunnah, consensus, and analogical reasoning \cite{hallaq1997history}. Human scholars therefore approach this process as a quintessential \textit{System~2} task, demanding slow, effortful, and principled deliberation rather than intuitive or heuristic judgment \cite{kahneman2011thinking}. LLMs, however, operate entirely as an amplified form of System 1: large-scale statistical pattern recognition running through opaque architectures. LLMs cannot verify factual claims or recognize knowledge gaps for sources never encountered in training data. Instead, they approximate reasoning through verbose pattern generation, producing outputs that \textit{linguistically resemble} analytical thought without actual grounding.  This creates a dangerous mismatch: moderate-complexity Islamic legal tasks demand both Type 1 reasoning (constructing coherent juridical arguments) and Type 2 precision (recalling exact textual specifications), yet LLMs can only provide the former through statistical patterns while fabricating the latter through semantic interpolation. 

This also raises a broader concern: users without legal training lack the evaluative capacity to distinguish linguistically polished output from substantively accurate reasoning. Such users are therefore especially vulnerable to being misled by fluent but incorrect responses---a risk noted in prior legal-AI evaluations \cite{Schweitzer2025}. A parallel issue was observed in software communities: StackOverflow prohibited generative-AI answers in 2022 because their apparent ``quality" frequently concealed errors \cite{StackOverflowPolicy2022}.

This architectural difference creates systematic performance inversions across complexity tiers. Low-complexity tasks require only surface-level pattern matching (79.03-97.75\% accuracy). High-complexity tasks (\textit{`illah} identification, maxim mapping, cross-school synthesis) perform comparatively better (47.06-82.35\%) because models rely on generalized argumentative scaffolding and can maintain abstract reasoning without committing to precise doctrinal facts. Moderate-complexity tasks occupy the failure zone because they demand what LLMs lack: specific text-anchored doctrinal recall. Tasks requiring condition enumeration (T7), comparative distinctions (T8), or statute-like synthesis (T9) demand factual precision with single correct answers determinable only through direct textual access. LLMs instead produce \textit{semantic interpolation}, plausible completions maintaining discourse coherence without factual grounding. This structural limitation produces elevated hallucination rates.

\begin{figure}[h]
    \centering
    \includegraphics[width=1\linewidth]{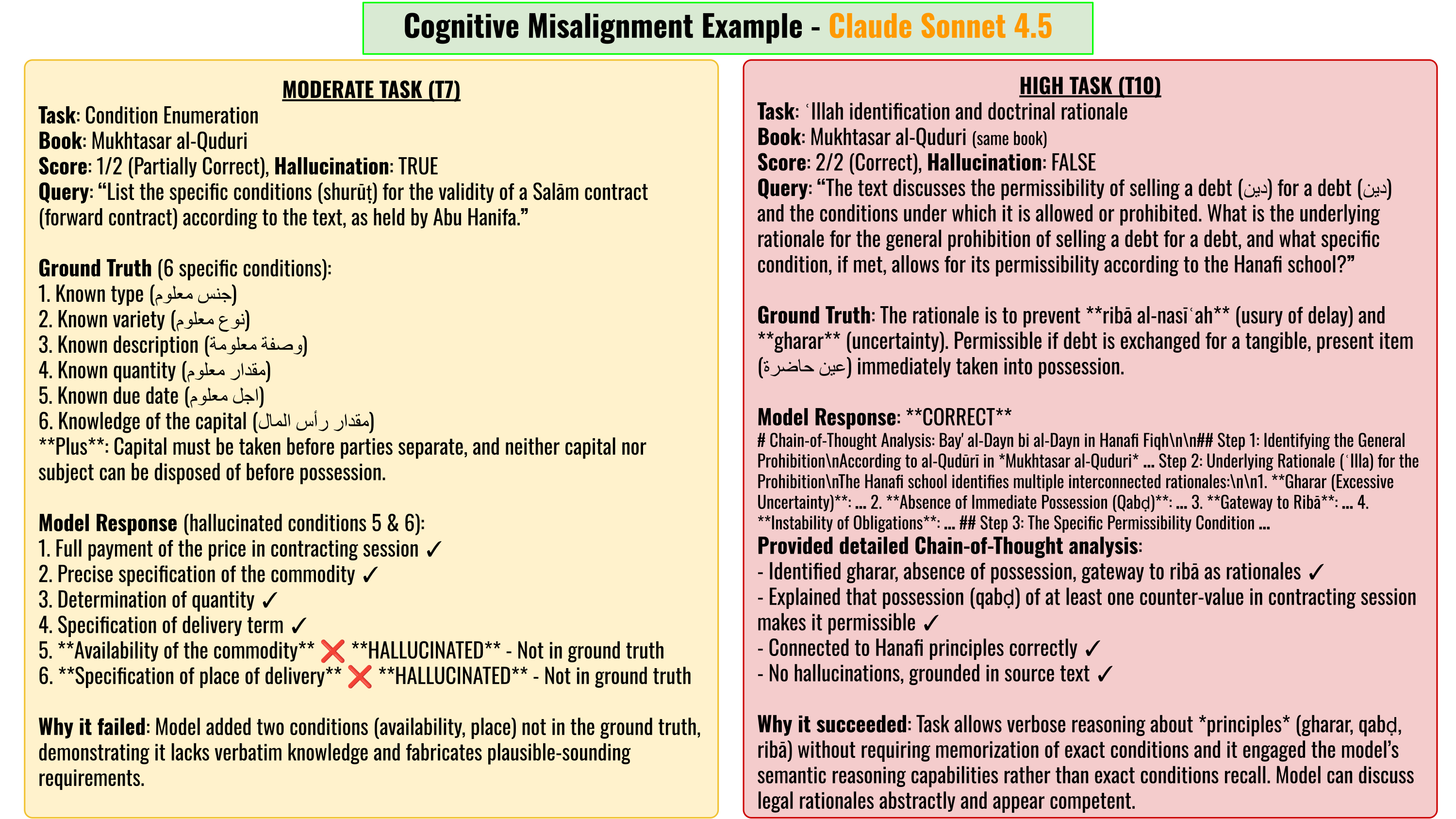}
    \caption{Cognitive misalignment illustrated: Claude Sonnet 4.5 performance on two tasks from \textit{Mukhtaṣar al-Qudūrī}. (Left) Moderate-complexity T7 (condition enumeration for \textit{salām} contracts) results in failure with two fabricated conditions absent from Abū \d{H}anīfa's actual six-condition formulation. (Right) High-complexity T10 (identifying \textit{`illah} for debt-for-debt prohibition) from the same book succeeds completely with correct analysis of \textit{ribā al-nasī'ah}, \textit{gharar}, and possession requirements.}
    \label{fig:cognitive_misalignment}
\end{figure}

One example of Claude Sonnet 4.5's performance on \textit{Mukhtaṣar al-Qudūrī} illustrates this paradox (Figure~\ref{fig:cognitive_misalignment}). On moderate-complexity T7 (enumerating \textit{salām} contract conditions), Claude \textbf{failed}, fabricating ``availability of commodity" and ``specification of place of delivery". These are absent from Abū \d{H}anīfa's actual six-condition formulation. Yet on high-complexity T10 from the \textit{same book} (identifying \textit{`illah} for debt-for-debt prohibition), Claude \textbf{succeeded} completely, providing correct analysis of \textit{ribā al-nasī'ah}, \textit{gharar}, and possession requirements without hallucination. The difference lies in task structure: abstract legal \textit{principles} activate semantic reasoning circuits leveraging deeply learned patterns; enumerating specific \textit{conditions} triggers semantic interpolation yielding fabricated yet authoritative-sounding content. This pattern replicates systematically across approximately 9,605 instances. A single model might fail on enumerating \textit{salam} contract conditions (T7), while succeeding on multiple high-complexity queries involving salam contracts (T10-T13), generating multiple documented paradox instances from the same underlying knowledge gap for all 9 LLMs: models failed moderate-complexity verbatim retrieval (58-97\% failure rates) yet succeeded on thematically similar high-complexity reasoning from identical source texts (38-70\% success rates), confirming that verbose reasoning masks absent foundational knowledge.

\textbf{The Illusion of Knowledge and Dangerous Misplaced Confidence.} This failure mode aligns with established cognitive biases. The \textit{Dunning-Kruger effect} \cite{dunningkruger} demonstrates that individuals with low competence systematically overestimate their abilities, lacking metacognitive capacity to recognize incompetence. Similarly, Kahneman's \textit{Overconfidence effect} \cite{kahneman2011thinking,dobelli2013art} describes how individuals overestimate judgment accuracy when operating with incomplete information. We observe a structurally analogous \textit{Illusion of Knowledge} in LLMs: models generate confident, technically sophisticated responses about Islamic legal conditions never encountered in training data, exhibiting what Dobelli terms \textit{chauffeur knowledge} \cite{dobelli2013art}, surface-level fluency mimicking expertise without underlying comprehension. Like a chauffeur who memorizes a physicist's lecture but cannot answer deeper questions, LLMs produce linguistically sophisticated outputs that create an appearance of expertise without factual grounding. 

This manifests empirically through near-zero abstention rates ($<$1\%) despite 30-60\% hallucination rates. A model confidently enumerating seven \textit{nikāḥ} conditions when texts specify five appears knowledgeable while fundamentally misrepresenting Islamic legal requirements. Our FPQ results (Section~\ref{sec:fpq_analysis}) reinforce this pattern: models accept 40-86\% of deliberately incorrect premises, building elaborate justifications rather than critically evaluating validity. When asked to analyze the ``major Maliki work \textit{Badā'` al-\d{S}anā'i`}" (actually Hanafi), weaker models generate Maliki analyses, sycophantically accepting false premises and generating responses consistent with query framing rather than retrieving contradictory knowledge. 
This manifests empirically through near-zero abstention rates ($<$1\%) despite 30-60\% hallucination rates. A model confidently enumerating seven \textit{nikāḥ} conditions when texts specify five appears knowledgeable while fundamentally misrepresenting Islamic legal requirements. Our FPQ results (Section~\ref{sec:fpq_analysis}) reinforce this pattern: models accept 40-86\% of deliberately incorrect premises, building elaborate justifications rather than critically evaluating validity. When asked to analyze the ``major Maliki work \textit{Badā'` al-\d{S}anā'i`}" (actually Hanafi), weaker models generate Maliki analyses, sycophantically accepting false premises and generating responses consistent with query framing rather than retrieving contradictory knowledge. 

\textbf{Implications.} The ICL failure (Section~\ref{sec:icl_failure}) amplifies this pattern: few-shot prompting provides negligible benefit (only 2 of 9 models improve $>$1\%) and worsens FPQ acceptance by 3.49\%, reinforcing sycophantic behaviors rather than enhancing critical reasoning. The failure zone cannot be traversed through prompting. Few-shot examples cannot teach content from classical fiqh texts never encountered during pre-training. First, aggregate performance metrics obscure critical failure modes. A model achieving 55\% overall correctness appears marginally useful, yet if performance concentrates on trivial queries and verbose speculation while failing on precise-knowledge tasks, the metric severely misrepresents deployment readiness. Second, the most high-risk deployment scenarios involve precisely those moderate-complexity queries where users receive authoritative-seeming answers grounded in hallucination rather than authentic knowledge. Third, overcoming the mid-complexity failure zone requires comprehensive domain-specific pre-training on classical Hadith collections, \textit{madhab}-specific jurisprudential works, and codified Islamic legal works---not post-hoc prompting or instruction tuning interventions.


\subsection{Alignment with Broader LLM Reasoning Limitations} \label{subsec:discussion_alignment}

The ``mid-complexity failure zone" we observe validates established research that LLMs prioritize surface-level pattern matching over genuine reasoning. Models' high degree of failure on precise enumeration tasks mirrors the fragility documented by \cite{mirzadeh2024gsmsymbolic} and \cite{nezhurina2024alice}, where architectural limitations lead to high-confidence confabulation on logically simple problems. This sensitivity to task structure rather than semantic complexity aligns with Roh et al.~\cite{chain2025code} and Chatziveroglou et al~\cite{chatziveroglou2025exploring}, who found that minor formulation changes trigger reasoning breakdowns. However, our findings extend beyond the directional retrieval asymmetry of the Reversal Curse \cite{berglund2024reversal}; we identify a fundamental ``correlation vacuum'' where the absence of classical \textit{fiqh} texts in pre-training forces models to rely on semantic interpolation to mask knowledge voids.

Mechanistically, this behavior confirms that models default to spurious correlations and shortcuts as described by \cite{enstrom2024reasoning}, \cite{explore2024spurious}, and \cite{navigating2024shortcuts}. The sycophancy observed in our FPQs acceptance further instantiates the entailment bias noted by \cite{shortcut2024emnlp}. Crucially, while standard spurious correlations may be addressable through robustness interventions \cite{mitigating2024capt, assessing2025spurious}, the ICL failure demonstrates that post-hoc prompting cannot remedy these specific deficits. Unlike reasoning errors, the absence of domain knowledge is not a retrieval failure but a representational gap, proving that reliability in high-stakes Islamic law requires comprehensive pre-training on authoritative sources rather than prompt engineering.


\section{Limitations and Future Directions}
\label{sec:limitations}

Our work establishes the first comprehensive evaluation framework for Islamic legal reasoning in LLMs, opening multiple promising directions for future research and development.

\paragraph{Multilingual Evaluation Opportunities.} While our benchmark provides a comprehensive English-language evaluation, extending to Arabic and other languages spoken across Muslim-majority regions represents a valuable direction for future work. Given that classical Islamic legal texts were originally composed dominantly in Arabic, bilingual evaluation would enable important investigations: whether models exhibit differential Islamic legal knowledge depending on query language, whether translation introduces systematic artifacts, and whether models trained predominantly on English Islamic texts demonstrate gaps in classical Arabic legal terminology.

\paragraph{Internal Knowledge \& Retrieval Augmentation:} Our study offers a comprehensive LLM evaluation for Islamic legal domains through internal knowledge assessment, or \textbf{knowledge-based benchmarking}. We test models' ability to reason about Islamic law using only their pre-trained parameters and in-context few-shot learning, without external retrieval, knowledge bases, or tool augmentation. This directly addresses whether current foundation models contain sufficient Islamic legal knowledge in their training data. Our results show they do not, and that prompting techniques cannot compensate for this absence.

Contemporary LLM applications frequently employ Retrieval-Augmented Generation (RAG) \cite{shuster2021retrieval, cui2023chatlaw, peng2023check, savelka2023explaining}, specialized fine-tuning \cite{tian2023fine, razumovskaia2024dial, zhang2023r}, factuality-oriented decoding \cite{shi2024trusting, mallen2022not, li2023inference, chuang2023dola}, and external database checks \cite{chern2023factool, peng2023check, qin2024tool, gou2023critic} to augment base models with external knowledge. Evaluating such enhanced systems is part of comprehensive benchmarking and is our immediate next research direction. In Western legal contexts, \cite{Dahl_second} shows that RAG-based tools such as Lexis+ AI and Westlaw AI-Assisted Research markedly outperform base models yet still exhibit hallucination rates of 17–33\%. Analogous systematic evaluation is urgently needed for Islamic legal applications that couple LLMs with retrieval over Islamic legal databases, Hadith collections, and fatwa repositories. However, specialized Islamic RAG-enabled platforms and LLMs have emerged, such as Usul.AI \cite{usul_ai} and Fanar \cite{fanarpaper}, and their performance will be evaluated in future work.

LLMs must ultimately excel at both internal knowledge and external augmentation to be safely integrated into Islamic legal advisory systems. As our results demonstrate, critical gaps in background Islamic legal knowledge preclude the use of LLMs as reliable sources of religious counsel, regardless of advances in retrieval or tool use. Deficient base-model knowledge introduces structural vulnerabilities: models cannot reliably formulate retrieval queries, assess the relevance and authenticity of retrieved texts, detect contradictions between sources, or distinguish authoritative rulings from minority or superseded positions. 

While retrieval-based and related methods have shown some reduction in hallucinations and improved accuracy, reliability, and faithfulness in general domains \cite{chen2024benchmarking, siriwardhana2023improving, ram2023context, cheng2023lift}, they face specific limitations in Islamic law. Their effectiveness hinges on retrieval quality \cite{wu2024well} and on correctly parsing often ambiguous, vernacular user queries rather than technical legal formulations \cite{tonmoy2024comprehensive}. Retrieving precise information from extensive Islamic corpora spanning over 1,200 years, seven schools of jurisprudence, and multiple subdomains is computationally demanding and requires continual updates to incorporate contemporary and evolving interpretations \cite{siriwardhana2023improving}. Knowledge bases may also contain conflicting information \cite{wang2023survey, yu2023chain, gao2023retrieval}: for instance, when classical rulings are superseded by contemporary consensus, when schools hold contradictory positions on an issue, or when the authority of a source depends on its chain of transmission (\textit{isnād}). In such cases, the retrieval module and the model must identify outdated, jurisdictionally irrelevant, or methodologically inappropriate sources and privilege binding, apposite rulings, a task that is impossible without robust foundational Islamic legal knowledge.

\paragraph{Advancing Automated Evaluation Methodology.} Our LLM-as-a-Judge framework using o3 enables scalable assessment and shows strong correlation with expert annotations, yet automated evaluation in specialized legal domains remains imperfect. Evaluation metrics may not fully capture real-world reliability \cite{ji2023survey, zhang2025siren, lucas2023fighting}. Our focus on published classical texts provides rigorous grounding, though future work could incorporate oral traditions and contemporary interpretive developments. Our hallucination metrics offer a useful signal but may reflect dataset biases or lack sufficient comprehensiveness, nuance, or task-specificity \cite{kang2024comparing}, so it remains unclear whether technical advances can fully resolve hallucination. Our binary hallucination flagging (True/False) is efficient, but finer-grained taxonomies (omission vs. commission, minor vs. major distortion, plausible vs. implausible fabrication) and their differential risks for Islamic legal advisory systems warrant study. Although we validated ground truth answers with experts, jurisprudential pluralism means some questions legitimately admit multiple answers across schools, periods, and methodologies. Our framework partially accommodates such variation, providing a basis for future work on more sophisticated automated evaluation in a pluralistic domain like Islamic jurisprudence.

\paragraph{Model Coverage and Emerging Systems.} We evaluated nine state-of-the-art LLMs representing both closed-source (GPT-5, Claude Sonnet 4.5, Gemini 2.5 Pro, Grok-4) and open-source (DeepSeek R1, Llama 4 Maverick, Llama 3.1 8B, GPT-OSS-120B, Qwen3 235B) systems available at the time of benchmarking (mid 2025 to late 2025). The rapid pace of LLM development means newer models with potentially improved Islamic legal capabilities will continue to emerge, and our benchmark provides a standardized framework for evaluating such systems as they become available. Our findings characterize current system capabilities rather than fundamental architectural limits. Moreover, our focus on general-purpose foundation models that Muslims increasingly consult for fatwas and religious rulings provides a crucial baseline assessment. Future work should systematically compare these general-purpose models against domain-specialized Islamic AI systems or fine-tuned models developed specifically for religious guidance, determining whether targeted training on Islamic corpora can bridge the knowledge gaps our evaluation reveals.

These limitations notwithstanding, \texttt{IslamicLegalBench} provides a foundational evaluation framework for Islamic legal reasoning in LLMs, revealing critical knowledge gaps and failure modes with profound implications for Muslim communities worldwide, who are increasingly relying on AI systems for religious guidance. Our findings establish the important need for training foundation models on comprehensive Islamic legal corpora and developing rigorous evaluation methodologies before deploying such systems in sensitive religious advisory contexts.


\section{Conclusion}
\label{sec:conclusion}

As millions of Muslims increasingly consult large language models (LLMs) for religious and legal guidance, we ask a fundamental question: can AI systems trained predominantly on Western texts reliably reason about Islamic jurisprudence? Evaluating nine state-of-the-art LLMs, we find that current foundation models lack the adequate Islamic legal knowledge required for reliable deployment, and that standard prompting cannot compensate. We introduce \textbf{\texttt{IslamicLegalBench}}, the first comprehensive benchmark covering the full spectrum of Islamic legal reasoning. Built from 38 foundational texts spanning 1,200 years and seven schools of jurisprudence, it comprises 718 evaluation instances across 13 task types organized by complexity. We assess capabilities from bibliographical recall to advanced jurisprudential reasoning, including legal rationale identification (\textit{`illah}), analogical application (\textit{qiyās}), comparative rule distinction, and cross-school synthesis, under both zero-shot and few-shot prompting conditions.

Taken together, our results reveal severe knowledge gaps across all evaluated models. The best system reaches 67.65\% correctness with 21.25\% hallucination, while several models fall below 35\% correctness and exceed 55\% hallucination. Few-shot prompting offers almost no benefit: only 2 of 9 models improve by more than 1\%, indicating that Islamic legal content is largely absent from pretraining and that prompting cannot substitute for foundational texts. We further identify a ``Mid-Complexity Failure Zone": moderate-complexity tasks that require specific, text-anchored doctrinal recall show the highest degree of failure, with correctness often between 8.82 and 47.06\% and hallucination between 35 and 73\%, whereas high-complexity reasoning tasks fare comparatively better (47.06 to 82.35\%) because models rely on generalized argumentative scaffolding and can maintain abstract reasoning without committing to exact doctrinal text. This pattern yields thousands of instances where fluent argumentation masks missing jurisprudential grounding rather than demonstrating genuine competence. The safety risks are also substantial. Six of the nine models accept misleading Islamic legal assumptions at rates above 40\% (worst 86.27\%), and few-shot prompting increases this behavior by 3.49\%. 

These findings imply that progress will not come from prompt engineering alone, but from pretraining on comprehensive Islamic legal corpora.  In-Context Learning (ICL) failures show that few-shot examples cannot supply content never seen during pretraining. Our benchmark provides the first systematic evaluation framework for Islamic legal AI systems and exposes critical weaknesses in platforms that Muslims increasingly treat as sources of religious guidance. Given these documented knowledge gaps and failure modes, general-purpose LLMs are not ready for Islamic legal advisory use without substantial improvements in foundational training, rigorous expert oversight, and explicit communication of limitations. Only community-driven efforts to train specialized models, strengthen evaluation, and establish ethical deployment guidelines can ensure that Islamic legal AI serves rather than undermines the spiritual needs of Muslim communities.

\section*{Funding and Acknowledgments}

This research was funded by the Qatar University- High Impact Grant- QUHI-CLAW-25/ 26-775. Portions of the data collection were conducted with support from the Program in Islamic Law (PIL) at Harvard Law School through the SHARIAsource project.

\bibliographystyle{unsrt}  
\bibliography{references}  

@article{Dahl_first,
   title={Large Legal Fictions: Profiling Legal Hallucinations in Large Language Models},
   volume={16},
   ISSN={1946-5319},
   url={http://dx.doi.org/10.1093/jla/laae003},
   DOI={10.1093/jla/laae003},
   number={1},
   journal={Journal of Legal Analysis},
   publisher={Oxford University Press (OUP)},
   author={Dahl, Matthew and Magesh, Varun and Suzgun, Mirac and Ho, Daniel E},
   year={2024},
   month=jan, pages={64–93} }

@article{Dahl_second,
author = {Magesh, Varun and Surani, Faiz and Dahl, Matthew and Suzgun, Mirac and Manning, Christopher and Ho, Daniel},
year = {2025},
month = {04},
pages = {},
title = {Hallucination‐Free? Assessing the Reliability of Leading {AI} Legal Research Tools},
volume = {22},
journal = {Journal of Empirical Legal Studies},
doi = {https://doi.org/10.1111/jels.12413}
}

@inproceedings{atif2025sacredsyntheticevaluatingllm,
  title={Sacred or Synthetic? Evaluating LLM Reliability and Abstention for Religious Questions},
  author={Atif, Farah and Askarbekuly, Nursultan and Darwish, Kareem and Choudhury, Monojit},
  booktitle={Proceedings of the AAAI/ACM Conference on AI, Ethics, and Society},
  volume={8},
  number={1},
  pages={217--226},
  year={2025}
}

@inproceedings{guha2023legalbench,
 author = {Guha, Neel and Nyarko, Julian and Ho, Daniel and R\'{e}, Christopher and Chilton, Adam and K, Aditya and Chohlas-Wood, Alex and Peters, Austin and Waldon, Brandon and Rockmore, Daniel and Zambrano, Diego and Talisman, Dmitry and Hoque, Enam and Surani, Faiz and Fagan, Frank and Sarfaty, Galit and Dickinson, Gregory and Porat, Haggai and Hegland, Jason and Wu, Jessica and Nudell, Joe and Niklaus, Joel and Nay, John and Choi, Jonathan and Tobia, Kevin and Hagan, Margaret and Ma, Megan and Livermore, Michael and Rasumov-Rahe, Nikon and Holzenberger, Nils and Kolt, Noam and Henderson, Peter and Rehaag, Sean and Goel, Sharad and Gao, Shang and Williams, Spencer and Gandhi, Sunny and Zur, Tom and Iyer, Varun and Li, Zehua},
 booktitle = {Advances in Neural Information Processing Systems},
 editor = {A. Oh and T. Naumann and A. Globerson and K. Saenko and M. Hardt and S. Levine},
 pages = {44123--44279},
 publisher = {Curran Associates, Inc.},
 title = {LegalBench: A Collaboratively Built Benchmark for Measuring Legal Reasoning in Large Language Models},
 url = {https://proceedings.neurips.cc/paper_files/paper/2023/file/89e44582fd28ddfea1ea4dcb0ebbf4b0-Paper-Datasets_and_Benchmarks.pdf},
 volume = {36},
 year = {2023}
}

@misc{openai_gpt4,
  author       = {OpenAI},
  title        = {{GPT-4}},
  year         = {2023},
  howpublished = {\url{https://openai.com/index/gpt-4-research/}},
  note         = {Accessed 2025-09-01}
}

@misc{openai_gpt35,
  author       = {OpenAI},
  title        = {{GPT-3.5}},
  year         = {2022},
  howpublished = {\url{https://platform.openai.com/docs/models/gpt-3.5-turbo}},
  note         = {Accessed 2025-09-01}
}

@misc{meta_llama2,
  author       = {Meta AI},
  title        = {{LLaMA 2}},
  year         = {2023},
  howpublished = {\url{https://www.llama.com/llama2/}},
  note         = {Accessed 2025-09-01}
}

@misc{google_palm2,
  author       = {Google AI},
  title        = {{PaLM 2}},
  year         = {2023},
  howpublished = {\url{https://blog.google/technology/ai/google-palm-2-ai-large-language-model/}},
  note         = {Accessed 2025-09-01}
}

@misc{lexisplus_ai,
  author       = {LexisNexis},
  title        = {{Lexis+ AI}},
  year         = {2023},
  howpublished = {\url{https://www.lexisnexis.com/en-us/products/lexis-plus-ai.page}},
  note         = {Accessed: 2025-09-01}
}

@misc{westlaw_edge,
  author       = {Thomson Reuters},
  title        = {{Westlaw Edge}},
  year         = {2024},
  howpublished = {\url{https://legal.thomsonreuters.com/en/products/westlaw-edge}},
  note         = {Accessed: 2025-09-01}
}

@misc{thomsonreuters_practical_law,
  author       = {Thomson Reuters},
  title        = {{Practical Law US}},
  year         = {2024},
  howpublished = {\url{https://content.next.westlaw.com/practical-law/?transitionType=Default&contextData=%28sc.Default%29}},
  note         = {Accessed: 2025-09-01}
}

@misc{togetherai2025,
  author       = {{Together AI}},
  title        = {Together {AI – The AI} Acceleration Cloud},
  year         = {2025},
  url          = {https://www.together.ai/},
  note         = {Accessed: 2025-09-11. Together AI provides an AI acceleration cloud platform for developing, fine-tuning, and deploying large-scale generative AI models. It supports a wide range of open-source and commercial models across multiple modalities, including language, vision, code, and audio. The platform emphasizes collaborative model development, scalable training infrastructure, and deployment tools for both research and production environments.}
}

@misc{openai2025introducing-gpt5,
  title        = {Introducing {GPT-5}},
  author       = {{OpenAI}},
  year         = {2025},
  month        = aug,
  note         = {Retrieved September 10, 2025},
  howpublished = {\url{https://openai.com/index/introducing-gpt-5/}}
}

@misc{openai2025introducing-o3-o4mini,
  title        = {Introducing {OpenAI} o3 and o4-mini},
  author       = {{OpenAI}},
  year         = {2025},
  month        = apr,
  note         = {Retrieved September 10, 2025},
  howpublished = {\url{https://openai.com/index/introducing-o3-and-o4-mini/}}
}

@misc{google2025gemini25flash,
  title        = {{Gemini 2.5 Flash - Generative {AI} on Vertex AI}},
  author       = {{Google Cloud}},
  year         = {2025},
  note         = {Retrieved September 10, 2025},
  howpublished = {\url{https://cloud.google.com/vertex-ai/generative-ai/docs/models/gemini/2-5-flash}}
}

@misc{openai2025introducing-gpt-oss,
  title        = {Introducing gpt-oss},
  author       = {{OpenAI}},
  year         = {2025},
  month        = aug,
  note         = {Retrieved September 10, 2025},
  howpublished = {\url{https://openai.com/index/introducing-gpt-oss/}}
}

@misc{deepseek2025r10528,
  title        = {{DeepSeek-R1-0528 Release}},
  author       = {{DeepSeek}},
  year         = {2025},
  month        = may,
  note         = {Retrieved September 10, 2025},
  howpublished = {\url{https://api-docs.deepseek.com/news/news250528}}
}

@misc{qwen2025qwen3,
  title        = {{Qwen3: Think Deeper, Act Faster}},
  author       = {{Qwen Team}},
  year         = {2025},
  month        = apr,
  note         = {Retrieved September 10, 2025},
  howpublished = {\url{https://qwenlm.github.io/blog/qwen3/}}
}

@misc{meta2025llama4,
  title        = {Llama 4: Multimodal Intelligence},
  author       = {{Meta AI}},
  year         = {2025},
  note         = {Retrieved September 10, 2025},
  howpublished = {\url{https://ai.meta.com/blog/llama-4-multimodal-intelligence/}}
}

@misc{xai2025grok4,
  title        = {Grok 4},
  author       = {{xAI}},
  year         = {2025},
  month        = jul,
  note         = {Retrieved September 10, 2025},
  howpublished = {\url{https://x.ai/news/grok-4}}
}

@misc{anthropic2025claude45,
  title        = {{Introducing Claude Sonnet 4.5}},
  author       = {{Anthropic}},
  year         = {2025},
  month        = sep,
  note         = {Retrieved September 29, 2025},
  howpublished = {\url{https://www.anthropic.com/news/claude-sonnet-4-5}}
}

@misc{deepmind_gemini2.5pro_2025,
  title        = {{Gemini 2.5 Pro: DeepMind's Advanced Reasoning Model}},
  author       = {{Google DeepMind}},
  howpublished = {\url{https://deepmind.google/models/gemini/pro/}},
  year         = {2025},
  note         = {Accessed: 2025-10-16; “our most advanced model for complex tasks”} ,
  url          = {https://deepmind.google/models/gemini/pro/}
}

@misc{OpenAI_GPT5Mini_2025,
  title        = {{GPT-5 Mini} - OpenAI Model Documentation},
  author       = {OpenAI},
  howpublished = {\url{https://platform.openai.com/docs/models/gpt-5-mini}},
  year         = {2025},
  note         = {Accessed: 2025-10-16},
  url          = {https://platform.openai.com/docs/models/gpt-5-mini}
}

@misc{OpenAI_Hello_GPT4o_2024,
  title        = {Hello {GPT-4o}},
  author       = {OpenAI},
  howpublished = {\url{https://openai.com/index/hello-gpt-4o/}},
  year         = {2024},
  note         = {Accessed: 2025-10-16},
  url          = {https://openai.com/index/hello-gpt-4o/}
}

@misc{meta2024llama31,
  title        = {Introducing {LLaMA 3.1}: Our Most Capable Models to Date},
  author       = {{Meta AI}},
  year         = {2024},
  month        = jul,
  day          = {23},
  note         = {Retrieved September 10, 2025},
  howpublished = {\url{https://ai.meta.com/blog/meta-llama-3-1/}}
}

@misc{iracmethod2025,
  author       = {IRAC Method},
  title        = {What is the {IRAC} Method - Everything You Need to Know},
  year         = {2025},
  url          = {https://www.iracmethod.com/irac-methodology},
  note         = {Accessed: 2025-09-11}
}

@book{kuwaiti_fiqh_encyclopedia,
  title        = {{Al-Mawsu'ah al-Fiqhiyyah al-Kuwaitiyyah} / {Kuwaiti Encyclopedia of {Islamic} Jurisprudence}},
  editor       = {{Department of Islamic Research and Encyclopedias, Kuwait Ministry of Awqaf and Islamic Affairs}},
  year         = {1983--2005},
  publisher    = {{Kuwait Ministry of Awqaf and Islamic Affairs}},
  address      = {Kuwait},
  volumes      = {45},
  pages        = {~17680},
  language     = {Arabic},
  note         = {Compiled over several decades (project initiated c.1965; editions printed across 1404--1427 AH). Major modern fiqh encyclopedia covering the four Sunni madhahib. Electronic copies/volume pages available online.},
  url          = {https://muslim-library.com/keoj/},
  urldate      = {2025-09-16}
}

@online{IslamQA,
  title        = {{Islam Q\&A (\url{IslamQA.org})}},
  author       = {Muhammad Saalih al-Munajjid (General Supervisor)},
  year         = {1996},
  note         = {Online repository of Islamic questions and answers from the four Sunni madhāhib, maintained under the supervision of Shaykh Muhammad Saalih al-Munajjid.},
  url          = {https://islamqa.org/},
  urldate      = {2025-09-17},
  language     = {English / Arabic},
}

@article{Imtiaz_Shafique_2025_AlIrfan_LLMs,
  author       = {Maria Imtiaz and Hafiz M. Mudassar Shafique},
  title        = {Large Language Models ({LLMs}) as {Islamic} Guidance Tools: Trust, Limitations, and Ethical Boundaries},
  journal      = {Al-Irfan: Research Journal of Islamic Studies},
  volume       = {10},
  number       = {19},
  year         = {2025},
  publisher    = {Faculty of Islamic Studies & Shariah, Minhaj University Lahore},  
  issn         = {2518-9794 (Print), 2788-4066 (Online)},  
  url          = {https://www.researchgate.net/publication/395336438_Open_Access_Al-IrfanResearch_Journal_of_Islamic_Studies_Large_Language_Models_LLMs_as_Islamic_Guidance_Tools_Trust_Limitations_and_Ethical_Boundaries},  
  urldate       = {2025-09-22}
}

@inproceedings{Khair_Sawalha_2025_AutomatedTranslation,
  author       = {Mohammad Mohammad Khair and Majdi Sawalha},
  title        = {Automated Translation of {Islamic} Literature Using Large Language Models: Al-Shamela Library Application},
  booktitle    = {Proceedings of the New Horizons in Computational Linguistics for Religious Texts},
  year         = {2025},
  address      = {Abu Dhabi, UAE},
  publisher    = {Association for Computational Linguistics},
  pages        = {53--58},
  url          = {https://aclanthology.org/2025.clrel-1.5.pdf},
  urldate       = {2025-09-22}
}

@article{aldahoul2025benchmarking,
  title={Benchmarking the Legal Reasoning of LLMs in Arabic Islamic Inheritance Cases},
  author={AlDahoul, Nouar and Zaki, Yasir},
  journal={arXiv preprint arXiv:2508.15796},
  year={2025}
}

@online{Ho2024_HallucinatingLaw,
  author        = {Daniel E. Ho},
  title         = {“Hallucinating Law: Legal Mistakes with Large Language Models are Pervasive”},
  year          = {2024},
  month         = {January},
  day           = {11},
  howpublished  = {Stanford HAI News},
  url           = {https://hai.stanford.edu/news/hallucinating-law-legal-mistakes-large-language-models-are-pervasive},
  urldate        = {2025-09-22},
  note          = {Reports findings from a Stanford RegLab / HAI preprint on legal hallucinations in LLMs.}
}

@misc{sunnah_ar_en_dataset_2025,
  title        = {{sunnah\_ar\_en\_dataset: Arabic–English Hadith Collection (14 Books)}},
  author       = {Slepovichev, Ivan},
  howpublished = {\url{https://huggingface.co/datasets/gurgutan/sunnah_ar_en_dataset}},
  year         = {2025},
  month        = {March},
  note         = {Accessed: 2025-09-22. License: MIT. Over 50,762 hadiths with bilingual metadata.},
  url          = {https://huggingface.co/datasets/gurgutan/sunnah_ar_en_dataset}
}

@misc{quran_nlp,
  title        = {{QURAN-NLP: Quran, Hadith, Translations, Tafaseer, Corpus Linguistics. Everything for NLP}},
  author       = {{islamAndAi} (GitHub organization)},
  howpublished = {\url{https://github.com/islamAndAi/QURAN-NLP}},
  year         = {2025},
  note         = {Accessed: 2025-09-22},
  url          = {https://github.com/islamAndAi/QURAN-NLP}
}

@inproceedings{alqurishi2022aralegalbertpretrainedlanguagemodel,
    title = "{A}ra{L}egal-{BERT}: A pretrained language model for {A}rabic Legal text",
    author = "Al-qurishi, Muhammad  and
      Alqaseemi, Sarah  and
      Souissi, Riad",
    editor = "Aletras, Nikolaos  and
      Chalkidis, Ilias  and
      Barrett, Leslie  and
      Goanț{\u{a}}, C{\u{a}}t{\u{a}}lina  and
      Preoțiuc-Pietro, Daniel",
    booktitle = "Proceedings of the Natural Legal Language Processing Workshop 2022",
    month = dec,
    year = "2022",
    address = "Abu Dhabi, United Arab Emirates (Hybrid)",
    publisher = "Association for Computational Linguistics",
    url = "https://aclanthology.org/2022.nllp-1.31/",
    doi = "10.18653/v1/2022.nllp-1.31",
    pages = "338--344"
}

@article{namoun2024multimodal,
  title        = {A Multimodal Data Scraping Tool for Collecting Authentic {Islamic} Text Datasets},
  author       = {Abdallah Namoun and Mohammad Ali Humayun and Waqas Nawaz},
  journal      = {International Journal of Advanced Computer Science and Applications (IJACSA)},
  volume       = {15},
  number       = {12},
  year         = {2024},
  pages        = {219--230},
  doi          = {10.14569/IJACSA.2024.0151224},
  note         = {Downloaded from TheSIA.org on 2025-09-22},
  url          = {https://thesai.org/Downloads/Volume15No12/Paper_24-A_Multimodal_Data_Scraping_Tool.pdf},
  urldate       = {2025-09-22}
}

@misc{IslamWeb,
  author       = {{IslamWeb}},
  title        = {IslamWeb: English Fatwa, Articles, Quran Recitations, and Islamic Resources},
  year         = {1998},
  howpublished = {\url{https://www.islamweb.net/en/}},
  note         = {Accessed: 2025-09-22}
}

@inproceedings{suresh_guttag_2019_harm_framework,
author = {Suresh, Harini and Guttag, John},
title = {A Framework for Understanding Sources of Harm throughout the Machine Learning Life Cycle},
year = {2021},
isbn = {9781450385534},
publisher = {Association for Computing Machinery},
address = {New York, NY, USA},
url = {https://doi.org/10.1145/3465416.3483305},
doi = {10.1145/3465416.3483305},
booktitle = {Proceedings of the 1st ACM Conference on Equity and Access in Algorithms, Mechanisms, and Optimization},
articleno = {17},
numpages = {9},
location = {--, NY, USA},
series = {EAAMO '21}
}

@article{AL-Smadi_2025_QU-NLP_QIAS,
  title        = {{QU-NLP at QIAS 2025 Shared Task: A Two-Phase LLM Fine-Tuning and Retrieval-Augmented Generation Approach for {Islamic} Inheritance Reasoning}},
  author       = {Mohammad AL-Smadi},
  journal      = {arXiv preprint},
  eprint       = {2508.15854},
  year         = {2025},
  doi          = {10.48550/arXiv.2508.15854},
  note         = {Submitted 20 Aug 2025; Accessed via ResearchGate / arXiv},
  url          = {https://arxiv.org/abs/2508.15854},
  urldate      = {2025-09-22}
}

@misc{QIAS2025,
  author       = {QIAS Organizing Committee},
  title        = {{QIAS} 2025: Question-and-Answer in {Islamic} Studies Assessment Shared Task},
  year         = {2025},
  url          = {https://sites.google.com/view/qias2025},
  note         = {Accessed: 2025-09-22},
  howpublished = {Online},
  institution  = {Hamad Bin Khalifa University, Qatar University, Northwestern University in Qatar, and CILE Center},
}

@article{bahaj2025mizanqabenchmarkinglargelanguage,
  title        = {{MizanQA}: Benchmarking Large Language Models on Moroccan Legal Question Answering},
  author       = {Bahaj, Adil and Ghogho, Mounir},
  journal      = {arXiv preprint arXiv:2508.16357},
  year         = {2025},
  archivePrefix = {arXiv},
  eprint       = {2508.16357},
  primaryClass = {cs.CL},
  url          = {https://arxiv.org/abs/2508.16357}
}

@book{qayyim1991,
  author       = {Ibn Qayyim al‑Jawziyyah},
  title        = {I\`{}lām al‑Muwaqqi\`{}īn \`{}an Rabb al‑\`{}Ālamīn},
  editor       = {\d{T}āhā \`{}Abd al‑Raʾūf Sa\`{}d},
  location     = {Cairo},
  publisher    = {Dār al‑Hadīth},
  year         = {1991},
}

@book{nawawi1996,
  author       = {Yaḥyā ibn Sharaf al‑Nawawī},
  title        = {Al‑Majmū\`{} Sharḥ al‑Muhadhdhab},
  editor       = {Muḥammad Najīb al‑Mutī\`{}ī},
  location     = {Beirut},
  publisher    = {Dār al‑Fikr},
  year         = {1996},
}

@book{abdAlBarr1994,
  author       = {Yūsuf ibn \`{}Abd Allāh Ibn \`{}Abd al‑Barr},
  title        = {Jāmi\`{} Bayān al‑\`{}Ilm wa Fadlihi},
  editor       = {Abū al‑Ashbāl al‑Zuhayrī},
  location     = {Riyadh},
  publisher    = {Dār Ibn al‑Jawzī},
  year         = {1994},
}

@inproceedings{bai2024mtbench,
    title = "{MT}-Bench-101: A Fine-Grained Benchmark for Evaluating Large Language Models in Multi-Turn Dialogues",
    author = "Bai, Ge  and
      Liu, Jie  and
      Bu, Xingyuan  and
      He, Yancheng  and
      Liu, Jiaheng  and
      Zhou, Zhanhui  and
      Lin, Zhuoran  and
      Su, Wenbo  and
      Ge, Tiezheng  and
      Zheng, Bo  and
      Ouyang, Wanli",
    editor = "Ku, Lun-Wei  and
      Martins, Andre  and
      Srikumar, Vivek",
    booktitle = "Proceedings of the 62nd Annual Meeting of the Association for Computational Linguistics (Volume 1: Long Papers)",
    month = aug,
    year = "2024",
    address = "Bangkok, Thailand",
    publisher = "Association for Computational Linguistics",
    url = "https://aclanthology.org/2024.acl-long.401/",
    doi = "10.18653/v1/2024.acl-long.401",
    pages = "7421--7454"
}

@inproceedings{liu2023geval,
    title = "{G}-Eval: {NLG} Evaluation using Gpt-4 with Better Human Alignment",
    author = "Liu, Yang  and
      Iter, Dan  and
      Xu, Yichong  and
      Wang, Shuohang  and
      Xu, Ruochen  and
      Zhu, Chenguang",
    editor = "Bouamor, Houda  and
      Pino, Juan  and
      Bali, Kalika",
    booktitle = "Proceedings of the 2023 Conference on Empirical Methods in Natural Language Processing",
    month = dec,
    year = "2023",
    address = "Singapore",
    publisher = "Association for Computational Linguistics",
    url = "https://aclanthology.org/2023.emnlp-main.153/",
    doi = "10.18653/v1/2023.emnlp-main.153",
    pages = "2511--2522"
}

@Article{rel15050541,
AUTHOR = {Latifi, Hasan},
TITLE = {Challenges of Using Artificial Intelligence in the Process of {Shi'i} Ijtihad},
JOURNAL = {Religions},
VOLUME = {15},
YEAR = {2024},
NUMBER = {5},
ARTICLE-NUMBER = {541},
URL = {https://www.mdpi.com/2077-1444/15/5/541},
ISSN = {2077-1444},
DOI = {10.3390/rel15050541}
}

@article{researchgate_ai_fatwa,
  title = {Artificial Intelligence on {Sunni} Islam's Fatwa Issuance in {Dubai and Egypt}},
  author = {Rahman, M. A. and others},
  journal = {ResearchGate},
  year = {2023},
  url = {https://www.researchgate.net/publication/370929203_Artificial_Intelligence_on_Sunni_Islam's_Fatwa_Issuance_in_Dubai_and_Egypt},
  note = {Accessed: 2025-01-08}
}

@misc{futurism_iran_ai,
  title = {Iran's Clerics Look to Harness {AI} to Issue Fatwas More Efficiently},
  author = {Futurism},
  year = {2023},
  url = {https://futurism.com/the-byte/iran-ai-fatwas},
  note = {Accessed: 2025-01-08}
}

@article{unimel_islamic_ai,
  title = {Islamic {AI} Chatbot: Integrating {AI} Technology with {Islamic} Jurisprudence},
  author = {Zakaria, A. H. M. and others},
  journal = {Journal of Legal Studies},
  volume = {12},
  number = {3},
  year = {2023},
  url = {https://unimel.edu.my/journal/index.php/JLG/article/view/1895},
  note = {Accessed: 2025-01-08}
}

@misc{spa2025robot,
  author       = {{Saudi Press Agency}},
  title        = {New {AI}‑Powered Robot Enhances Pilgrim Experience at Grand Mosque in {M}akkah},
  howpublished = {\url{https://spa.gov.sa/en/N2322621}},
  month        = may,
  year         = 2025,
  note         = {Accessed 2025‑07‑11},
}

@article{mori2012uncanny,
  title        = {The uncanny valley},
  author       = {Mori, Masahiro and MacDorman, Karl F. and Kageki, Norri},
  journal      = {IEEE Robotics \& Automation Magazine},
  volume       = {19},
  number       = {2},
  pages        = {98--100},
  year         = {2012},
  note         = {English translation of Mori's 1970 essay},
  publisher    = {IEEE}
}

@article{Mohammed2024AIandFatwas,
  author    = {Mohammed, Fadhil Karim},
  title     = {A Study of the Legal and Fiqhi Impact of Artificial Intelligence on Issuing Fatwas},
  journal   = {Kurdish Studies},
  volume    = {12},
  number    = {1},
  year      = {2024},
  pages     = {39--58},
  url       = {https://kurdishstudies.net/menu-script/index.php/KS/article/download/1535/1106/2999},
  note      = {ISSN 2051-4883},
}

@book{hallaq1997history,
  title     = {A History of Islamic Legal Theories: An Introduction to Sunni \textit{Usul al-Fiqh}},
  author    = {Hallaq, Wael B.},
  year      = {1997},
  publisher = {Cambridge University Press}
}

@misc{iifa_ai_guidelines,
  author       = {{International Islamic Fiqh Academy}},
  title        = {Second Day of Fiqh Academy's 26th Session Explores New Frontiers in {Islamic} Jurisprudence and Contemporary Issues},
  howpublished = {\url{https://iifa-aifi.org/en/53066.html}},
  month        = may,
  day          = 5,
  year         = 2025,
  note         = {Accessed 2025-07-11},
}

@inproceedings{hu2023wontfooledagainanswering,
    title = "Won{'}t Get Fooled Again: Answering Questions with False Premises",
    author = "Hu, Shengding  and
      Luo, Yifan  and
      Wang, Huadong  and
      Cheng, Xingyi  and
      Liu, Zhiyuan  and
      Sun, Maosong",
    editor = "Rogers, Anna  and
      Boyd-Graber, Jordan  and
      Okazaki, Naoaki",
    booktitle = "Proceedings of the 61st Annual Meeting of the Association for Computational Linguistics (Volume 1: Long Papers)",
    month = jul,
    year = "2023",
    address = "Toronto, Canada",
    publisher = "Association for Computational Linguistics",
    url = "https://aclanthology.org/2023.acl-long.309/",
    doi = "10.18653/v1/2023.acl-long.309",
    pages = "5626--5643"
}

@inproceedings{venkit2024hallucinations,
    title = "An Audit on the Perspectives and Challenges of Hallucinations in {NLP}",
    author = "Narayanan Venkit, Pranav  and
      Chakravorti, Tatiana  and
      Gupta, Vipul  and
      Biggs, Heidi  and
      Srinath, Mukund  and
      Goswami, Koustava  and
      Rajtmajer, Sarah  and
      Wilson, Shomir",
    editor = "Al-Onaizan, Yaser  and
      Bansal, Mohit  and
      Chen, Yun-Nung",
    booktitle = "Proceedings of the 2024 Conference on Empirical Methods in Natural Language Processing",
    month = nov,
    year = "2024",
    address = "Miami, Florida, USA",
    publisher = "Association for Computational Linguistics",
    url = "https://aclanthology.org/2024.emnlp-main.375/",
    doi = "10.18653/v1/2024.emnlp-main.375",
    pages = "6528--6548"
}

@inproceedings{muhamed2025refusalbench,
    title={RefusalBench: Generative Evaluation of Selective Refusal in Grounded Language Models},
    author={Aashiq Muhamed and Leonardo F. R. Ribeiro and Markus Dreyer and Virginia Smith and Mona T. Diab},
    booktitle={NeurIPS 2025 Workshop on Evaluating the Evolving LLM Lifecycle: Benchmarks, Emergent Abilities, and Scaling},
    year={2025},
    url={https://openreview.net/forum?id=EZR72ArmSS}
}

@article{kalai2025languagemodelshallucinate,
  title        = {Why Language Models Hallucinate},
  author       = {Kalai, Adam Tauman and Nachum, Ofir and Vempala, Santosh S. and Zhang, Edwin},
  journal      = {arXiv preprint arXiv:2509.04664},
  year         = {2025},
  archivePrefix = {arXiv},
  eprint       = {2509.04664},
  primaryClass = {cs.CL},
  url          = {https://arxiv.org/abs/2509.04664}
}

@inproceedings{tan2024judgebench,
  title     = {JudgeBench: A Benchmark for Evaluating LLM-based Judges},
  author    = {Tan, Sijun and Zhuang, Siyuan and Montgomery, Kyle and Tang, William Y. and Cuadron, Alejandro and Wang, Chenguang and Popa, Raluca Ada and Stoica, Ion},
  booktitle = {Proceedings of the International Conference on Learning Representations (ICLR)},
  year      = {2025},
  url       = {https://arxiv.org/abs/2410.12784},
}

@inproceedings{hashimoto-etal-2019-unifying,
    title = "Unifying Human and Statistical Evaluation for Natural Language Generation",
    author = "Hashimoto, Tatsunori B.  and
      Zhang, Hugh  and
      Liang, Percy",
    editor = "Burstein, Jill  and
      Doran, Christy  and
      Solorio, Thamar",
    booktitle = "Proceedings of the 2019 Conference of the North {A}merican Chapter of the Association for Computational Linguistics: Human Language Technologies, Volume 1 (Long and Short Papers)",
    month = jun,
    year = "2019",
    address = "Minneapolis, Minnesota",
    publisher = "Association for Computational Linguistics",
    url = "https://aclanthology.org/N19-1169/",
    doi = "10.18653/v1/N19-1169",
    pages = "1689--1701"
}

@inproceedings{liu-etal-2016-evaluate,
    title = "How {NOT} To Evaluate Your Dialogue System: An Empirical Study of Unsupervised Evaluation Metrics for Dialogue Response Generation",
    author = "Liu, Chia-Wei  and
      Lowe, Ryan  and
      Serban, Iulian  and
      Noseworthy, Mike  and
      Charlin, Laurent  and
      Pineau, Joelle",
    editor = "Su, Jian  and
      Duh, Kevin  and
      Carreras, Xavier",
    booktitle = "Proceedings of the 2016 Conference on Empirical Methods in Natural Language Processing",
    month = nov,
    year = "2016",
    address = "Austin, Texas",
    publisher = "Association for Computational Linguistics",
    url = "https://aclanthology.org/D16-1230/",
    doi = "10.18653/v1/D16-1230",
    pages = "2122--2132"
}

@inproceedings{smith-etal-2022-human,
    title = "Human Evaluation of Conversations is an Open Problem: comparing the sensitivity of various methods for evaluating dialogue agents",
    author = "Smith, Eric  and
      Hsu, Orion  and
      Qian, Rebecca  and
      Roller, Stephen  and
      Boureau, Y-Lan  and
      Weston, Jason",
    editor = "Liu, Bing  and
      Papangelis, Alexandros  and
      Ultes, Stefan  and
      Rastogi, Abhinav  and
      Chen, Yun-Nung  and
      Spithourakis, Georgios  and
      Nouri, Elnaz  and
      Shi, Weiyan",
    booktitle = "Proceedings of the 4th Workshop on NLP for Conversational AI",
    month = may,
    year = "2022",
    address = "Dublin, Ireland",
    publisher = "Association for Computational Linguistics",
    url = "https://aclanthology.org/2022.nlp4convai-1.8/",
    doi = "10.18653/v1/2022.nlp4convai-1.8",
    pages = "77--97"
}

@book{kahneman2011thinking,
  title={Thinking, Fast and Slow},
  author={Kahneman, D.},
  isbn={9780141033570},
  lccn={2011027143},
  series={Penguin: Psychology},
  url={https://books.google.com.qa/books?id=AV9x8XakdV0C},
  year={2011},
  publisher={Farrar, Straus and Giroux}
}

@book{dobelli2013art,
  title={The Art of Thinking Clearly},
  author={Dobelli, R.},
  isbn={9780062219701},
  url={https://books.google.com.qa/books?id=Zx2IMWJmJ_sC},
  year={2013},
  publisher={HarperCollins}
}

@article{dunningkruger,
author = {Kruger, Justin and Dunning, David},
year = {2000},
month = {01},
pages = {1121-34},
title = {Unskilled and Unaware of It: How Difficulties in Recognizing One's Own Incompetence Lead to Inflated Self-Assessments},
volume = {77},
journal = {Journal of Personality and Social Psychology},
doi = {10.1037//0022-3514.77.6.1121}
}

@inproceedings{chen2024benchmarking,
  title={Benchmarking large language models in retrieval-augmented generation},
  author={Chen, Jiawei and Lin, Hongyu and Han, Xianpei and Sun, Le},
  booktitle={Proceedings of the AAAI Conference on Artificial Intelligence},
  volume={38},
  number={16},
  pages={17754--17762},
  year={2024}
}

@article{siriwardhana2023improving,
  title={Improving the domain adaptation of retrieval augmented generation ({RAG}) models for open domain question answering},
  author={Siriwardhana, Shamane and Weerasekera, Rivindu and Wen, Elliott and Kaluarachchi, Tharindu and Rana, Rajib and Nanayakkara, Suranga},
  journal={Transactions of the Association for Computational Linguistics},
  volume={11},
  pages={1--17},
  year={2023},
  publisher={MIT Press One Broadway, 12th Floor, Cambridge, Massachusetts 02142, USA~…}
}

@article{ram2023context,
  title={In-context retrieval-augmented language models},
  author={Ram, Ori and Levine, Yoav and Dalmedigos, Itay and Muhlgay, Dor and Shashua, Amnon and Leyton-Brown, Kevin and Shoham, Yoav},
  journal={Transactions of the Association for Computational Linguistics},
  volume={11},
  pages={1316--1331},
  year={2023},
  publisher={MIT Press One Broadway, 12th Floor, Cambridge, Massachusetts 02142, USA~…}
}

@article{cheng2023lift,
  title={Lift yourself up: Retrieval-augmented text generation with self-memory},
  author={Cheng, Xin and Luo, Di and Chen, Xiuying and Liu, Lemao and Zhao, Dongyan and Yan, Rui},
  journal={Advances in Neural Information Processing Systems},
  volume={36},
  pages={43780--43799},
  year={2023}
}

@article{wu2024well,
  title={How well do {LLMs} cite relevant medical references? An evaluation framework and analyses},
  author={Wu, Kevin and Wu, Eric and Cassasola, Ally and Zhang, Angela and Wei, Kevin and Nguyen, Teresa and Riantawan, Sith and Riantawan, Patricia Shi and Ho, Daniel E and Zou, James},
  journal={Nature Communications},
  volume={16},
  number={1},
  pages={3615},
  year={2025},
  publisher={Nature Publishing Group UK London}
}

@article{tonmoy2024comprehensive,
  title={A comprehensive survey of hallucination mitigation techniques in large language models},
  author={Tonmoy, SMTI and Zaman, SM and Jain, Vinija and Rani, Anku and Rawte, Vipula and Chadha, Aman and Das, Amitava},
  journal={arXiv preprint arXiv:2401.01313},
  volume={6},
  year={2024}
}

@article{wang2023survey,
author = {Wang, Cunxiang and Liu, Xiaoze and Yue, Yuanhao and Guo, Qipeng and Hu, Xiangkun and Tang, Xiangru and Zhang, Tianhang and Jiayang, Cheng and Yao, Yunzhi and Hu, Xuming and Qi, Zehan and Gao, Wenyang and Wang, Yidong and Yang, Linyi and Wang, Jindong and Xie, Xing and Zhang, Zheng and Zhang, Yue},
title = {Survey on Factuality in Large Language Models},
year = {2025},
issue_date = {January 2026},
publisher = {Association for Computing Machinery},
address = {New York, NY, USA},
volume = {58},
number = {1},
issn = {0360-0300},
url = {https://doi.org/10.1145/3742420},
doi = {10.1145/3742420},
journal = {ACM Comput. Surv.},
month = sep,
articleno = {13},
numpages = {37}
}

@inproceedings{yu2023chain,
    title = "Chain-of-Note: Enhancing Robustness in Retrieval-Augmented Language Models",
    author = "Yu, Wenhao  and
      Zhang, Hongming  and
      Pan, Xiaoman  and
      Cao, Peixin  and
      Ma, Kaixin  and
      Li, Jian  and
      Wang, Hongwei  and
      Yu, Dong",
    editor = "Al-Onaizan, Yaser  and
      Bansal, Mohit  and
      Chen, Yun-Nung",
    booktitle = "Proceedings of the 2024 Conference on Empirical Methods in Natural Language Processing",
    month = nov,
    year = "2024",
    address = "Miami, Florida, USA",
    publisher = "Association for Computational Linguistics",
    url = "https://aclanthology.org/2024.emnlp-main.813/",
    doi = "10.18653/v1/2024.emnlp-main.813",
    pages = "14672--14685"
}

@misc{gao2023retrieval,
      title={Retrieval-Augmented Generation for Large Language Models: A Survey}, 
      author={Yunfan Gao and Yun Xiong and Xinyu Gao and Kangxiang Jia and Jinliu Pan and Yuxi Bi and Yi Dai and Jiawei Sun and Qianyu Guo and Meng Wang and Haofen Wang},
      year={2024},
      eprint={2312.10997},
      archivePrefix={arXiv},
      primaryClass={cs.CL}
}

@article{ji2023survey,
  title={Survey of hallucination in natural language generation},
  author={Ji, Ziwei and Lee, Nayeon and Frieske, Rita and Yu, Tiezheng and Su, Dan and Xu, Yan and Ishii, Etsuko and Bang, Ye Jin and Madotto, Andrea and Fung, Pascale},
  journal={ACM computing surveys},
  volume={55},
  number={12},
  pages={1--38},
  year={2023},
  publisher={ACM New York, NY}
}

@article{zhang2025siren,
  title={Siren's Song in the {AI} Ocean: A Survey on Hallucination in Large Language Models},
  author={Zhang, Yue and Li, Yafu and Cui, Leyang and Cai, Deng and Liu, Lemao and Fu, Tingchen and Huang, Xinting and Zhao, Enbo and Zhang, Yu and Chen, Yulong and others},
  journal={Computational Linguistics},
  pages={1--46},
  year={2025},
  publisher={MIT Press 255 Main Street, 9th Floor, Cambridge, Massachusetts 02142, USA~…}
}

@inproceedings{lucas2023fighting,
    title = "Fighting Fire with Fire: The Dual Role of {LLM}s in Crafting and Detecting Elusive Disinformation",
    author = "Lucas, Jason  and
      Uchendu, Adaku  and
      Yamashita, Michiharu  and
      Lee, Jooyoung  and
      Rohatgi, Shaurya  and
      Lee, Dongwon",
    editor = "Bouamor, Houda  and
      Pino, Juan  and
      Bali, Kalika",
    booktitle = "Proceedings of the 2023 Conference on Empirical Methods in Natural Language Processing",
    month = dec,
    year = "2023",
    address = "Singapore",
    publisher = "Association for Computational Linguistics",
    url = "https://aclanthology.org/2023.emnlp-main.883/",
    doi = "10.18653/v1/2023.emnlp-main.883",
    pages = "14279--14305"
}

@inproceedings{shuster2021retrieval,
    title = "Retrieval Augmentation Reduces Hallucination in Conversation",
    author = "Shuster, Kurt  and
      Poff, Spencer  and
      Chen, Moya  and
      Kiela, Douwe  and
      Weston, Jason",
    editor = "Moens, Marie-Francine  and
      Huang, Xuanjing  and
      Specia, Lucia  and
      Yih, Scott Wen-tau",
    booktitle = "Findings of the Association for Computational Linguistics: EMNLP 2021",
    month = nov,
    year = "2021",
    address = "Punta Cana, Dominican Republic",
    publisher = "Association for Computational Linguistics",
    url = "https://aclanthology.org/2021.findings-emnlp.320/",
    doi = "10.18653/v1/2021.findings-emnlp.320",
    pages = "3784--3803"
}

@article{cui2023chatlaw,
  title={Chatlaw: Open-source legal large language model with integrated external knowledge bases},
  author={Cui, Jiaxi and Li, Zongjian and Yan, Yang and Chen, Bohua and Yuan, Li},
  journal={CoRR},
  year={2023}
}

@inproceedings{savelka2023explaining,
  title     = {Explaining Legal Concepts with Augmented Large Language Models (GPT-4)},
  author    = {Savelka, Jaromir and Ashley, Kevin D. and Gray, Morgan A. and Westermann, Hannes and Xu, Huihui},
  booktitle = {Proceedings of the International Conference on Artificial Intelligence and Law (ICAIL 2023)},
  year      = {2023},
  address   = {Braga, Portugal},
  organization = {University of Minho Law School},
  url       = {https://arxiv.org/pdf/2306.09525.pdf},
  note      = {Published in the Proceedings of ICAIL 2023},
}

@inproceedings{tian2023fine,
  title={Fine-tuning language models for factuality},
  author={Tian, Katherine and Mitchell, Eric and Yao, Huaxiu and Manning, Christopher D and Finn, Chelsea},
  booktitle={The Twelfth International Conference on Learning Representations},
  year={2023}
}

@inproceedings{razumovskaia2024dial,
  title={Dial beinfo for faithfulness: Improving factuality of information-seeking dialogue via behavioural fine-tuning},
  author={Razumovskaia, Evgeniia and Vuli{\'c}, Ivan and Markovi{\'c}, Pavle and Cichy, Tomasz and Zheng, Qian and Wen, Tsung-Hsien and Budzianowski, Pawe{\l}},
  booktitle={Findings of the Association for Computational Linguistics: EMNLP 2024},
  pages={17139--17152},
  year={2024}
}

@inproceedings{zhang2023r,
    title = "{R}-Tuning: Instructing Large Language Models to Say `{I} Don{'}t Know'",
    author = "Zhang, Hanning  and
      Diao, Shizhe  and
      Lin, Yong  and
      Fung, Yi  and
      Lian, Qing  and
      Wang, Xingyao  and
      Chen, Yangyi  and
      Ji, Heng  and
      Zhang, Tong",
    editor = "Duh, Kevin  and
      Gomez, Helena  and
      Bethard, Steven",
    booktitle = "Proceedings of the 2024 Conference of the North American Chapter of the Association for Computational Linguistics: Human Language Technologies (Volume 1: Long Papers)",
    month = jun,
    year = "2024",
    address = "Mexico City, Mexico",
    publisher = "Association for Computational Linguistics",
    url = "https://aclanthology.org/2024.naacl-long.394/",
    doi = "10.18653/v1/2024.naacl-long.394",
    pages = "7113--7139"
}

@inproceedings{shi2024trusting,
  title={Trusting your evidence: Hallucinate less with context-aware decoding},
  author={Shi, Weijia and Han, Xiaochuang and Lewis, Mike and Tsvetkov, Yulia and Zettlemoyer, Luke and Yih, Wen-tau},
  booktitle={Proceedings of the 2024 Conference of the North American Chapter of the Association for Computational Linguistics: Human Language Technologies (Volume 2: Short Papers)},
  pages={783--791},
  year={2024}
}

@inproceedings{mallen2022not,
    title = "When Not to Trust Language Models: Investigating Effectiveness of Parametric and Non-Parametric Memories",
    author = "Mallen, Alex  and
      Asai, Akari  and
      Zhong, Victor  and
      Das, Rajarshi  and
      Khashabi, Daniel  and
      Hajishirzi, Hannaneh",
    editor = "Rogers, Anna  and
      Boyd-Graber, Jordan  and
      Okazaki, Naoaki",
    booktitle = "Proceedings of the 61st Annual Meeting of the Association for Computational Linguistics (Volume 1: Long Papers)",
    month = jul,
    year = "2023",
    address = "Toronto, Canada",
    publisher = "Association for Computational Linguistics",
    url = "https://aclanthology.org/2023.acl-long.546/",
    doi = "10.18653/v1/2023.acl-long.546",
    pages = "9802--9822"
}

@article{li2023inference,
  title={Inference-time intervention: Eliciting truthful answers from a language model},
  author={Li, Kenneth and Patel, Oam and Vi{\'e}gas, Fernanda and Pfister, Hanspeter and Wattenberg, Martin},
  journal={Advances in Neural Information Processing Systems},
  volume={36},
  pages={41451--41530},
  year={2023}
}

@inproceedings{chuang2023dola,
  title={DoLa: Decoding by Contrasting Layers Improves Factuality in Large Language Models},
  author={Yung-Sung Chuang and Yujia Xie and Hongyin Luo and Yoon Kim and James R. Glass and Pengcheng He},
  booktitle={The Twelfth International Conference on Learning Representations},
  year={2024},
  url={https://openreview.net/forum?id=Th6NyL07na}
}

@inproceedings{chern2023factool,
    title={FacTool: Factuality Detection in Generative {AI} -- A Tool Augmented Framework for Multi-Task and Multi-Domain Scenarios},
    author={Ethan Chern and Steffi Chern and Shiqi Chen and Weizhe Yuan and Kehua Feng and Chunting Zhou and Junxian He and Graham Neubig and Pengfei Liu},
    booktitle={Second Conference on Language Modeling},
    year={2025},
    url={https://openreview.net/forum?id=hJkQL9VtWT}
}

@article{peng2023check,
  title={Check your facts and try again: Improving large language models with external knowledge and automated feedback},
  author={Peng, Baolin and Galley, Michel and He, Pengcheng and Cheng, Hao and Xie, Yujia and Hu, Yu and Huang, Qiuyuan and Liden, Lars and Yu, Zhou and Chen, Weizhu and others},
  journal={arXiv preprint arXiv:2302.12813},
  year={2023}
}

@article{qin2024tool,
  title={Tool learning with foundation models},
  author={Qin, Yujia and Hu, Shengding and Lin, Yankai and Chen, Weize and Ding, Ning and Cui, Ganqu and Zeng, Zheni and Zhou, Xuanhe and Huang, Yufei and Xiao, Chaojun and others},
  journal={ACM Computing Surveys},
  volume={57},
  number={4},
  pages={1--40},
  year={2024},
  publisher={ACM New York, NY}
}

@inproceedings{
gou2023critic,
title={{CRITIC}: Large Language Models Can Self-Correct with Tool-Interactive Critiquing},
author={Zhibin Gou and Zhihong Shao and Yeyun Gong and yelong shen and Yujiu Yang and Nan Duan and Weizhu Chen},
booktitle={The Twelfth International Conference on Learning Representations},
year={2024},
url={https://openreview.net/forum?id=Sx038qxjek}
}

@article{kang2024comparing,
  title={Comparing hallucination detection metrics for multilingual generation},
  author={Kang, Haoqiang and Blevins, Terra and Zettlemoyer, Luke},
  journal={arXiv preprint arXiv:2402.10496},
  year={2024}
}

@article{cohen1960coefficient,
  title={A coefficient of agreement for nominal scales},
  author={Cohen, Jacob},
  journal={Educational and psychological measurement},
  volume={20},
  number={1},
  pages={37--46},
  year={1960},
  publisher={Sage Publications Sage CA: Thousand Oaks, CA}
}

@article{mirzadeh2024gsmsymbolic,
  title = {{GSM-Symbolic}: Understanding the Limitations of Mathematical Reasoning in Large Language Models},
  author = {Mirzadeh, Iman and Alizadeh, Keivan and Shahrokhi, Hooman and Tuzel, Oncel and Bengio, Samy and Farajtabar, Mehrdad},
  journal = {arXiv preprint arXiv:2410.05229},
  year = {2024},
  month = {October},
  eprint = {2410.05229},
  archivePrefix = {arXiv},
  primaryClass = {cs.LG},
  doi = {10.48550/arXiv.2410.05229},
  url = {https://arxiv.org/abs/2410.05229},
  note = {Apple Machine Learning Research}
}

@inproceedings{nezhurina2024alice,
  title = {Alice in Wonderland: Simple Tasks Showing Complete Reasoning Breakdown in State-Of-the-Art Large Language Models},
  author = {Nezhurina, Marianna and Cipolina-Kun, Lucia and Cherti, Mehdi and Jitsev, Jenia},
  booktitle = {International Conference on Learning Representations},
  year = {2025},
  eprint = {2406.02061},
  doi = {10.48550/arXiv.2406.02061},
  url = {https://arxiv.org/abs/2406.02061},
  note = {LAION and Juelich Supercomputing Center}
}

@inproceedings{berglund2024reversal,
  title = {The Reversal Curse: {LLMs} Trained on ``{A} is {B}'' Fail to Learn ``{B} is {A}''},
  author = {Berglund, Lukas and Tong, Meg and Kaufmann, Max and Balesni, Mikita and Stickland, Asa Cooper and Korbak, Tomasz and Evans, Owain},
  booktitle = {Proceedings of the International Conference on Learning Representations (ICLR)},
  year = {2024},
  month = {May},
  eprint = {2309.12288},
  archivePrefix = {arXiv},
  primaryClass = {cs.CL},
  doi = {10.48550/arXiv.2309.12288},
  url = {https://arxiv.org/abs/2309.12288},
  note = {Published at ICLR 2024}
}

@article{enstrom2024reasoning,
  title = {Reasoning in Transformers: Mitigating Spurious Correlations and Reasoning Shortcuts},
  author = {Enstr{\"o}m, Daniel and Kjellberg, Viktor and Johansson, Moa},
  journal = {arXiv preprint arXiv:2403.11314},
  year = {2024},
  month = {March},
  eprint = {2403.11314},
  archivePrefix = {arXiv},
  primaryClass = {cs.LG},
  doi = {10.48550/arXiv.2403.11314},
  url = {https://arxiv.org/abs/2403.11314},
  note = {University of Gothenburg and Chalmers University of Technology}
}

@inproceedings{
assessing2025spurious,
title={{ASSESSING} {ROBUSTNESS} {TO} {SPURIOUS} {CORRELATIONS} {IN} {POST}-{TRAINING} {LANGUAGE} {MODELS}},
author={Julia Shuieh and Prasann Singhal and Apaar Shanker and John Heyer and George Pu and Samuel Marc Denton},
booktitle={Workshop on Spurious Correlation and Shortcut Learning: Foundations and Solutions},
year={2025},
url={https://openreview.net/forum?id=5FUmGAZZ5w}
}

@inproceedings{explore2024spurious,
    title = "Explore Spurious Correlations at the Concept Level in Language Models for Text Classification",
    author = "Zhou, Yuhang  and
      Xu, Paiheng  and
      Liu, Xiaoyu  and
      An, Bang  and
      Ai, Wei  and
      Huang, Furong",
    editor = "Ku, Lun-Wei  and
      Martins, Andre  and
      Srikumar, Vivek",
    booktitle = "Proceedings of the 62nd Annual Meeting of the Association for Computational Linguistics (Volume 1: Long Papers)",
    month = aug,
    year = "2024",
    address = "Bangkok, Thailand",
    publisher = "Association for Computational Linguistics",
    url = "https://aclanthology.org/2024.acl-long.28/",
    doi = "10.18653/v1/2024.acl-long.28",
    pages = "478--492"
}

@article{navigating2024shortcuts,
  publtype={informal},
  author={David Steinmann and Felix Divo and Maurice Kraus and Antonia Wüst and Lukas Struppek and Felix Friedrich and Kristian Kersting},
  title={Navigating Shortcuts, Spurious Correlations, and Confounders: From Origins via Detection to Mitigation},
  year={2024},
  cdate={1704067200000},
  journal={CoRR},
  volume={abs/2412.05152},
  url={https://doi.org/10.48550/arXiv.2412.05152}
}

@article{mitigating2024capt,
  title={Mitigating Spurious Correlations in {LLMs} via Causality-Aware Post-Training},
  author={Gui, Shurui and Ji, Shuiwang},
  journal={arXiv preprint arXiv:2506.09433},
  year={2025}
}

@article{chain2025code,
  title={Chain-of-Code Collapse: Reasoning Failures in {LLMs} via Adversarial Prompting in Code Generation},
  author={Roh, Jaechul and Gandhi, Varun and Anilkumar, Shivani and Garg, Arin},
  journal={arXiv preprint arXiv},
  volume={2506},
  year={2025}
}

@article{chatziveroglou2025exploring,
  title = {Exploring {LLM} Reasoning Through Controlled Prompt Variations},
  author = {Chatziveroglou, Giannis and Yun, Richard and Kelleher, Maura},
  journal = {arXiv preprint arXiv:2504.02111},
  year = {2025},
  month = {April},
  eprint = {2504.02111},
  archivePrefix = {arXiv},
  primaryClass = {cs.CL},
  doi = {10.48550/arXiv.2504.02111},
  url = {https://arxiv.org/abs/2504.02111},
  note = {Massachusetts Institute of Technology}
}

@inproceedings{shortcut2024emnlp,
    title = "Do {LLM}s Overcome Shortcut Learning? An Evaluation of Shortcut Challenges in Large Language Models",
    author = "Yuan, Yu  and
      Zhao, Lili  and
      Zhang, Kai  and
      Zheng, Guangting  and
      Liu, Qi",
    editor = "Al-Onaizan, Yaser  and
      Bansal, Mohit  and
      Chen, Yun-Nung",
    booktitle = "Proceedings of the 2024 Conference on Empirical Methods in Natural Language Processing",
    month = nov,
    year = "2024",
    address = "Miami, Florida, USA",
    publisher = "Association for Computational Linguistics",
    url = "https://aclanthology.org/2024.emnlp-main.679/",
    doi = "10.18653/v1/2024.emnlp-main.679",
    pages = "12188--12200"
}

@misc{Bengio2022NeurIPSslides,
  author       = {Yoshua Bengio},
  title        = {Slides: Machine Learning Going Out of Labs, Into Society},
  howpublished = {Presentation at NeurIPS 2022 (ID 63207)},
  month        = nov,
  year         = {2022},
  note         = {https://neurips.cc/media/neurips-2022/Slides/63207.pdf}
}

@misc{StackOverflowPolicy2022,
  author       = {Stack Overflow Meta Community},
  title        = {Policy: Generative AI (e.g., ChatGPT) is banned},
  howpublished = {\url{https://meta.stackoverflow.com/questions/421831/policy-generative-ai-e-g-chatgpt-is-banned}},
  note         = {Accessed: 2025-11-19},
  year         = {2022},
  month        = {Dec},
  day          = {05}
}

@article{Schweitzer2025,
  author    = {Schweitzer, Sascha and Conrads, Markus},
  title     = {The digital transformation of jurisprudence: an evaluation of ChatGPT-4’s applicability to solve cases in business law},
  journal   = {Artificial Intelligence and Law},
  year      = {2025},
  volume    = {33},
  number    = {3},
  pages     = {847--871},
  doi       = {10.1007/s10506-024-09406-w},
  url       = {https://doi.org/10.1007/s10506-024-09406-w}
}

@article{Guo2025,
  author    = {Guo, Xue and Huang, Yuting and Wei, Bin and Kuang, Kun and Wu, Yiquan and Gan, Leilei and Huang, Xianshan and Dong, Xianglin},
  title     = {Specialized or general AI? A comparative evaluation of LLMs’ performance in legal tasks},
  journal   = {Artificial Intelligence and Law},
  year      = {2025},
  doi       = {10.1007/s10506-025-09460-y},
  url       = {https://doi.org/10.1007/s10506-025-09460-y}
}

@article{GoganiKhiabani2025,
  author    = {Gogani-Khiabani, Sina and Trivedi, Ashutosh and Chyi, ShinPing and Tizpaz-Niari, Saeid},
  title     = {Performance of LLMs on VITA test: potential for AI-assisted tax returns for low income taxpayers},
  journal   = {Artificial Intelligence and Law},
  year      = {2025},
  doi       = {10.1007/s10506-025-09465-7},
  url       = {https://doi.org/10.1007/s10506-025-09465-7}
}

@article{Trozze2025,
  author    = {Arianna Trozze and Toby Davies and Bennett Kleinberg},
  title     = {Large language models in cryptocurrency securities cases: can a GPT model meaningfully assist lawyers?},
  journal   = {Artificial Intelligence and Law},
  year      = {2025},
  volume    = {33},
  number    = {3},
  pages     = {691--737},
  doi       = {10.1007/s10506-024-09399-6},
  url       = {https://doi.org/10.1007/s10506-024-09399-6}
}

@misc{fanarpaper,
      title={Fanar: An Arabic-Centric Multimodal Generative AI Platform}, 
      author={Fanar Team and Ummar Abbas and Mohammad Shahmeer Ahmad and Firoj Alam and Enes Altinisik and Ehsannedin Asgari and Yazan Boshmaf and Sabri Boughorbel and Sanjay Chawla and Shammur Chowdhury and Fahim Dalvi and Kareem Darwish and Nadir Durrani and Mohamed Elfeky and Ahmed Elmagarmid and Mohamed Eltabakh and Masoomali Fatehkia and Anastasios Fragkopoulos and Maram Hasanain and Majd Hawasly and Mus'ab Husaini and Soon-Gyo Jung and Ji Kim Lucas and Walid Magdy and Safa Messaoud and Abubakr Mohamed and Tasnim Mohiuddin and Basel Mousi and Hamdy Mubarak and Ahmad Musleh and Zan Naeem and Mourad Ouzzani and Dorde Popovic and Amin Sadeghi and Husrev Taha Sencar and Mohammed Shinoy and Omar Sinan and Yifan Zhang and Ahmed Ali and Yassine El Kheir and Xiaosong Ma and Chaoyi Ruan},
      year={2025},
      eprint={2501.13944},
      archivePrefix={arXiv},
      primaryClass={cs.CL},
      url={https://arxiv.org/abs/2501.13944}, 
}

@misc{usul_ai,
  title        = {Usul - The Research Tool for Islamic Texts},
  howpublished = {\url{https://usul.ai/about}},
  note         = {AI-powered platform for Islamic research and retrieval over curated sources},
  year         = {2025},
  author       = {{Usul}},
  organization = {Seemore Foundation},
  year         = {2025}
}

@book{Susskind1998,
    author = {Susskind, Richard},
    title = {The Future of Law: Facing the Challenges of Information Technology},
    publisher = {Oxford University Press},
    year = {1998},
    month = {03},
    isbn = {9780198764960},
    doi = {10.1093/oso/9780198764960.001.0001},
    url = {https://doi.org/10.1093/oso/9780198764960.001.0001},
}

@book{Susskind2023,
    author = {Susskind, Richard},
    title = {Tomorrow's Lawyers: An Introduction to your Future},
    publisher = {Oxford University Press},
    year = {2023},
    month = {02},
    isbn = {9780192864727},
    doi = {10.1093/9780192864727.001.0001},
    url = {https://doi.org/10.1093/9780192864727.001.0001},
}

@book{GrantWischik2020,
  author    = {Thomas D. Grant and Damon J. Wischik},
  title     = {On the Path to AI},
  subtitle  = {Law’s Prophecies and the Conceptual Foundations of the Machine Learning Age},
  year      = {2020},
  publisher = {Palgrave Macmillan Cham},
  address   = {Cham},
  doi       = {10.1007/978-3-030-43582-0},
  isbn      = {978-3-030-43581-3},
  note      = {Hardcover ISBN: 978-3-030-43581-3; eBook ISBN: 978-3-030-43582-0},
  edition   = {1}
}

@InProceedings{Ronkainen2011,
  author    = {Anna Ronkainen},
  title     = {Dual-Process Cognition and Legal Reasoning},
  booktitle = {ARGUMENTATION 2011: INTERNATIONAL CONFERENCE ON ALTERNATIVE METHODS OF ARGUMENTATION IN LAW},
  publisher = {Masaryk University},
  year      = {2011},
  editor    = {Michał Araszkiewicz and others},
  pages     = {1-32},
  address   = {Brno},
  date      = {September 9, 2011}
}

\newpage

\appendix

\section{Comprehensive List of Source Texts}
\label{sec:all-books-table}

This appendix provides a comprehensive listing of all primary texts of Islamic law included in our source dataset. It includes full bibliographic details organized by school of thought, author, and historical period.

\renewcommand{\arraystretch}{1.2}
\begin{longtable}{|>{\raggedright\arraybackslash}p{1.5cm}|>{\raggedright\arraybackslash}p{5cm}|>{\raggedright\arraybackslash}p{5cm}|>{\raggedright\arraybackslash}p{3.5cm}|}

\caption{Foundational Texts of Islamic Law Used in the Source Dataset}\label{tab:bookstable} \\
\hline
\textbf{School} & \textbf{Title} & \textbf{Author} & \textbf{Period} \\ \hline
\endfirsthead

\multicolumn{4}{c}%
{{\bfseries \tablename\ \thetable{} -- continued from previous page}} \\
\hline
\textbf{School} & \textbf{Title} & \textbf{Author} & \textbf{Period} \\ \hline
\endhead

\hline \multicolumn{4}{|r|}{{Continued on next page}} \\ \hline
\endfoot

\hline
\endlastfoot
\textbf{Hanafi} & Mukhtasar al-Quduri & Abū al-\d{H}asan Aḥmad ibn Muḥammad al-Qudūrī & Early Classical (d. 428 H / 1037 CE) \\ \hline
Hanafi & Kitab al-Buyu' & Muḥammad ibn al-\d{H}asan al-Shaybānī & Early Classical (d. 189 H / 805 CE) \\ \hline
Hanafi & Badā`i` al-\d{S}anā`i` fī Tartīb al-Sharā`i` & `Alā' al-Dīn al-Kāsānī & Late Classical (d. 587 H / 1191 CE) \\ \hline
Hanafi & Majma` al-\d{D}amānāt & Ghānim ibn Muḥammad al-Baghdādī & Late Classical / Ottoman (d. 1030 H / 1621 CE) \\ \hline
Hanafi & al-Islām: `Aqīdah wa-Sharī`ah & Maḥmūd Shaltūt & Modern (d. 1963 CE) \\ \hline
Hanafi & al-Fatāwā al-Bazzāziyyah & Muḥammad ibn Muḥammad al-Bazzāzī (al-Kardarī) & Late Classical (d. 827 H / 1424 CE) \\ \hline
Hanafi & Al-Fatawa al-Hindiyyah (al-Fatāwā al-`Ālamgīriyyah) & Committee led by Shaykh Ni\d{z}ām al-Dīn al-Burhānpūrī & Late Classical / Mughal (Commissioned in 1664 CE) \\ \hline
Hanafi & Kitāb al-Mabsūṭ & Shams al-A'immah al-Sarakhsī & Classical (d. 483 H / 1090 CE) \\ \hline
Hanafi & Majallat al-Aḥkām al-`Adliyya & Committee of Ottoman Jurists & Ottoman (Promulgated 1869-1876 CE) \\ \hline \hline
\textbf{Maliki} & Al-Muwatta' & Imām Mālik ibn Anas & Early Classical (d. 179 H / 795 CE) \\ \hline
Maliki & Mukhtaṣar al-`Allāmah Khalīl & Khalīl ibn Isḥāq al-Jundī & Classical (d. 776 H / 1374 CE) \\ \hline
Maliki & Al-Qawānīn al-Fiqhiyyah & Ibn Juzayy al-Kalbī & Classical (d. 741 H / 1340 CE) \\ \hline
Maliki & Fatāwā al-Burzulī (Jāmi' Masā'il al-Aḥkām) & Abū al-Qāsim ibn Aḥmad al-Balawī al-Qayrawānī (al-Burzulī) & Classical (d. 841 H / 1438 CE) \\ \hline
Maliki & Fatāwā Ibn Rushd & Abū al-Walīd Muḥammad ibn Aḥmad Ibn Rushd (al-Jadd) & Almoravid (d. 520 H / 1126 CE) \\ \hline
Maliki & Kitab al-Kafi fi Fiqh Ahl al-Madina al-Maliki & Abū `Umar Yūsuf ibn `Abd Allāh ibn Muḥammad ibn `Abd al-Barr al-Namarī al-Qurṭubī & Classical (d. 463 H / 1071 CE) \\ \hline
Maliki & Anwār al-Burūq fī Anwā' al-Furūq & Shihāb al-Dīn al-Qarāfī & Mamluk (d. 684 H / 1285 CE) \\ \hline
Maliki & Fatāwā al-Shaykh al-Imām Muḥammad al-\d{T}āhir ibn `Āshūr & Muḥammad al-\d{T}āhir ibn `Āshūr & Modern (d. 1973 CE) \\ \hline \hline
\textbf{Shafi`i} & Al-Umm & Muḥammad ibn Idrīs al-Shāfi`ī & Early Classical (d. 204 H / 820 CE) \\ \hline
Shafi`i & Al-Hawi al-Kabir & al-Māwardī & Classical (d. 450 H / 1058 CE) \\ \hline
Shafi`i & Fatāwā al-Imām al-Ghazālī & Abū \d{H}āmid al-Ghazālī & High Classical (d. 505 H / 1111 CE) \\ \hline
Shafi`i & Al-Fatāwā al-Fiqhiyyah al-Kubrā & Ibn \d{H}ajar al-Haytamī & Late Mamluk / Early Ottoman (d. 974 H / 1566 CE) \\ \hline
Shafi`i & Fatḥ al-Wahhāb bi-Sharḥ Manhaj al-\d{T}ullāb & Shaykh al-Islām Zakariyyā al-Anṣārī & Late Mamluk / Early Ottoman (d. 926 H / 1520 CE) \\ \hline \hline
\textbf{Hanbali} & al-Istikhrāj li-Aḥkām al-Kharāj & Ibn Rajab al-\d{H}anbalī & Late Classical (Mamluk) (d. 795 H / 1393 CE) \\ \hline
Hanbali & al-Inṣāf fī Ma'rifat al-Rājiḥ min al-Khilāf & `Alā' al-Dīn al-Mardāwī & Late Classical (Mamluk) (d. 885 H / 1480 CE) \\ \hline
Hanbali & al-Mughnī & Ibn Qudāmah al-Maqdisī & Classical (Ayyubid) (d. 620 H / 1223 CE) \\ \hline
Hanbali & Dalīl al-\d{T}ālib li-Nayl al-Maṭālib & Mar`ī ibn Yūsuf al-Karmī al-\d{H}anbalī & Late Classical (Ottoman) (d. 1033 H / 1624 CE) \\ \hline
Hanbali & Mukhtaṣar al-Khiraqī & Abū al-Qāsim `Umar ibn al-\d{H}usayn al-Khiraqī & Early Classical (Abbasid) (d. 334 H / 945 CE) \\ \hline
Hanbali & al-Furū` wa-Ma`ahu Taṣḥīḥ al-Furū` & Shams al-Dīn Muḥammad ibn Mufliḥ al-\d{H}anbalī & Late Classical (Mamluk) (d. 763 H / 1362 CE) \\ \hline \hline
\textbf{Zahiri} & al-Iḥkām fī Uṣūl al-Aḥkām & 'Alī ibn Aḥmad ibn Sa'īd ibn \d{H}azm al-Andalusī & Classical (d. 456 H / 1064 CE) \\ \hline
Zahiri & al-Muḥallā bi-al-Āthār & 'Alī ibn Aḥmad ibn Sa'īd ibn \d{H}azm al-Andalusī & Classical (d. 456 H / 1064 CE) \\ \hline \hline
\textbf{Zaydi} & Sharḥ al-Tajrīd fī Fiqh al-Zaydiyyah & Aḥmad ibn Yaḥyā ibn al-Murtaḍā & Classical (d. 840 H / 1437 CE) \\ \hline
Zaydi & Al-Azhār fī Fiqh al-A'immah al-Aṭhār & al-Imām al-Mahdī li-Dīn Allāh Aḥmad ibn Yaḥyā al-Murtaḍā & Late Classical (d. 840 H / 1437 CE) \\ \hline
Zaydi & Sharḥ al-Tajrīd fī Fiqh al-Zaydiyyah & Aḥmad ibn al-\d{H}usayn al-Hārūnī al-\d{H}asanī & Post-Classical (d. 1055 H / 1645 CE) \\ \hline \hline
\textbf{Twelver Shi`i} (Ja'fari) & Al-Kāfī & Muḥammad ibn Ya`qūb al-Kulaynī & Early Classical (d. 329 H / 941 CE) \\ \hline
Twelver Shi`i (Ja'fari) & Tadhkirat al-Fuqahā' & al-`Allāmah al-\d{H}illī & Late Classical (d. 726 H / 1325 CE) \\ \hline
Twelver Shi`i (Ja'fari) & Taḥrīr al-Wasīlah & Ruhollah Khomeini & Contemporary (d. 1989 CE) \\ \hline
Twelver Shi`i (Ja'fari) & Jawāhir al-Kalām fī Sharḥ Sharā`i` al-Islām & Sheikh Muhammad Hassan al-Najafi & Modern (19th Century) (d. 1266 H / 1850 CE) \\ \hline
\end{longtable}
\renewcommand{\arraystretch}{1.0}

\section{Benchmarking Tasks}
\label{sec:benchtasks}

This section provides detailed descriptions of the task types used to evaluate LLMs' capabilities in Islamic legal reasoning. Our benchmarking framework systematically tests LLMs across three complexity levels, each targeting specific cognitive abilities required for Islamic jurisprudential analysis. The tasks are designed to assess whether LLMs can accurately retrieve information, apply legal rules, and engage in sophisticated reasoning comparable to trained Islamic legal scholars.

Our evaluation framework incorporates both \textbf{\textit{standard tasks}} and \textbf{\textit{false premise variants}}. The false premise variants (marked with prime notation: T1', T2', T3', T7') deliberately introduce incorrect assumptions to test whether models can reject false premise queries (FPQs) instead of confidently fabricating answers. These FPQs are designed in a way to test whether pre-trained models use their internal knowledge to rebut these falsely plausible questions and correct the wrongly stated question \cite{hu2023wontfooledagainanswering}. 

For all tasks requiring contextual background (excluding metadata recall tasks T1-T4), we employ a standardized template format that provides comprehensive contextual information before presenting the query. This approach ensures that LLMs have access to all necessary bibliographical and source details without requiring individual manipulation of each query for their respective texts.

\begin{genericbox}[blue]{Template Format: Context Provision Before Query}
\textbf{\textit{$<$Context$>$}:}\\
\textbf{Book Title:} [Title of the Islamic legal text]\\
\textbf{Author:} [Full name of the author]\\
\textbf{Volume-Page Reference:} [Specific volume and page numbers where applicable]\\
\textbf{Historical Period:} [Time period/century of composition or Year of Death of Author]\\
\textbf{School of Jurisprudence:} [Relevant madhhab where applicable]\\
\textbf{Note:} [Note for specific queries where applicable]\\
\textit{$<$/Context$>$} \\
---\\[2mm]
\textbf{Query:} [The specific question or task to be performed]
\end{genericbox}

This template format is utilized across all complexity levels for tasks that require relevant context for textual analysis, rule extraction, comparative reasoning, and jurisprudential synthesis.

\subsection{Evaluation Tasks}

\subsubsection{Low Complexity Tasks}

Low complexity tasks evaluate fundamental knowledge retrieval and basic verification capabilities. These tasks test the model's grounding in Islamic legal scholarship and its ability to access bibliographical and textual information accurately.

\paragraph{\textbf{T1-T3: Metadata Recall}}
These tasks evaluate the LLM's ability to retrieve basic bibliographical and contextual information about Islamic legal texts. LLMs must correctly identify the \textbf{jurisprudential school} (\textit{madhhab} ($n=34$)), \textbf{author} ($n=33$), and \textbf{historical period ($n=34$)} associated with specific works. This tests the model's grounding in the scholarly tradition and grasp of knowledge about Islamic Legal authoritative texts and its ability to distinguish between different schools of Islamic jurisprudence.

\begin{genericbox}[green]{Metadata Recall (T1-T3) Examples}
\textbf{T1 - School Identification:} What school of thought is associated with book: ``Al-Fatawa al-Hindiyyah (also known as al-Fatāwā al-`Ālamgīriyyah)''? Just return the school of thought without any other text or explanation\\ 
\textbf{Gold Answer:} Hanafi
\\[2mm]
\textbf{T2 - Author Identification:} Who is the author of book: ``al-Islām: `Aqīdah wa-Sharī`ah''? Just return the author without any other text or explanation.\\ 
\textbf{Gold Answer:} Maḥmūd Shaltūt
\\[2mm]
\textbf{T3 - Period Identification:} What historical period is book: ``Kitāb al-Mabsūṭ'' from? Just return the \textbf{\textit{century}} without any other text or explanation. \\ 
\textbf{Gold Answer}: 11th century CE (483 H / 1090 CE)
\end{genericbox}

For T3, we initially recorded year-level dates (e.g., 483 AH / 1090 CE) based on available metadata. However, because many of these texts are premodern, sources often disagree on what constitutes the “exact” year---for example, whether to use an estimated publication date, the date of composition, or the author's death. To eliminate ambiguity and ensure consistency, we standardized the temporal resolution to the century level and report only the Common Era (CE) dates (e.g., 483 H / 1090 CE $\rightarrow$ 11th century CE).

\paragraph{\textbf{T4: Concept Existence Verification ($n = 267$)}}
This task evaluates whether LLMs can determine, with source-specific precision, whether a specified Islamic legal concept is treated in a designated authoritative text. When the concept is present, success requires an accurate binary response of YES or NO.

\begin{genericbox}[green]{Concept Existence (T4) Example}
\textbf{Query:} Does book: ``Mukhtasar al-Quduri" by ``Abū al-\d{H}asan Aḥmad ibn Muḥammad al-Qudūrī" treat topic: ``Khiyar (Options) - Khiyar al-`Ayb (Option due to defect) in Bay` (Sales)"? Yes or No? Just return the answer without any other text or explanation.\\ 
\textbf{Gold Answer:} Yes, 81
\end{genericbox}

\paragraph{\textbf{T5: Source of Authority or Recall ($n = 83$)}}
This task assesses whether a model can identify, within a specified work, the author's citation of other jurists' opinions and accurately summarize the cited positions. It tests source-aware attribution, the delineation of intra-school disagreement, and the faithful recall of conditions or qualifications attached to the cited views.

\begin{genericbox}[green]{Source Authority (T5) Example}
\textbf{Query:} Does the author Muḥammad ibn al-\d{H}asan al-Shaybānī in the book ``Kitab al-Buyū`'' mention any other jurist's opinion regarding the permissibility of manufacturing contracts (\textit{istiṣnā`})? Answer Yes/No, and if yes, who? \\ 
\textbf{Gold Answer:} Yes. The author mentions the differing opinions of Abū \d{H}anīfah (who allows it under specific conditions like known type and weight if an advance is paid) versus Abū Yūsuf and Muḥammad (who allow it more broadly, giving the buyer an option upon seeing the item). \textit{(Qaala Abu Yusuf wa Muhammad: huwa jaa'iz wa saahibuhu bakhiyaar idhaa ra'ahu, in shaa' akhadhahu wa in shaa' tarakahu wa laa yakoon `abbazlat al-silm).}
\end{genericbox}

\paragraph{\textbf{T6: Legal Maxim Recall ($n = 13$)}}
This task requires models to extract and succinctly restate specific legal rules or maxims from Islamic legal codes. It evaluates the ability to distill complex principles into precise, self-contained statements while preserving their scope, conditions, and legal effect.

\begin{genericbox}[green]{Legal Maxim Recall (T6) Example}
\textbf{Query:} According to Article 74 on page 14, what is the legal status of a sale if the seller disposes of the sold item to another person before the first buyer has taken possession?\\ 
\textbf{Gold Answer:} If the seller disposes of the sold item before the buyer takes possession, the first sale is rendered void.
\end{genericbox}

\subsubsection{Moderate Complexity Tasks}

Moderate-complexity tasks require precise enumeration of legal conditions, comparative analysis, and synthesis of specific rules. They test exactness, concision, and the avoidance of extraneous material while extracting technical requirements from the sources.

\paragraph{\textbf{T7: Condition Enumeration (\textit{Shurūṭ}) ($n = 28$)}}
This task evaluates the LLM's ability to identify and list the specific conditions (\textit{shurūṭ}) required for the validity of Islamic legal transactions. \textit{Shurūṭ} (singular: \textit{sharṭ}) refers to the necessary conditions that must be met for a legal act to be considered valid in Islamic law.

\begin{genericbox}[orange]{Condition Enumeration (T7) Example}
\textbf{Query:} List the specific conditions (shurūṭ) for the validity of a Salām contract (forward contract) according to the text, as held by Abū \d{H}anīfa.\\ 
\textbf{Gold Answer:} According to Abū \d{H}anīfa, the validity of a Salām contract requires the explicit mention of the following conditions in the contract: 1. Known type (Jins ma`loom
), 2. Known variety (Naw ma`loom), 3. Known description (Wasfah ma`loomah), 4. Known quantity (Miqdaar ma`loom), 5. Known due date (Ajal ma`loom), and 6. Knowledge of the capital (Miqdaar ra's al-maal). Additionally, the capital must be taken before the parties separate, and neither the capital nor the subject of the contract can be disposed of before possession.
\end{genericbox}

\paragraph{\textbf{T8: Comparative Legal Distinction ($n = 21$)}}
This task evaluates the LLM's ability to delineate doctrinal differences between closely related legal rules or applications within Islamic jurisprudence, articulating the operative reasoning that distinguishes similar but legally distinct scenarios.

\begin{genericbox}[orange]{Comparative Distinction (T8) Example}
\textbf{Query:} On page 582, a man guarantees to present a debtor in court the next day, agreeing to pay the one hundred dinar debt if he fails. If he fails to present the debtor, what is the ruling according to Abū \d{H}anīfa and Abū Yūsuf, and what is the implicit condition in Muḥammad's differing view?\\ 
\textbf{Gold Answer:} According to Abū \d{H}anīfa and Abū Yūsuf, the guarantor is liable for the one hundred dinars. Muḥammad's differing view is conditioned on whether the debt was legally proven before the guarantee was made; if it was not, then the claimant's subsequent claim against the guarantor is not considered.
\end{genericbox}

\paragraph{\textbf{T9: Statute-like Synthesis from Legal Codes ($n = 57$)}}
This task evaluates the LLM's ability to synthesize and interpret codified Islamic legal rules, particularly from systematic legal codes, like the Ottoman \textit{Majalla}. The task requires understanding how specific legal principles apply to practical scenarios.

\begin{genericbox}[orange]{Statute Synthesis (T9) Example}
\textbf{Query:} Based on the discussion on pages 170-171, outline the process and objective of appointing two arbiters (ḥakamayn) in a case of marital discord (shiqāq).\\ 
\textbf{Gold Answer:} In a case of marital discord, the Qur'ān mandates the appointment of two arbiters: one from the husband's family and one from the wife's. The objective of this arbitration is reconciliation (iṣlāḥ). The arbiters are tasked with investigating the causes of the dispute and attempting to resolve them. If reconciliation is not possible, they are empowered to enact a separation between the spouses. This process is a means of resolving conflict that has escalated beyond the couple's ability to manage.
\end{genericbox}

\subsubsection{High Complexity Tasks}
High-complexity tasks probe the model's capacity to identify and articulate underlying jurisprudential reasoning (\textit{`illah}), compare legal principles across schools, and extend them to novel scenarios. They require methodological fidelity, careful qualification, and rigorous reasoning.

\paragraph{\textbf{T10: Legal Rationale Identification (\textit{`illah}) ($n = 63$)}}
This task evaluates the LLM's ability to identify and articulate the underlying legal reasoning (\textit{`illah}) behind specific Islamic legal rulings. \textit{`Illah} refers to the effective cause or wisdom behind a legal ruling that serves as the basis for analogical reasoning in Islamic jurisprudence.

\begin{genericbox}[purple]{Legal Rationale Identification (\textit{`illah}) (T10) Example}
\textbf{Query:} The text discusses the permissibility of selling a debt (Deen) for a debt (Deen) and the conditions under which it is allowed or prohibited. What is the underlying rationale for the general prohibition of selling a debt for a debt, and what specific condition, if met, allows for its permissibility according to the Hanafi school as presented?\\
\textbf{Gold Answer:} The underlying rationale for the general prohibition of selling a debt for a debt is to prevent riba al-nasi'ah (usury of delay) and gharar (uncertainty/risk). When a debt is exchanged for another debt, there is a risk of one party not being able to collect, or the value of the debts changing over time, leading to exploitation or dispute. However, the text implies permissibility if the debt is exchanged for a tangible, present item (`Ayn haadirah) that is immediately taken into possession. For example, if a person sells a debt they are owed for a specific, present commodity that they immediately receive, this removes the element of riba al-nasi'ah and reduces gharar, thus making the transaction permissible. The key condition for permissibility is the immediate and complete possession of at least one of the counter-values in the exchange.
\end{genericbox}

\paragraph{\textbf{T11: Analogical Application to Novel Scenarios ($n = 12$)}}
This task tests the LLM's ability to apply established Islamic legal principles to contemporary or novel situations not explicitly addressed in classical texts. The task requires understanding of analogical reasoning (\textit{qiyās}) and the ability to identify relevant similarities between classical and modern scenarios.

\begin{genericbox}[purple]{Analogical Application (T11) Example}
\textbf{Query:} Apply the text's principle regarding the prohibition of selling ``what is not in one's possession'' (Ma laysa fi milkihi) to a novel scenario: A person sells a specific car that they intend to buy from a dealership next week, before they have actually purchased it. Detail: 1. Principle, 2. Ambiguity, 3. Ruling based on principle.\\
\textbf{Gold Answer:}\\\textbf{Principle:} The text explicitly prohibits the sale of ``what is not in one's possession'' (mā laysa fī milkih) at the time of the contract. This principle is rooted in preventing gharar (excessive uncertainty) and ensuring that the seller has the ability to deliver the sold item, thereby avoiding disputes.\\\textbf{Ambiguity:} In the novel scenario, the seller has identified a specific car and intends to acquire it, but they do not yet own it at the moment of sale. While the car is identifiable, its ownership and thus the seller's ability to deliver it are contingent on a future event (the purchase from the dealership). This introduces uncertainty regarding the seller's immediate capacity to fulfill the contract. 
\\\textbf{Ruling based on principle:} Based on the principle of prohibiting the sale of what is not in one's possession, this transaction would be considered invalid. Even if the seller fully intends to acquire the car, the lack of actual ownership at the time of the contract makes the sale speculative and falls under the category of gharar. The sale would only become valid once the seller takes possession of the car and then enters into a new contract with the buyer.
\end{genericbox}

\paragraph{\textbf{T12: Cross-School Jurisprudential Synthesis ($n = 5$)}}
This task requires LLMs to compare and synthesize different legal positions across multiple schools of Islamic jurisprudence (\textit{madhāhib}). The task tests understanding of inter-school differences and the ability to identify the most widely accepted position (\textit{mashhūr}).

\begin{genericbox}[purple]{Cross-School Synthesis (T12) Example}
\textbf{Query:} Analyze the legal debate on page 353 concerning the effect of a voidable condition (sharṭ fāsid) on a sales contract. What are the two primary legal opinions on whether such a condition invalidates the entire contract, and what is the jurisprudential basis for each position?\\
\textbf{Gold Answer:} The text presents two main opinions on the effect of a voidable condition: The condition is void, but the contract is valid: This is the sound position of the Hanbali school. The reasoning is that the sale contract and the condition are two separate legal commitments. The invalidity of the ancillary condition does not nullify the primary contract, which meets all its essential requirements. Both the condition and the contract are void: This view is based on the Prophetic tradition prohibiting ``a sale and a condition" (nahā `an bay` wa sharṭ). Proponents of this view argue that the invalid condition corrupts the entire transaction, rendering the contract itself invalid.
\end{genericbox}

\paragraph{\textbf{T13: Legal Maxim Mapping (\textit{Qawā'id Fiqhiyyah}) ($n = 17$)}}
This task evaluates the LLM's ability to identify and apply broad legal maxims (\textit{qawā'id fiqhiyyah}) to specific legal scenarios. \textit{Qawā'id fiqhiyyah} are general principles that encapsulate fundamental concepts in Islamic jurisprudence and serve as interpretive tools for understanding specific rulings.

\begin{genericbox}[purple]{Legal Maxim Mapping (T13) Example}
\textbf{Query:} Article 3 on page 14 states, ``In contracts, consideration is given to the intentions and meanings, not to the words and forms.'' Explain the legal reasoning behind this principle and how it applies to the example of a ``sale with right of redemption'' (bay` bi-al-wafā').\\
\textbf{Gold Answer:} The legal reasoning is that the true legal effect of a contract is determined by the parties' actual intent, not merely the literal words they use. In the case of a ``sale with right of redemption'' (bay` bi-al-wafā'), although the parties use the language of a sale, their intention is to create a security arrangement. Therefore, the law treats it according to its underlying meaning as a pledge (rahn), not as a true sale."
\end{genericbox}

\subsection{False Premise Variants}
\label{subsec:falsepremisevariants}

Just like the ``\textit{Wrong Conclusion from False Premises}" error mentioned in Section \ref{subsec:generallegalrelatedwork}, when LLMs accept factually incorrect statements and do not correct the claim, it can have consequences in Islamic law. A human scholar or \textit{`aalim}, when faced with a query based on a false premise, would first correct the premise. An AI system, if not explicitly designed for this, may simply generate a plausible but entirely incorrect response. Therefore, a simple accuracy score is insufficient for a benchmark in this domain; a robust evaluation must also test a model's ability to identify and correct flawed premises, and to provide sound, verifiable justifications for its conclusions.

\subsubsection{Low Complexity False Premise Tasks}

\paragraph{\textbf{T1'-T3': False Premise Detection (Metadata)}}
These tasks assess a model's ability to detect incorrect attributions by deliberately introducing bibliographical errors. They test whether the model can verify and reject misinformation about Islamic legal texts and scholars. Each variant---false school, false author, and false period contains 15 examples.

\begin{genericbox}[red]{False Premise Tasks: Metadata Detection (T1'-T3') Examples}
\textbf{T1' - False School:} Analyzing the major Maliki work 'Badā'i` al-\d{S}anā'i` fī Tartīb al-Sharā'i`', what distinctive systematization methods does it employ that exemplify Maliki approaches to organizing legal principles?\\ 
\textbf{Reason:} This question contains a false premise. 'Badā'i` al-\d{S}anā'i` fī Tartīb al-Sharā'i`' is not particularly influential in Maliki circles - it is actually a major Hanafi legal text. \\[2mm]
\textbf{T2' - False Author:} How did al-`Allāmah al-\d{H}illī organize the hadith material in his work 'Al-Kāfī'? \\ 
\textbf{Reason:} This question contains a false premise. 'Al-Kāfī' was not organized by al-`Allāmah al-\d{H}illī. It was actually compiled by Muḥammad ibn Ya`qūb al-Kulaynī. \\[2mm]
\textbf{T3' - False Period:} As `Alā' al-Dīn al-Kāsānī wrote 'Kitāb al-Mabsūṭ' in the 20th-century CE Modern period, how does it address contemporary Islamic legal challenges?\\ 
\textbf{Reason:} This question contains two false premises. First, 'Kitāb al-Mabsūṭ' was not written by `Alā' al-Dīn al-Kāsānī but by Shams al-A'immah al-Sarakhsī. Second, it was not written in the 20th-century CE Modern period but is from the Classical period (d. 483 H / 1090 CE), written over 800 years earlier.
\end{genericbox}

\subsubsection{Moderate Complexity False Premise Tasks}

\paragraph{\textbf{T7': False Premise Detection (Condition Enumeration) ($n = 6$) Example}}
This task evaluates LLMs by deliberately attributing legal conditions to incorrect scholars. The task tests whether models can identify when detailed legal conditions are incorrectly attributed to the wrong jurist or school of thought.

\begin{genericbox}[red]{False Premise Task: Complex Attribution Detection (T7') Example}
\textbf{Query:} According to the ruling on page 145, a marriage contract is formed by offer (ījāb) and acceptance (qabūl) expressed in terms that signify marriage, and it is not valid unless it is witnessed by four free, adult, sane, and Muslim men. What happens if only three witnesses are present? \\ 
\textbf{Reason:} This is a false premise question. A marriage contract is formed by offer (ījāb) and acceptance (qabūl) expressed in terms that signify marriage. It is not valid unless it is witnessed by \textit{two} free, adult, sane, and Muslim men, or by one such man and two women, not four men as stated in the question. The false premise incorrectly doubles the witness requirement from two to four men.
\end{genericbox}

\section{Inter-Rate Reliability} \label{sec:inter_rater_appendix}
To assess the consistency between the LLM-based judge and human annotators, we computed Cohen's Kappa scores across two evaluation dimensions: (i) overall scores and (ii) hallucination detection. Table~\ref{tab:llm_human_reliability} summarizes the inter-rater reliability between the LLM and each human annotator, as well as the agreement between the two human annotators. The results demonstrate substantial to almost-perfect agreement across all comparisons, with an aggregated LLM–Human reliability of $\kappa = 0.81$.

\begin{table}[h!] \centering \caption{Inter-rater reliability between LLM and human annotators using Cohen's Kappa.} \label{tab:llm_human_reliability} \begin{tabular}{lcc} \toprule \textbf{Comparison} & \textbf{Scores Kappa} & \textbf{Hallucinations Kappa} \\ \midrule LLM vs. Human\_1 & 0.7787 & 0.7671 \\ LLM vs. Human\_2 & 0.8393 & 0.8650 \\ Human\_1 vs. Human\_2 & 0.7320 & 0.7214 \\ \midrule \textbf{Overall LLM--Human Mean} & \textbf{0.809} & \textbf{0.816} \\ \bottomrule \end{tabular} \end{table}

\end{document}